\documentclass{article}

\usepackage{microtype}
\usepackage{graphicx}
\usepackage{subfigure}
\usepackage{booktabs} 
\usepackage{tikz}
\usetikzlibrary{automata,positioning,arrows,arrows.meta}

\usepackage{hyperref}
\hypersetup{colorlinks,allcolors=black}
\usepackage{enumitem}
\setlist[itemize]{itemsep=0pt, parsep=0.5em, topsep=0pt}

\usepackage[accepted]{icml2023}

\usepackage{amsmath}
\usepackage{amssymb}
\usepackage{mathtools}
\usepackage{amsthm}

\usepackage[nameinlink,capitalize,noabbrev]{cleveref}
\crefname{equation}{Eq.}{Eqs.}
\crefname{assumption}{Assumption}{Assumptions}

\theoremstyle{plain}
\newtheorem{theorem}{Theorem}

\theoremstyle{definition}
\newtheorem{definition}{Definition}
\newtheorem{assumption}{Assumption}
\theoremstyle{remark}

\usepackage[textsize=tiny]{todonotes}

\icmltitlerunning{Leveraging Factored Action Spaces for Off-Policy Evaluation}

\begin{document}

\twocolumn[
\icmltitle{Leveraging Factored Action Spaces for Off-Policy Evaluation}



\icmlsetsymbol{equal}{*}

\begin{icmlauthorlist}
\icmlauthor{Aaman Rebello}{imperial}
\icmlauthor{Shengpu Tang}{umich}
\icmlauthor{Jenna Wiens}{umich}
\icmlauthor{Sonali Parbhoo}{imperial}
\end{icmlauthorlist}

\icmlaffiliation{imperial}{Department of Engineering, Imperial College London, London, UK}
\icmlaffiliation{umich}{Division of Computer Science \& Engineering, University of Michigan, Ann Arbor, Michigan, USA}

\icmlcorrespondingauthor{Aaman Rebello}{\href{mailto:aaman.rebello18@imperial.ac.uk}{aaman.rebello18@imperial.ac.uk}}
\icmlcorrespondingauthor{Sonali Parbhoo}{\href{mailto:s.parbhoo@imperial.ac.uk}{s.parbhoo@imperial.ac.uk}}

\icmlkeywords{Machine Learning, ICML}

\vskip 0.3in
]



\printAffiliationsAndNotice{}  

\begin{abstract}
Off-policy evaluation (OPE) aims to estimate the benefit of following a counterfactual sequence of actions, given data collected from executed sequences. However, existing OPE estimators often exhibit high bias and high variance in problems involving large, combinatorial action spaces. We investigate how to mitigate this issue using factored action spaces i.e. expressing each action as a combination of independent sub-actions from smaller action spaces. This approach facilitates a finer-grained analysis of how actions differ in their effects. In this work, we propose a new family of ``decomposed'' importance sampling (IS) estimators based on factored action spaces. Given certain assumptions on the underlying problem structure, we prove that the decomposed IS estimators have less variance than their original non-decomposed versions, while preserving the property of zero bias. Through simulations, we empirically verify our theoretical results, probing the validity of various assumptions. Provided with a technique that can derive the action space factorisation for a given problem, our work shows that OPE can be improved ``for free'' by utilising this inherent problem structure. 
\end{abstract}

\section{Introduction} \label{sec:introduction}

Off-policy evaluation (OPE) is useful in high-stakes decision making domains such as healthcare \citep{Gottesman2018}, robot control \cite{Pagnozzi2021} and recommender systems \citep{Theocharous2015}, where real-world testing of potential decisions is infeasible, unethical, or expensive. Given a batch of data collected under a tested policy, OPE is used to estimate the utility of taking untested counterfactual decisions. As described in \citet{Parbhoo2022} and \citet{pmlr-v97-oberst19a}, these estimates may be prospective i.e. the future outcome of following the counterfactual policy from a particular state, or retrospective i.e. how the outcome of the tested policy seen in the data would have been affected by counterfactual decisions.

Performing OPE in real-world applications is challenging; the decisions we wish to evaluate may differ from those observed in the batch of data available for evaluation, resulting in poor sample overlap. This causes high variance in the estimates of a policy's performance and low data sample efficiency i.e. more samples are needed to guarantee a particular level of variance. Meanwhile, combinatorial action spaces are common for many applications: e.g., in healthcare, an action may correspond to combinations of different drugs or treatments \citep{parbhoo2017combining,komorowski2018artificial,prasad2017reinforcement}. For these problems, viewing each combination of ``sub-actions'' as a unique action results in an exponentially large action space, where only a few actions correspond to those in our batch data, thereby exacerbating the large variance problem in OPE estimates. 

In this paper, we develop a new approach for OPE that leverages the idea of factored action spaces \citep{tang2022leveraging}. The intuition is that though it may be difficult to ensure overlap between the policy we wish to evaluate and the data-generating policy, each of them may instead be decomposed into factors over actions, making it easier to ensure overlap among the policies with similar factors. In doing so, we improve sample efficiency and obtain OPE estimates with better bias and variance guarantees. This approach can also facilitate a more detailed analysis of the effects of counterfactual decisions, as they can be observed at the level of action factors. Our contributions are as follows: 
\begin{itemize}[itemsep=-2pt, topsep=-2pt]
    \item We introduce a family of decomposed IS estimators that leverage factored action spaces for performing off-policy evaluation (OPE).
    \item We prove theoretically that under certain assumptions, our decomposed estimators are unbiased and have lower variance than standard OPE estimators.
    \item We demonstrate empirically that decomposed OPE estimators have lower variance and similar bias to existing OPE baselines, while having better data sample efficiency, as measured by effective sample size (ESS).
\end{itemize}

\section{Related Work} \label{sec:related-work}
\paragraph{Approaches to Off-Policy Evaluation.}
Standard methods for OPE struggle when dealing with combinatorial action spaces. Direct methods (DM) use a model of the environment to simulate trajectories to compute the value of a policy \citep{paduraru2013off,chow2015robust,hanna2017bootstrapping,fonteneau2013batch,liu2018representation}; learning an accurate model of combinatorial actions from offline data is difficult because some combinations may not be observed. In general, DM can suffer from large bias where the actions chosen by the evaluation policy differ significantly from the behaviour policy. Importance sampling (IS) estimators i.e. inverse propensity score (IPS)-based estimators use reweighting to correct the sampling bias in off-policy data, such that an unbiased estimator may be obtained (e.g. \citet{precup2000,horvitz1952generalization,thomas2016data}). Unfortunately, as the size of the action space increases, these methods suffer from poor sample overlap, resulting in value estimates with high variance. Doubly robust (DR) estimators (e.g. \citet{jiang2016doubly,farajtabar2018more}) combine DM and IS; these can be low variance if either DM or IS is accurate. DR estimators can be used along with our approach; however, \emph{they do not offer a way to explicitly factorise action spaces} to improve overlap as we propose here. \citet{keramati2021identification} analyse how identifying \textit{subgroups} with similar benefits when performing OPE can produce more reliable estimates. In contrast, we focus on explicitly \emph{factorising combinatorial action spaces} to reduce variance of the estimate.

\paragraph{Factored Actions in RL.}
\citet{tang2022leveraging} proposed a linear function decomposition to express the Q-function based on factored action spaces. The authors provide several theoretical guarantees of when the approach can lead to unbiased and low-variance \emph{on-policy} estimation of the Q-function i.e. where the policy generating the data is being evaluated. They illustrated the bias-variance trade-off in offline RL. Other work, such as \citet{tavakoli2020learning,sunehag2017value,zhou2019factorized} leveraged action factorisation for improved exploration, handling multiple agents or multiple rewards, primarily in online RL. 

We share a similar premise with these prior works in assuming the problem comes with factored action spaces and we utilise the theory from \citet{tang2022leveraging} as a starting point. However, unlike these works which address \textit{policy learning}, we explicitly focus on the task of \emph{off-policy evaluation}, where we are given offline data under a particular behaviour policy but would like to evaluate the performance following a counterfactual sequence of actions generated by a different policy. The OPE setting brings about new challenges and complexities in the theoretical guarantees and assumptions (e.g. conditions for unbiasedness and variance reduction), which are the focus of this work. 

\section{Preliminaries} \label{sec:problem-setup}
\paragraph{Markov Decision Processes.}
We consider Markov decision processes (MDPs) defined by a tuple $\mathcal{M} = (\mathcal{S}, \mathcal{A}, p, r, d_1, \gamma, T)$, where $\mathcal{S}$ and $\mathcal{A}$ are the state and action spaces, $p: \mathcal{S} \times \mathcal{A} \to \Delta(\mathcal{S})$ and $r: \mathcal{S} \times \mathcal{A} \to \Delta(\mathbb{R})$ are the transition and reward functions, $d_1 \in \Delta(\mathcal{S})$ is the initial state distribution, $\gamma \in [0,1]$ is the discount factor, $T \in \mathbb{Z}^{+}$ is the fixed horizon. A policy $\pi: \mathcal{S} \to \Delta(\mathcal{A})$ specifies a mapping from each state to a probability distribution over actions. A $T$-step trajectory following policy $\pi$ is denoted by $\tau = \smash{[(s_t, a_{t+1}, r_t, s_{t+1})]}_{t=1}^{T}$ where $s_0 \sim d_1, a_{t+1} \sim \pi(s_t), r_t \sim r(s_t,a_{t+1}), s_{t+1} \sim p(s_t, a_{t+1})$. Here, $a \sim \pi(s)$ is short for $a \sim \pi(\cdot | s)$ and $s' \sim p(s, a)$ for $s' \sim p(\cdot|s, a)$. Let $J = \sum_{t=1}^{T} \gamma^{t-1} r_{t}$ denote the return of the trajectory, which is the discounted sum of rewards. The value a of policy $\pi$, denoted by $V_{\pi}: \mathcal{S} \to \mathbb{R}$, maps each state to the expected return starting from that state following policy $\pi$. That is, $V_\pi(s) = \mathbb{E}_{\pi}[J|s_1 = s]$. Similarly, the action-value function (i.e., the Q-function), $Q_{\pi}: \mathcal{S} \times \mathcal{A} \to \mathbb{R}$, is defined by further restricting the action taken from the starting state $Q_\pi(s,a) = \mathbb{E}_\pi[J|s_1 = s, a_1 = a]$. The performance of a policy (i.e., the value of policy $\pi$) is defined as the expected value over initial states, $V_{\pi} = \mathbb{E}_{s \sim d_1}[V_{\pi}(s)]$.

\paragraph{Off-Policy Evaluation.}
In OPE, we are given a dataset of $T$-step trajectories $\mathcal{D}=\{\tau^{(n)}\}_{n=1}^{N}$ each independently generated by some \emph{behaviour policy} $\pi_b$. We aim to produce a good estimate of $V_{\pi_e}$ or $Q_{\pi_e}$, the performance of a different policy, $\pi_e$, known as the \emph{evaluation policy}. In general, the estimator $\hat{V}_{\pi_e}$ or $\hat{Q}_{\pi_e}$ is good if it achieves low mean squared error (MSE), 
\begin{equation}
\label{eq:MSE-def-1}
MSE(Q_{\pi_e}, \hat{Q}_{\pi_e}) = \mathbb{E}_{P^\tau_{\pi_b}}[(Q_{\pi_e} - \hat{Q}_{\pi_e})^2],
\end{equation}
where $P^\tau_{\pi_b}$ denotes the distribution of trajectory $\tau$ under behaviour policy $\pi_b$. 

\paragraph{Factored Action Spaces in MDPs.}
While the standard MDP definition does not consider the underlying structure within action space $\mathcal{A}$, we follow \citet{tang2022leveraging} and explicitly express a factored action space $\mathcal{A}$ as a Cartesian product of $D$ sub-action spaces $\mathcal{A}^d$, with $d \ \in \ \{1, \cdots, D\}$. Formally, $\mathcal{A} = \otimes_{d=1}^{D} \mathcal{A}^d$. Accordingly, each action can be written as a vector of sub-actions, $a = [a^1, \cdots, a^D]$. The key insight of \citet{tang2022leveraging} is that the Q-function (of certain policies) can be additively decomposed in terms of the factored action spaces: 
\begin{equation}
    \label{eq:factored-Q-decomposition}
    \tilde{Q}_{\pi}(s,a) = \sum_{d=0}^{D} \hat{Q}^{d}_{\pi}(s,a^d)
\end{equation}
Sufficient conditions (on the MDP and policy $\pi$) for Equation \ref{eq:factored-Q-decomposition} to be valid are outlined in Theorem \ref{theorem-1-shengpu paper}. It should be noted that while not guaranteed, Equation \ref{eq:factored-Q-decomposition} can hold even when the sufficient conditions are violated. 

\begin{theorem}[adapted from Theorem 1 of \citet{tang2022leveraging}]
\label{theorem-1-shengpu paper}
Equation \ref{eq:factored-Q-decomposition} holds if, for the MDP, the following conditions hold: 
 \begin{equation}
 \label{eq:transition-factorisation}
     \sum_{\tilde{s} \ \in \ \boldsymbol{\phi}^{-1}(\boldsymbol{\phi}(s'))} p(\tilde{s}|s,a) = \prod_{d=1}^{D} p^d((z')^{d}|z^d,a^d)
 \end{equation}
 \begin{equation}
 \label{eq:reward-factorisation}
     r(s,a) = \sum_{d=1}^{D} r^d(z^d, a^d), 
 \end{equation}
where $p^d$ and $r^d$ are sub-transition and sub-reward functions corresponding to sub-action space $\mathcal{A}^d$, and $\boldsymbol{\phi}(s) = [z^1, \cdots, z^D]$, such that each $z^d$ is an abstraction of $s$ with respect to sub-action space $\mathcal{A}^d$. Additionally, for policy $\pi$, the following should hold:
\begin{equation}
\label{eq:policy-factorisation}
     \pi(a|s) = \prod_{d=1}^{D} \pi^d(a^d|z^d) 
 \end{equation}
where each $\pi^d$ is a sub-policy corresponding to sub-action space $\mathcal{A}^d$.
\end{theorem}

Below, we state the assumptions that will be used in our theoretical analyses. The first regularity assumption is a relaxed version of the standard regularity assumption in OPE, where absolute continuity is only required and assumed over factors of the action space rather than the entire action space. The second assumption states that our policies do not change with time. 
\begin{assumption}(Absolute Continuity).
\label{abscont}
For all $(s, a^d) \in \mathcal{S} \times \mathcal{A}^d$, if $\pi_b(a^d|s) = 0$ then $\pi_e(a^d|s) = 0$. 
\end{assumption}


\begin{assumption}(All Policies are Stationary).
\label{ass:stationarity}
The probability distributions of a policy do not change with time within the trajectory or the number of trajectories completed.
\end{assumption}

Additionally, to prove bounds on the variance of the decomposed estimators, we make the following assumptions:

\begin{assumption}(Conditions for Variance Bounds on Decomposed Estimators).
\label{ass:variance-bounds}
The following conditions hold $\forall \ d, d', t, t', n$, where $d' \ne d$ and $t' \ne t$:
    \begin{equation}
        \label{eq:reward-correlation}
        Cov\left(r^{d}(z_{t}^{(n),d},a_{t}^{(n),d}), \  r^{d^{\prime}}(z_{t^{\prime}}^{(n),d^{\prime}},a_{t^{\prime}}^{(n),d^{\prime}})\right) \ge 0
    \end{equation}
    \begin{equation}
        \label{eq:ratio-correlation}
        Cov\left((\rho_{0:T}^{(n), d}), \ (\rho_{0:T}^{(n), d^{\prime}})\right) = 0
    \end{equation}
    \begin{equation}
        \label{eq:reward-ratio-correlation}
        Cov\left((r^d(z_{t}^{(n),d},a_{t}^{(n),d}), \ (\rho_{0:T}^{(n), d^{\prime}})\right) = 0
    \end{equation}

Here, $\rho_{0:T}^{(n),d} = \prod_{t^{\prime}=0}^{T} \frac{\pi^d_e(a_{t^{\prime}}^{(n),d} | z_{t^{\prime}}^{(n), d})}{\pi^d_b(a_{t^{\prime}}^{(n), d} | z_{t^{\prime}}^{(n), d})}$ is the importance sampling weight specifically for the abstracted trajectory $[(z^{(n), d}_t, a^{(n), d}_t, z^{(n), d}_{t+1})]_{t=1}^{T}$ corresponding to $\mathcal{A}^d$. $Cov(X,Y)$ denotes the covariance between random variables $X$ and $Y$.

\end{assumption}

These conditions require that the rewards at different times $t$ and IS ratios in different factored action spaces $d$ are all uncorrelated. The rewards at different $t$ and $d$ can be positively correlated i.e. the rewards increase together. Finally, an extra assumption is needed to prove the variance bound of one of the decomposed estimators (decomposed PDWIS):

\begin{assumption} (Additional Variance Bound Condition)
\label{ass:PDWIS-variance-bounds}
$\forall \ d, t, t', n$, where $t' \ne t$:
    \begin{equation}
        \label{eq:same-ratio-correlation}
        Cov \left(  \rho_{0:t}^{(n), d}, \  \rho_{0:t^{\prime}}^{(n), d} \right) \ge 0
    \end{equation}

\end{assumption}

This assumption imposes that importance sampling weights in any given factored action space and episode should be non-negatively correlated when calculated on sub-trajectories of different lengths.

\section{Method} \label{sec:method}
We seek to leverage factored action spaces to improve the bias and variance guarantees of OPE estimators; in this work, we focus on the IS-based estimators. To do this, we utilise our knowledge that the action space can be expressed in terms of smaller, independent sub-action spaces. The overall intuition of our approach is that while it may be difficult to ensure overlap between behaviour and evaluation policies, by decomposing both behaviour and evaluation policies into factors over actions, we may be able to ensure overlap among the policies with similar factors. In doing so, we may be able to improve sample efficiency and obtain OPE estimates with better bias and variance guarantees.

To begin, we impose that the conditions in \cref{theorem-1-shengpu paper} hold, i.e., \cref{eq:factored-Q-decomposition,eq:transition-factorisation,eq:reward-factorisation,eq:policy-factorisation} hold. Applying these equations allows us to derive new versions of the IS estimators defined in \citet{precup2000}. These new estimators leverage the factorisation structure of the action spaces, and are presented in \cref{def:decomposed-IS-PDIS,def:decomposed-PDWIS} below.

\begin{definition} (Decomposed IS and Decomposed Per-Decision IS) The decomposed IS estimator is defined as
\label{def:decomposed-IS-PDIS}
\begin{equation}
    \label{eq:decomposed-IS}
    \tilde{Q}^{DecIS}_{\pi_e} = \sum_{d=0}^{D} \frac{1}{N} \sum_{n=1}^{N} \sum_{t=0}^{T} \gamma^t \cdot  \rho_{0:T}^{(n),d} \cdot r^d(z_t^{(n), d}, a_{t}^{(n), d})
\end{equation}

where $\rho_{0:T}^{(n),d}$ is defined in Assumption \ref{ass:variance-bounds}. The decomposed Per-Decision IS (PDIS) estimator $\tilde{Q}^{DecPDIS}_{\pi_e}$ is defined by replacing $\rho_{0:T}^{(n),d}$ with $\rho_{0:t}^{(n),d}$ i.e. IS weights up to each time step $t$.

By \cref{eq:factored-Q-decomposition}, we can calculate an IS estimate for each $d$ i.e. an estimate factor and take the sum. Utilising \cref{eq:policy-factorisation}, we define the IS weights $\rho^{d}$ for each estimate factor. To calculate the estimate factor for each $d$, we utilise $r^d$ based on \cref{eq:reward-factorisation}. 
\end{definition}

Similarly, we can also define weighted variants of our new decomposed estimator:

\begin{definition}(Decomposed PDWIS)
\label{def:decomposed-PDWIS}
The decomposed Per-Decision Weighted IS estimator based on factored action spaces is given by, where $r^{(n),d}_t \equiv r^d(z_t^{(n), d}, a_{t}^{(n), d})$,
\begin{equation}
    \label{eq:decomposed-PDWIS}
    \hat{Q}_{\pi_e}^{DecPDWIS} = \sum_{d=0}^{D} \frac{\sum_{n=1}^{N} \sum_{t=0}^{T} \gamma^t \cdot \rho_{0:t}^{(n),d} \cdot r^{(n),d}_t}{\sum_{n=1}^{N} \sum_{t=0}^{T} \gamma^t \cdot \rho_{0:t}^{(n),d}}
\end{equation}
The PDWIS estimator is derived from PDIS by dividing the IS weighted discounted reward sum by the sum of the IS weights instead of $N$. We note that the decomposed PDIS estimator is a sum of PDIS estimators, one for each $d$. We convert these PDIS estimators to PDWIS as discussed. The sum of these sub-PDWIS estimators gives $\hat{Q}_{\pi_e}^{DecPDWIS}$. 
\end{definition}
We next utilise \cref{eq:factored-Q-decomposition,eq:transition-factorisation,eq:reward-factorisation,eq:policy-factorisation} to derive theoretical guarantees on the bias and variance of these decomposed estimators, set out in \cref{theorem-unbiased-factored-estimator,theorem-lower-variance-factored-estimator}.

\begin{theorem}
\label{theorem-unbiased-factored-estimator}
    When the assumptions in Theorem \ref{theorem-1-shengpu paper} hold, the decomposed IS estimator $\hat{Q}_{\pi_e}^{DecIS}$ and decomposed PDIS estimator $\hat{Q}_{\pi_e}^{DecPDIS}$ are unbiased estimators of the true Q-function $Q_{\pi_e}$. 
\end{theorem}
\textit{Proof Sketch.} The decomposed IS estimator in \cref{eq:decomposed-IS} is a sum of IS estimators, one for each Q-function factor $Q_{\pi_e}^d$ for the factored MDP and factored policy $\pi_e^d$ corresponding to $\mathcal{A}^d$. Each of these IS estimators is unbiased, as this is a property of an IS estimator \citep{precup2000}. Since \cref{eq:factored-Q-decomposition} holds, we can sum these unbiased estimates to get an unbiased estimate of the overall Q-function. A similar argument can be applied for decomposed PDIS. The full proof is provided in \cref{sec:dec-bias}.

\begin{theorem}
\label{theorem-lower-variance-factored-estimator}
    The decomposed IS and PDIS estimators are guaranteed to have at most the variance of their respective non-decomposed equivalent estimators i.e. $\mathbb{V}_{\pi_b}[\hat{Q}_{\pi_e}^{DecIS}] \le \mathbb{V}_{\pi_b}[\hat{Q}_{\pi_e}^{IS}]$ and $\mathbb{V}_{\pi_b}[\hat{Q}_{\pi_e}^{DecPDIS}] \le \mathbb{V}_{\pi_b}[\hat{Q}_{\pi_e}^{PDIS}]$, provided that Theorem \ref{theorem-1-shengpu paper} and the conditions in Assumption \ref{ass:variance-bounds} hold. For $\mathbb{V}_{\pi_b}[\hat{Q}_{\pi_e}^{DecPDWIS}] \le \mathbb{V}_{\pi_b}[\hat{Q}_{\pi_e}^{PWDIS}]$, Assumption \ref{ass:PDWIS-variance-bounds} must hold in addition to the mentioned conditions.
\end{theorem}

\textit{Proof Sketch.} Given the variance expressions of IS, PDIS and PDWIS estimators, and their decomposed versions, we use \cref{eq:policy-factorisation,eq:reward-factorisation} to write all of them in terms of the sub-actions $a^d$ and sub-rewards $r^d$. Between these expressions, we identify comparable corresponding terms and we utilise guarantees from these comparisons to compare the overall variance expressions. The full proof is in \cref{sec:dec-var}.

When \cref{ass:variance-bounds} does not hold, there is no guarantee on the variance of the decomposed estimators, as interactions between $r_t^d$ and $\rho^d$ across $t$ and $d$ can in some cases lead to the variance of the decomposed estimator being more than its non-decomposed counterpart. It should also be noted that there is no guarantee that the variance of a decomposed estimate will always scale at a slower rate than its non-decomposed counterpart with length of the trajectory or mismatch between policies. 

An important result of \cref{theorem-lower-variance-factored-estimator} relates to the effective sample size (ESS) of an IS (or PDIS or PDWIS) estimator $\hat{Q}^{IS}_{\pi_e}$, which is a measure of data sample efficiency:
\begin{equation}
\label{eq:ESS-definition}
    ESS[\hat{Q}^{IS}_{\pi_e}] = N \times \frac{\mathbb{V}_{\pi_e}[\hat{Q}^{\ on \ policy}_{\pi_e}]}{\mathbb{V}_{\pi_b}[\hat{Q}^{IS}_{\pi_e}]}
\end{equation}
where both variances are calculated on the same number of samples and $\hat{Q}^{\ on \ policy}_{\pi_e}$ is a Q-function estimate based on data generated by $\pi_e$. $\mathbb{V}_{\pi_e}[\hat{Q}^{\ on \ policy}_{\pi_e}]$ stays the same for different IS estimators, hence the ESS is inversely proportional to $\mathbb{V}_{\pi_b}[\hat{Q}^{IS}_{\pi_e}]$. Clearly then, \cref{theorem-lower-variance-factored-estimator} states that a decomposed estimator has a higher ESS i.e. it is more efficient than its non-decomposed version.

When the sufficient conditions hold, factored action spaces can reduce the variance of IS estimators without increasing bias. A qualitative explanation is that each factored action space has fewer possible actions than the overall action space. In each factored space, the distributions $\pi_b$ and $\pi_e$ are likely to overlap more i.e. greater coverage. \cref{eq:factored-Q-decomposition} also means the Q-function can be calculated for each factored action space and then summed to give an unbiased estimate of $Q_{\pi_e}$. Thus, variance is reduced without affecting bias.

When the conditions in \cref{theorem-1-shengpu paper} are relaxed, \cref{eq:factored-Q-decomposition} may not hold, causing the decomposed estimators to be biased or to have higher variance. One way to address this is to re-group the factored action spaces into larger factored action spaces that satisfy \cref{theorem-1-shengpu paper}. For example, if we grouped factored spaces as $\mathcal{A}^{d_{\{1,2,3\}}} = \mathcal{A}^{d_1} \cup \mathcal{A}^{d_2} \cup \mathcal{A}^{d_3}$, we can define the reward factor $r^{d_{\{1, 2, 3\}}}(z^{d_1},z^{d_2},z^{d_3},a^{d_1},a^{d_2},a^{d_3})$ and IS ratio $\rho_{0:T}^{d_{\{1, 2, 3\}}}$ based on policy factors $\pi_b^{d_{\{1, 2, 3\}}}(a^{d_1},a^{d_2},a^{d_3} | z^{d_1},z^{d_2},z^{d_3})$ and $\pi_e^{d_{\{1, 2, 3\}}}(a^{d_1},a^{d_2},a^{d_3} | z^{d_1},z^{d_2},z^{d_3})$. Thus, we can re-obtain the guarantees on bias and variance, at the cost of decreasing overlap between $\pi_b$ and $\pi_e$. An illustration is in \cref{sec:grouping-action-spaces}. If all the factored action spaces were grouped together, we would get the original, non-decomposed IS estimator operating in the full action space $\mathcal{A}$. 

\section{Experimental Setup}
We empirically validated \cref{theorem-unbiased-factored-estimator,theorem-lower-variance-factored-estimator} through code simulations of two MDP's specified below. Simultaneously, we investigated how varying $N$, $T$, divergence between $\pi_b$ and $\pi_e$, and relaxing \cref{theorem-1-shengpu paper} affected the relative performance of the decomposed and non-decomposed estimators. The code is publicly available at \url{https://github.com/ai4ai-lab/Factored-Action-Spaces-for-OPE}.

\paragraph{MDP-1.} This MDP is the 2-dimensional bandit problem from \citet{tang2022leveraging}. A diagram of the MDP is provided in \cref{sec:diag-MDP-1}. $\mathcal{S} = \{state, terminal\}$, $\mathcal{A} = \{up\_right, \ up\_left, \ down\_right, \ down\_left\}$. Every action taken from $state$ or $terminal$ always leads to $terminal$. In terms of rewards, $r(state, \ up\_right) = 1 + \alpha + \beta$,  $r(state, \ up\_left) = \alpha$, $r(state, \ down\_right) = 1$, where $\alpha = 1$ and $\beta = 0$. Otherwise, $r = 0$. The start state is always $state$. $T=1$ always, hence $\gamma$ is irrelevant. The MDP satisfies \cref{theorem-1-shengpu paper} for $\beta=0$ and can be factored into two sub-action spaces as described in \cref{sec:diag-MDP-1}.

\paragraph{MDP-2.} This 4-state MDP is inspired by Figure 2(a) from \citet{tang2022leveraging}. A diagram is provided in \cref{sec:diag-MDP-2}. $\mathcal{S} = \{0,0 \ , \ 0,1 \ , \ 1,0 \ , \ 1,1\}$, $\mathcal{A} = \{up\_right, \ up\_left, \ down\_right, \ down\_left\}$. $P$ is defined such that always, $up\_right$ leads to $1,1$, $up,left$ to $0,1$, $down\_right$ to $1,0$ and $down\_left$ to $0,0$. The reward structure is complex and described in \cref{sec:diag-MDP-2}. The start state is always $0,0$. Both $T$ and $\gamma$ are varied, as described in the training details. The MDP satsfies \cref{theorem-1-shengpu paper} and factors into two action spaces as described in \cref{sec:diag-MDP-2}.

Each MDP has been constructed by combining two smaller MDP's with independent dynamics. This suggests that the rewards and IS ratios in different factored action spaces would be uncorrelated, thus satisfying \cref{ass:variance-bounds}. It is also expected that \cref{ass:PDWIS-variance-bounds} is satisfied.

\paragraph{Metrics.}
The following metrics characterise experimental configurations and performance of the OPE estimators:

\textit{Bias}: The bias of OPE estimator $\hat{Q}_{\pi_e}$ is the difference between the expected value of the estimator over all possible observed datasets generated by $\pi_b$ and the true value $Q_{\pi_e}$. It measures how accurate the estimator is on average.
\begin{equation}
    \label{eq:bias-definition}
    Bias(\hat{Q}_{\pi_e}, Q_{\pi_e}) = \mathbb{E}_{\pi_b}[\hat{Q}_{\pi_e}] - Q_{\pi_e} 
\end{equation}
\textit{Variance}: The variance of estimator $\hat{Q}_{\pi_e}$ is given in \cref{eq:variance-definition}. It measures how precise the estimator is.
\begin{equation}
    \label{eq:variance-definition}
    \mathbb{V}_{\pi_b}[\hat{Q}_{\pi_e}] = \mathbb{E}_{\pi_b}[(\hat{Q}_{\pi_e})^2] - \mathbb{E}_{\pi_b}[\hat{Q}_{\pi_e}]^2
\end{equation}
\textit{Mean Squared Error (MSE)}: In addition to \cref{eq:MSE-def-1}, MSE can be calculated via the expression below, which combines bias and variance into a single metric measuring the accuracy of the OPE estimates:
\begin{equation}
\label{eq:MSE-def-2}
MSE(Q_{\pi_e}, \hat{Q}_{\pi_e}) = Bias(Q_{\pi_e}, \hat{Q}_{\pi_e})^2 + \mathbb{V}[\hat{Q}_{\pi_e}]
\end{equation}
\textit{Effective Sample Size (ESS)}: This is defined in \cref{eq:ESS-definition}, assuming that $N$ data samples are taken from $\pi_b$ for the off-policy estimator to make its estimate. The ESS reflects the equivalent number of \textit{on-policy} samples required from $\pi_e$ to obtain the same variance. The higher the ESS, the more sample efficient the estimator is. 

\textit{Policy Divergence}: This metric was defined by \citet{VoloshinEtAl2019} to quantify the difference between $\pi_b$ and $\pi_e$ in the context of an OPE problem:
\begin{equation}
    \label{eq:policy-divergence-definition}
    PD(\pi_b, \pi_e) = \Big(\sup_{a \in \mathcal{A}, s \in \mathcal{S}} \frac{\pi_e(a|s)}{\pi_b(a|s)}\Big)^T
\end{equation}
where $T$ is the length of each trajectory. The minimum $PD(\pi_b, \pi_e)$ of $1$ indicates $\pi_b = \pi_e$, while $PD(\pi_b, \pi_e)$ increases as the overlap between $\pi_b$ and $\pi_e$ decreases.

\paragraph{Training Details.}

The following OPE estimators were compared: IS, PDIS, PDWIS, decomposed IS (DecIS), decomposed PDIS (DecPDIS) and decomposed PDWIS (DecPDWIS). Additionally, an on-policy estimate of the Q-function was recorded in each experiment. 

For each MDP, we generated two datasets: $\mathcal{D}_{\pi_e}$ following $\pi_e$, and $\mathcal{D}_{\pi_b}$ following $\pi_b$. The $\mathcal{D}_{\pi_e}$ was used to calculate the \textit{on-policy} estimates, which represent the true value that the OPE estimators are estimating. $\mathcal{D}_{\pi_b}$ was used by the OPE estimators. To allow experimentation with a wide range of $N$ and (for MDP-2) $T$, $\mathcal{D}_{\pi_e}$ and $\mathcal{D}_{\pi_b}$ each consisted of $10,000,000$ trajectories of length $1$ for MDP-1, and $100,000$ trajectories of length $1000$ for MDP-2. For values of $N$ and $T$ smaller than these sizes, multiple subsets of $\mathcal{D}_{\pi_b}$ and $\mathcal{D}_{\pi_e}$ could be taken. Whenever the policies and MDP rewards were varied, new datasets had to be generated.

To empirically measure bias, variance, and MSE, we generated $\mathrm{R} = 100$ subsets of $\mathcal{D}_{\pi_b}$ and $\mathcal{D}_{\pi_e}$ for $\mathrm{R}$ estimates from each estimator in each experiment. From these, the MSE was calculated for each OPE estimator using the definition in \cref{eq:MSE-def-1}, where the true value was the on-policy estimate in the experiment. The variance of the $\mathrm{R}$ estimates was also calculated for each estimator; this allowed us to find the bias by applying \cref{eq:MSE-def-2}. To cross-check, the bias was also calculated with by applying \cref{eq:bias-definition}, using the on-policy estimate as the true value. The bias of the on-policy estimator was always taken to be zero, hence its MSE is equal to its variance. The ESS of each estimator was calculated by \cref{eq:ESS-definition} using the number of trajectories $N$, the variance of $\mathrm{R}$ OPE estimates and the variance of $\mathrm{R}$ on-policy estimates.

To generate each of our plots, we repeated the procedure in the above paragraph with five different $(\mathcal{D}_{\pi_e}, \ \mathcal{D}_{\pi_b})$ pairs. All plotted values (denoted by '$\times$' in the figures) are mean values over these five trials. Our plots also display error bars at each plotted value; the one-sided size of each error bar is equal to the standard deviation over the five trials.

The tested values of each parameter in the experiments were: $N \in \{10, \ 50, \ 100, \ ... \ 100000\}$, $T \in \{1, \ 5, \ 10, \ 50,  \ 100,  \ 500, \ 1000\}$, $\gamma \in \{0.7, \ 0.9, \ 0.9999\}$ and $PD(\pi_b, \ \pi_e) \in \{1.44^T, \ 2.56^T, \ 3.61^T, \ 4.46^T, \ 5.64^T, \\ 10.03^T, \ 22.53^T, \ 90.25^T, \ 361.0^T \}$ where we note that all the policy divergence values are raised to the value of $T$.

\section{Results}
\paragraph{Empirical Finding 1.} \textit{The decomposed estimators have lower variance and MSE than their corresponding non-decomposed versions for all tested values of $N$}.

\begin{figure}[h!]
    \centering
    \vspace{0.2cm}
    \includegraphics[scale=0.45]{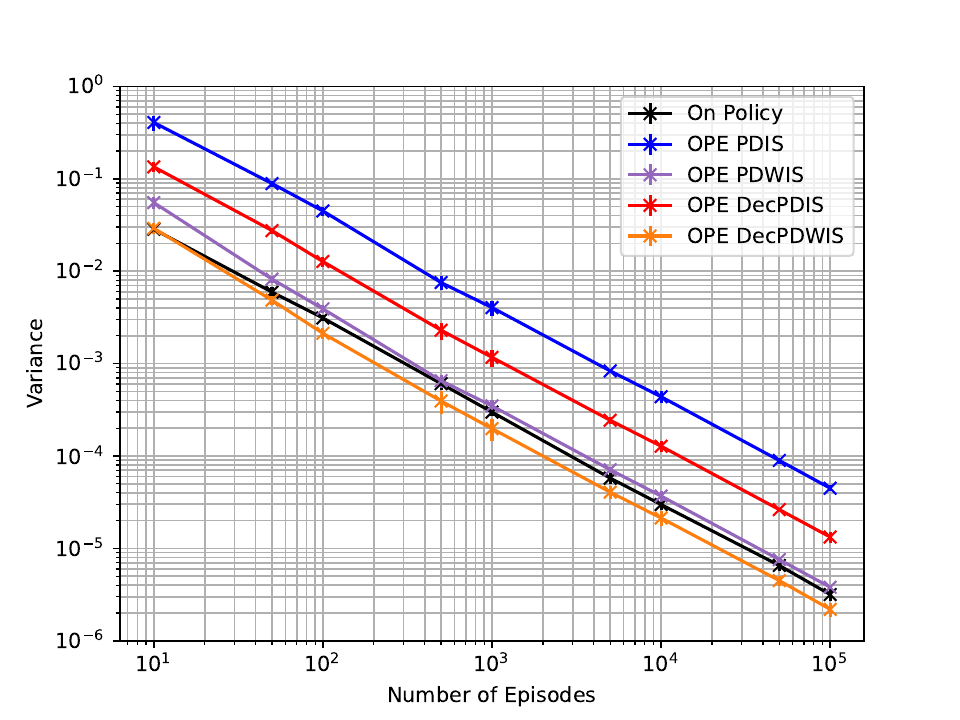}
    \includegraphics[scale=0.45]{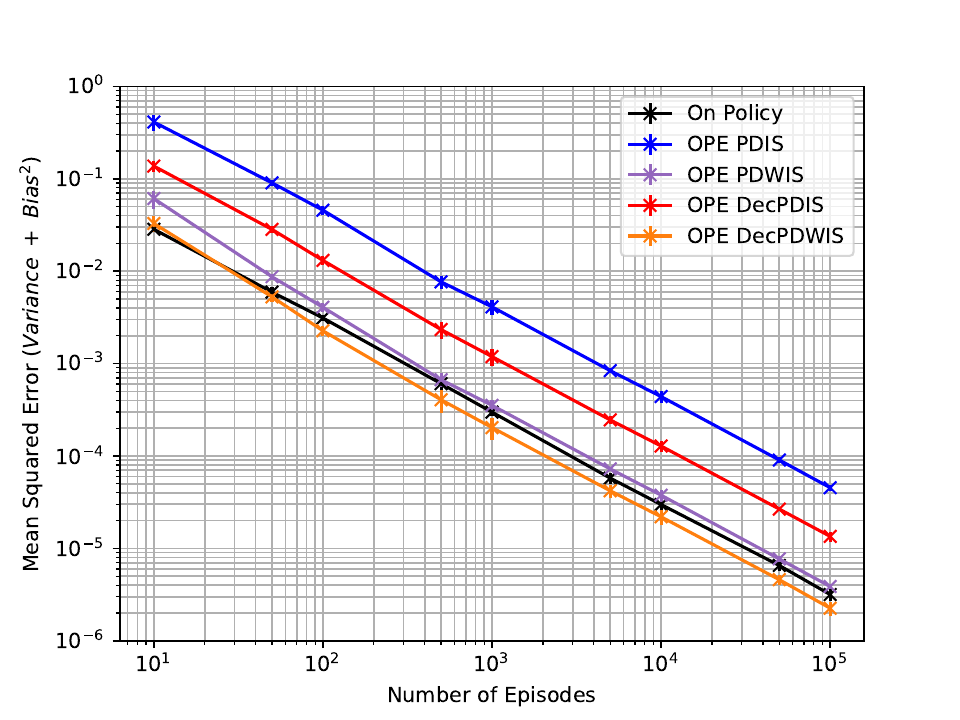}
    \caption{Variance (top) and MSE (bottom) of different OPE estimators as the number of episodes ($N$) varies, for MDP-1 with $PD(\pi_b, \pi_e) = 2.56$. The variances and MSEs of all estimators decrease with $N$, while maintaining a clear and consistent ordering among estimators.}
    \label{fig:var-vs-trajectories-policy-divergence-2-56}
\end{figure}

In \cref{fig:var-vs-trajectories-policy-divergence-2-56}, for all tested values of $N$, the decomposed estimators (DecPDIS, DecPDWIS) have less variance than their non-decomposed counterparts (PDIS, PDWIS, respectively). Since MDP-1 has only one transition, IS is equivalent to PDIS and is thus not shown. The variances of all estimators scale similarly with increasing $N$. Compared to the variance of PDIS, using the weighted version (as in PDWIS) offers a greater variance reduction than leveraging factored action spaces (as in DecPDIS), while applying both ideas (as in DecPDWIS) achieves the smallest variance, implying that these are two complementary sources of variance reduction. Finally, since all estimators have negligible bias, the MSE plot shows the same trend as the variance.

\paragraph{Empirical Finding 2.} \textit{The decomposed IS and PDIS estimators have approximately zero bias as long as \cref{theorem-1-shengpu paper} holds and there is sufficient coverage of $\pi_e$ by $\pi_b$}.

\begin{figure}[h!]
    \centering
    \vspace{-0.2cm}
    \includegraphics[scale=0.45]{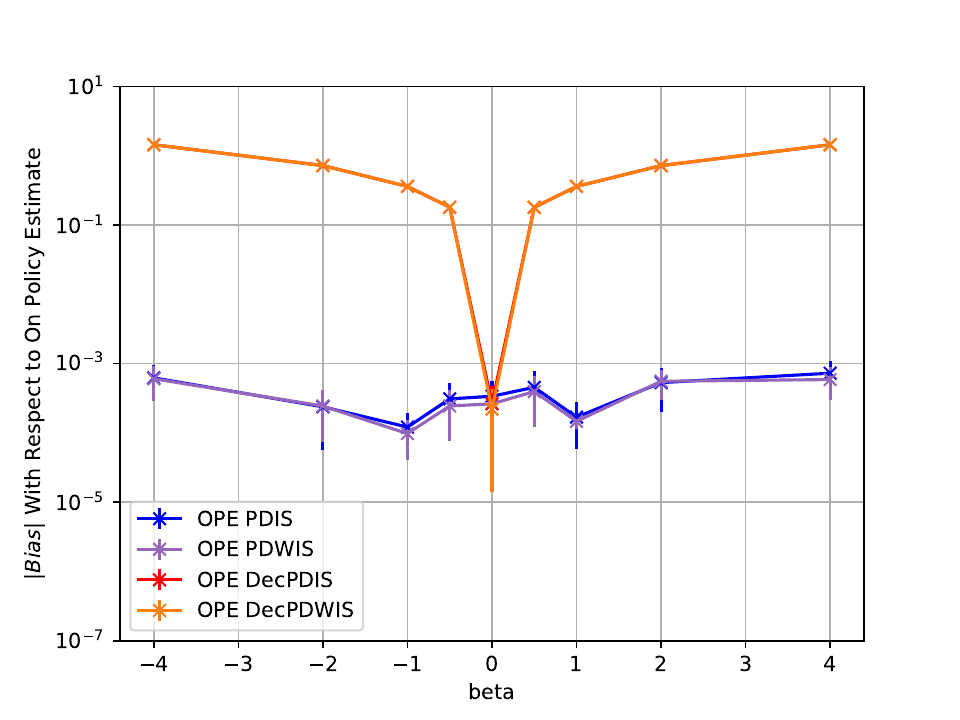}
    \vspace{-0.2cm}
    \caption{The magnitude of bias of different OPE estimators as $\beta$ varies, for MDP-1 with $N = 100,000$, $PD(\pi_b, \pi_e) = 1.44$. While non-decomposed estimators have near-zero bias for all tested values of $\beta$ as expected, decomposed estimators only have near-zero bias for $\beta=0$ and increasing bias as $|\beta|$ increases.}
    \label{fig:bias-100000-trajectories-vs-beta}
\end{figure}

In the definition of MDP-1, $\beta$ is part of the reward for taking action $up,right$ from $state$. When $\beta = 0$, it is possible to satisfy the condition in \cref{eq:reward-factorisation}; and thus, satisfy \cref{theorem-1-shengpu paper}. When $\beta \ne 0$, this condition cannot be satisfied; the more we increase $|\beta|$, the more strongly the condition is violated. In \cref{fig:bias-100000-trajectories-vs-beta}, we compare the bias of the estimators for varying values of $\beta$. For $\beta=0$, the bias of the decomposed estimators is comparable with the non-decomposed estimators i.e. $Bias(\hat{Q}_{\pi_e}, Q_{\pi_e}) \approx 0$. For $\beta \ne 0$, since the decomposed estimators implicitly assume $\beta = 0$ in their definitions, their biases increase rapidly as $|\beta|$ increases.

\begin{figure}[h!]
    \centering
    \includegraphics[scale=0.525]{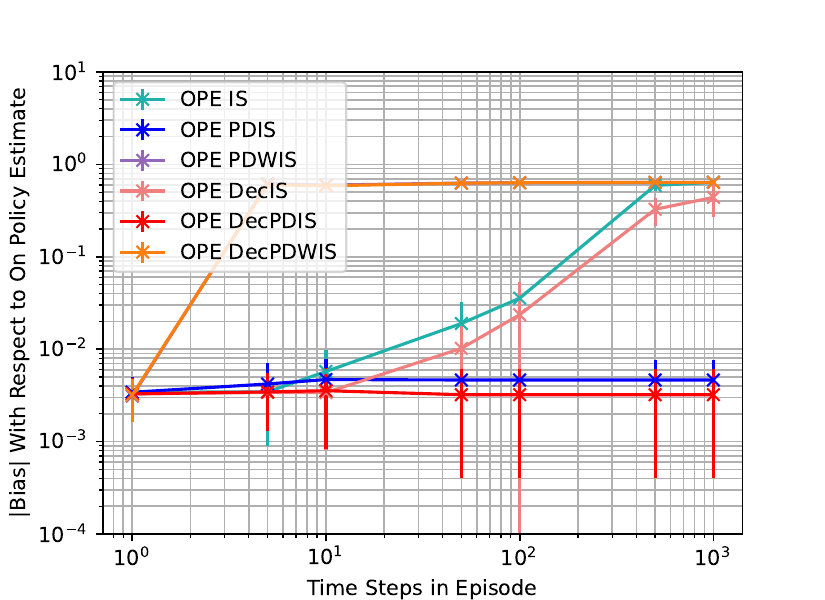}
     \vspace{-0.2cm}
    \caption{The magnitude of bias of different OPE estimators as the episode length ($T$) varies, for MDP-2 with $\gamma = 0.7$, $N=1000$ and $PD(\pi_b, \pi_e) = 1.44^T$. Typically, the IS and PDIS estimators should always have near-zero bias, however as $T$ increases, the divergence between $\pi_b$ and $\pi_e$ increases, resulting in loss of coverage of $\pi_e$ by samples from $\pi_b$.}
    \label{fig:bias-vs-trajectory-length-discount-0-7-policy-divergence-1-44}
\end{figure}

In \cref{fig:bias-vs-trajectory-length-discount-0-7-policy-divergence-1-44}, the PDWIS estimators have higher bias than IS and PDIS $\forall \ T$; this is expected, as these estimators are known to be biased \citep{precup2000}. The PDIS estimators have $|Bias| \approx 0 \ \forall \ T$, while IS has $|Bias| \approx 0$ for $T \le 10$. For $T > 10$, the bias of IS scales rapidly. The reason is that as $T$ increases, it is more likely that trajectories occurring under $\pi_e$ would not be covered by the dataset sampled from $\pi_b$. This is particularly relevant to IS estimators, where IS weights are assigned to entire trajectories. \citet{Sachdeva2020} stated that in the absence of coverage, IS estimators are biased, with bias equal to the expected reward under $\pi_e$ of following the non-covered trajectories. PDIS improves coverage by using IS weights for shorter trajectories at each time step, thereby lowering the bias. Incorporating factored action spaces (as in DecPDIS) further improves coverage and reduces bias $\forall \ T$. 

\paragraph{Empirical Finding 3.} \textit{The variances of the decomposed estimators grow more slowly than their non-decomposed versions, with increasing $T$ and/or $PD(\pi_b, \pi_e)$. This is because $\pi_b$ and $\pi_e$ overlap more in the sub-action spaces i.e. the decomposed estimators improve coverage}.

\begin{figure}[h!]
    \centering
    \vspace{-0.2cm}
    \includegraphics[scale=0.52]{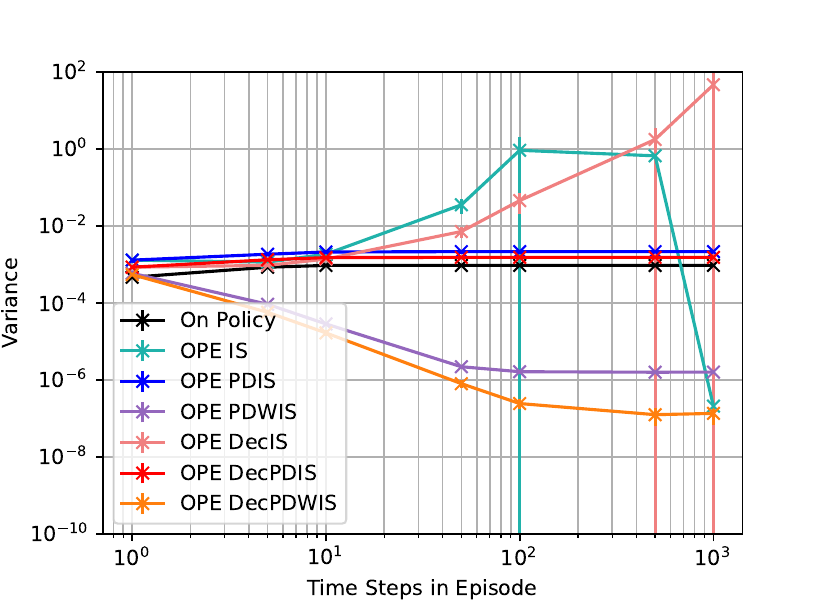}
    \vspace{-0.2cm}
    \includegraphics[scale=0.52]{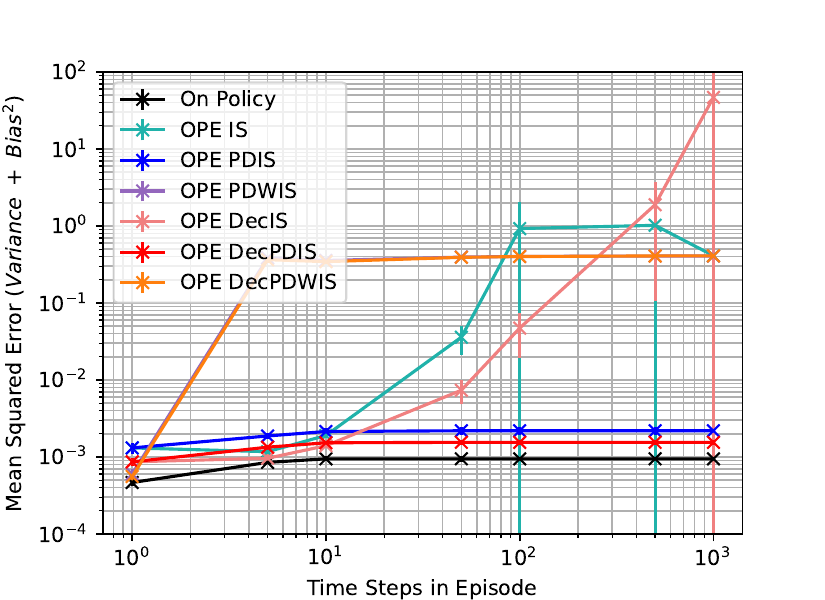}
    \caption{Variance (top) and MSE (bottom) of different OPE estimators as the episode length ($T$) varies, for MDP-2 with $\gamma = 0.7$, $N=1000$ and $PD(\pi_e, \pi_b) = 1.44^T$. Comparing the speed of scaling of variance/MSE, we generally see that OPE IS $>$ OPE DecIS $>$ OPE PDIS $\approx$ OPE DecPDIS. PDWIS has the best variance scaling behavior i.e. negative scaling.}
    \label{fig:var-vs-trajectory-length-discount-0-7-policy-divergence-1-44}
\end{figure}

\cref{fig:var-vs-trajectory-length-discount-0-7-policy-divergence-1-44} shows the variance and MSE for the same experiments in \cref{fig:bias-vs-trajectory-length-discount-0-7-policy-divergence-1-44}. The variances of PDIS and IS estimators both scale with $T$, although the scaling is less pronounced for PDIS estimators. As illustrated in \cref{images-MDP-2-gamma-0-9,images-MDP-2-gamma-0-9999}, the scaling becomes more pronounced when $\gamma$ is increased. The decomposed estimators always have lower variance than their non-decomposed counterparts, sometimes scaling differently - this is true even for the PDWIS estimators where the variance decreases with $T$. The drop in variance of the IS estimator for large $T$ is again explained by low coverage - for these cases, the IS estimator is so biased that it only estimates values close to zero, which results in low variance. Once $T$ is large enough, all estimators would reach this state as they lose coverage of $\pi_e$.

\begin{figure}[h!]
    \centering
    \vspace{-0.3cm}
    \includegraphics[scale=0.44]{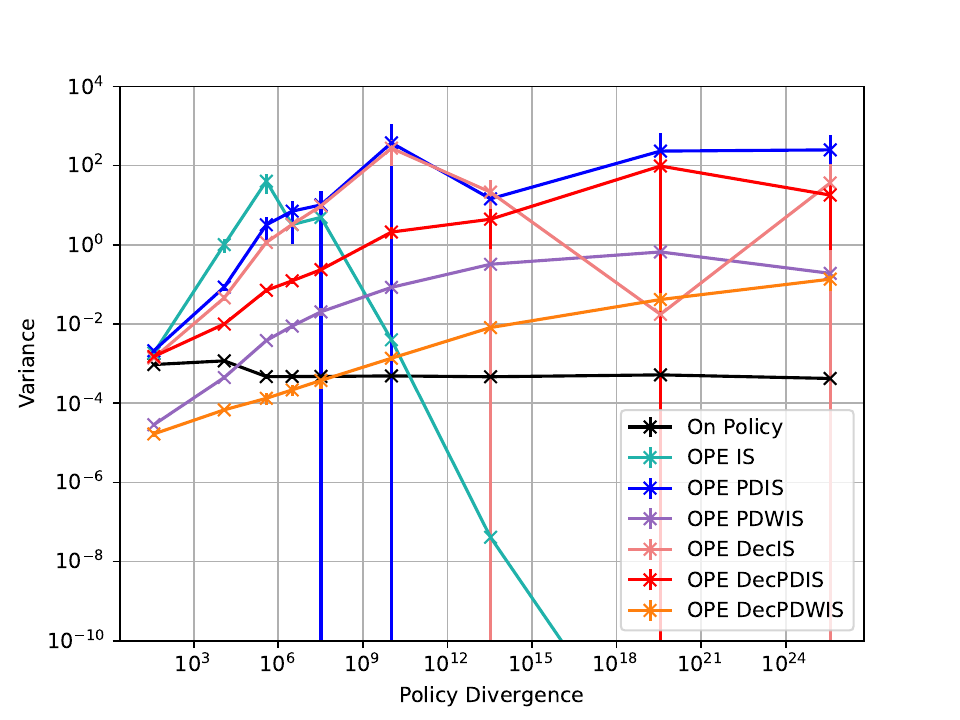}
    \vspace{-0.3cm}
    \includegraphics[scale=0.44]{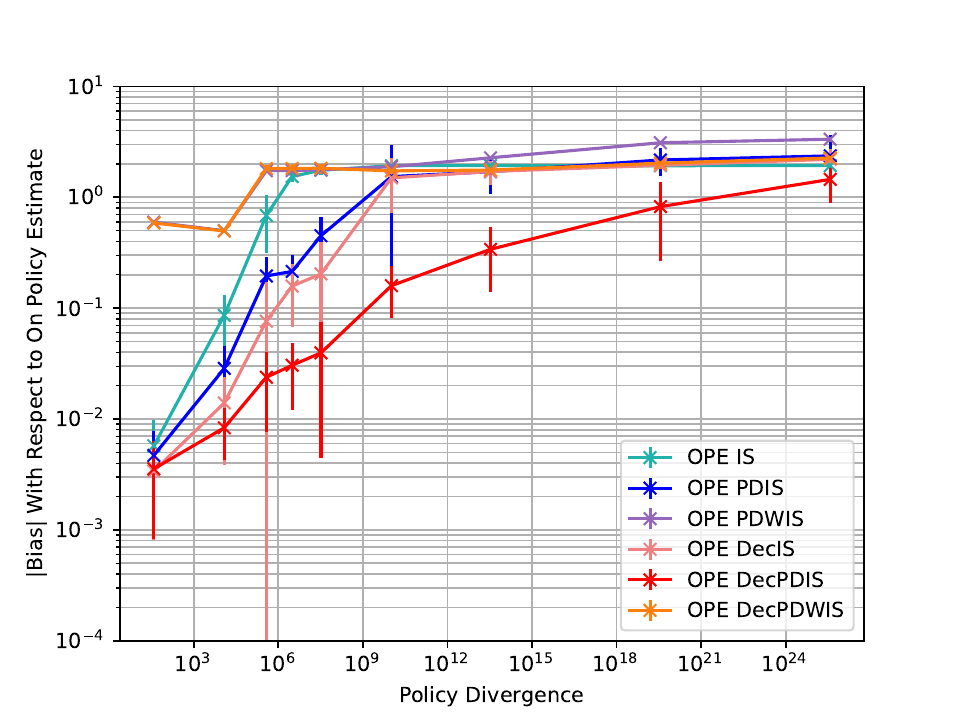}
    \vspace{-0.2cm}
    \caption{Variance (top) and MSE (bottom) of different OPE estimators as the policy divergence ($PD$) varies, for MDP-2 with $\gamma = 0.7$, $N=1000$ and $T=10$. As policy divergence increases, the coverage of $\pi_e$ by the $\pi_b$ data decreases, causing increases in bias (and reductions in variance). The PDWIS estimators were biased even when policy divergence is small.}
    \label{fig:var-10-steps-vs-policy-divergence}
\end{figure}

In \cref{fig:var-10-steps-vs-policy-divergence}, $\pi_b$ and $\pi_e$ were varied to adjust $PD(\pi_e, \pi_b)$. The exact policy configurations are discussed in \cref{subsec:policies-MDP2}. Here, it is seen that all the variances initially scale with the policy divergence but, one-by-one, the IS and PDIS estimators lose coverage and drop in variance. This drop in variance is accompanied by a rise in bias. It is seen that the decomposed PDIS and IS scale the most slowly in bias, which indicates that they improve coverage.

\paragraph{Empirical Finding 4.} \textit{The decomposed estimators have higher ESS than their non-decomposed versions for most tested values of $N$ and $T$. This implies that they have higher data sample efficiency}.

\begin{figure}[h!]
    \centering
    \includegraphics[scale=0.45]{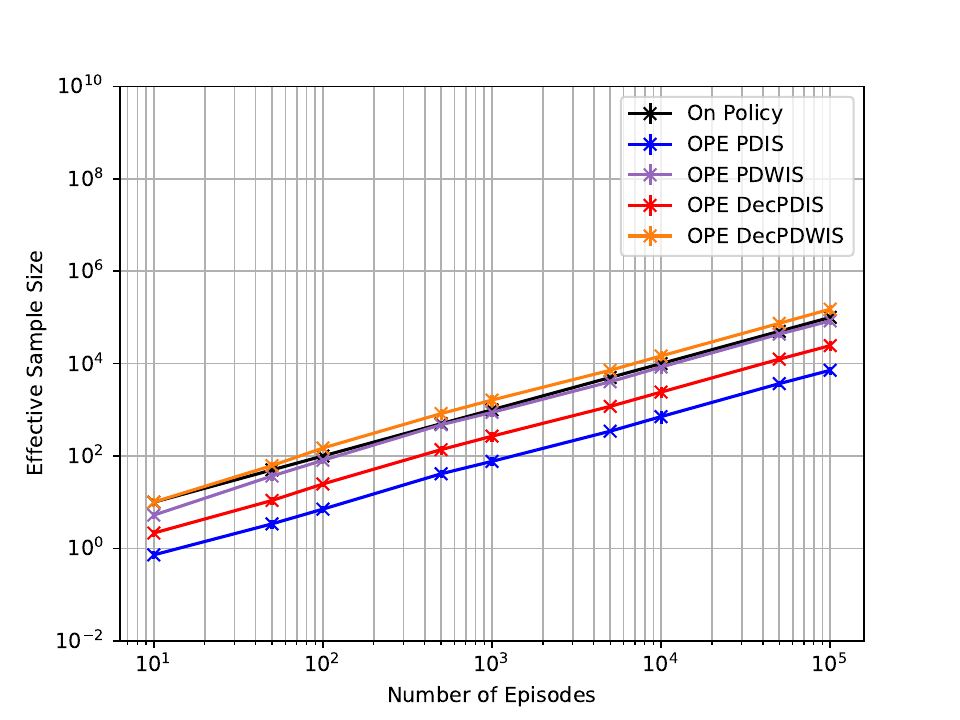}
    \caption{The effective sample size (ESS) of different OPE estimators as the number of episodes ($N$) varies, for MDP-1 with $PD(\pi_b, \pi_e) = 2.56$. The graph is almost an inverted version of the variance graph in \cref{fig:var-vs-trajectories-policy-divergence-2-56}, showing the same precedence in estimators.}
    \label{fig:ESS-vs-trajectories-policy-divergence-2-56}
\end{figure}

\begin{figure}[h!]
    \centering
    \includegraphics[scale=0.53]{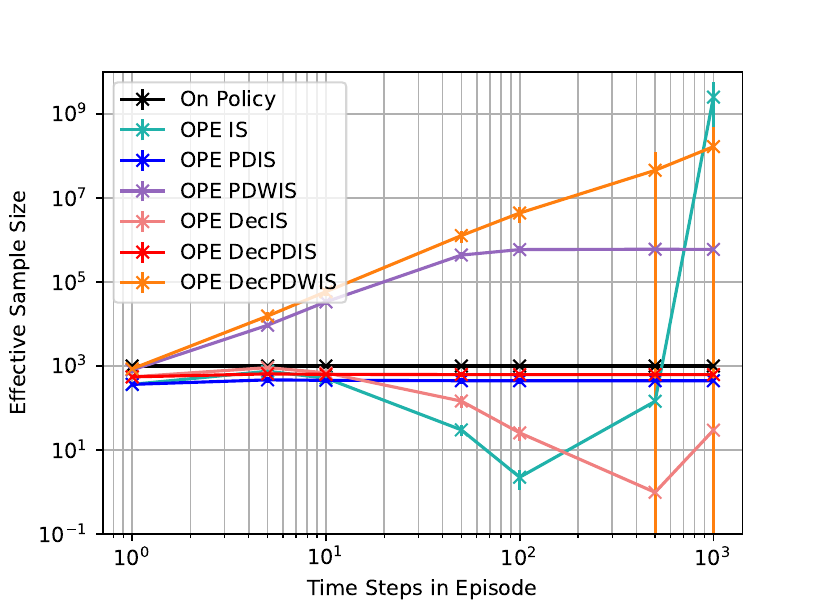}
    \caption{The effective sample size (ESS) of different OPE estimators as the episode length ($T$) varies, for MDP-2 with $\gamma = 0.7$, $N=1000$ and $PD(\pi_b, \pi_e) = 1.44^T$. The graph is almost an inverted version of \cref{fig:var-vs-trajectory-length-discount-0-7-policy-divergence-1-44}. The IS estimators demonstrate how the ESS decreases with increasing $T$ due to greater mismatch in trajectory trajectories from $\pi_b$ and $\pi_e$.}
    \label{fig:ESS-vs-trajectory-length-discount-0-7-policy-divergence-1-44}
\vspace{1cm}
\end{figure}

The ESS of an IS estimator, as defined in \cref{eq:ESS-definition}, is inversely proportional to the variance of the estimator. While it is also directly proportional to the number of trajectories $N$, most of the ESS graphs plotted appear as inverted variance graphs, as seen in \cref{fig:ESS-vs-trajectories-policy-divergence-2-56,fig:ESS-vs-trajectory-length-discount-0-7-policy-divergence-1-44}, which respectively correspond to \cref{fig:var-vs-trajectories-policy-divergence-2-56,fig:var-vs-trajectory-length-discount-0-7-policy-divergence-1-44}.

An interesting insight is the high efficiency of the PDWIS estimators, especially the decomposed version, compared to others. For most values of $N$ and $T$ in \cref{fig:ESS-vs-trajectories-policy-divergence-2-56} and \cref{fig:ESS-vs-trajectory-length-discount-0-7-policy-divergence-1-44}, the decomposed PDWIS estimator has higher ESS than the on-policy estimate itself - this may be due to the advantages of factored actions and weighting, which are not available to the latter. Due to lower variance, the decomposed estimators have higher ESS than their non-decomposed equivalent estimators, which implies they use available data samples more efficiently.

\vspace{-0.5em}
\section{Conclusion}
\vspace{-0.25em}
In this paper, we study the role of factored action spaces in improving the variance, mean squared error and data sample efficiency of off-policy evaluation (OPE) without increasing the bias. Since OPE is a counterfactual query on a set of collected data, such improvements enable more accurate knowledge of the benefit/disadvantage of these counterfactual outcomes.  By proposing a new family of decomposed importance sampling (IS) estimators that utilise the factorisation structure of actions, we have demonstrated theoretically and empirically the potential of leveraging factored action spaces to improve IS OPE. 

Of note, unlike \citet{tang2022leveraging} where only \textit{implicit} MDP factorisation is required, our decomposed IS estimators require \textit{explicit} knowledge of the state abstractions, the reward factorisation and policy factorisations (for both behavior and evaluation policies), as seen in the definitions of DecPDIS and DecPDWIS (interestingly, we do not require knowledge of transition factorisations). Admittedly, an action space factorisation (along with the reward/policy factorisations) that satisfies the sufficient conditions we have outlined for variance reduction and/or zero bias may not always exist, or may be challenging to derive in practice. Future work should investigate the practicality of our approach; for example, using offline data to design the state abstraction, policy and reward factorisations based on action factorisations for general and/or specific OPE problems. We believe these are important avenues of future research before the decomposed estimators can be applied to real-life OPE problems. By utilising domain knowledge about problem structures - in this case, how the action space can be factorised - our approach is an important step in improving OPE for high-stakes decision-making domains such as healthcare, which often have complex and combinatorial action spaces.

\bibliography{ref}
\bibliographystyle{icml2023}

\newpage
\appendix
\onecolumn
\section{Derivation of the Decomposed IS Estimator}
\label{sec:decIS-expression}

Let us consider the definition of the importance sampling (IS) estimator, which was introduced in \citet{precup2000}:

\begin{equation}
    \label{IS-raw-notation}
    \hat{Q}_{\pi_e}^{IS} = \frac{1}{N} \sum_{n=1}^{N} \rho_{0:T}^{(n)} \sum_{t=0}^{T} \gamma^t \cdot r(s_t^{(n)}, a_t^{(n)})
\end{equation}

We first note that the number of trajectories $N$, the discount factor $\gamma$ and the number of transitions $T$ in each trajectory are all constants. They do not change with $i$, $t$ or $d$. The remaining terms: $r$ and $\rho$, do depend on $i$ and/or $t$. Next, we apply Equation \ref{eq:reward-factorisation} to write:

\begin{equation}
     \hat{Q}_{\pi_e}^{IS} = \frac{1}{N} \sum_{n=1}^{N} \rho_{0:T}^{(n)} \sum_{t=0}^{T} \gamma^t  \sum_{d=1}^{D} r^d(z_{t}^{(n),d},a_{t}^{(n),d})
\end{equation}

Since $i$, $t$ and $d$ do not depend on each other, we may make the summation over $d$ the outer summation:

\begin{equation}
    \hat{Q}^{IS}_{\pi_e} = 
    \sum_{d=1}^{D} \frac{1}{N} \sum_{n=1}^{N} \rho_{0:T}^{(n)} \sum_{t=0}^{T} \gamma^t \cdot r^d(z_{t}^{(n),d},a_{t}^{(n),d})
\end{equation}

At this point, we may write:

\begin{equation}
\label{q-decomposition}
    \hat{Q}_{\pi_e}^{IS} = 
    \sum_{d=1}^{D} \hat{Q}^{IS,d} 
\end{equation}

where:

\begin{equation}
    \label{sub-IS-estimator}
   \hat{Q}^{IS,d}_{\pi_e} = \frac{1}{N} \sum_{n=1}^{N} \rho_{0:T}^{(n)} \sum_{t=0}^{T} \gamma^t \cdot r^d(z_{t}^{(n),d},a_{t}^{(n),d})
\end{equation}

In essence, we have decomposed $\hat{Q}_{\pi_e}^{IS}$ based on the factored action spaces, in the form of Equation \ref{eq:factored-Q-decomposition}. However, each estimate $\hat{Q}_{\pi_e}^{IS,d}$ has the full $\rho_{0:T}^{(n)}$ and we now examine whether this is necessary. To do this we observe the definition of $\rho_{0:T}^{(n)}$ in Equation \ref{IS-raw-notation}: 

\begin{equation}
    \rho_{0:T}^{(n)} = \prod_{t=0}^{T} \frac{\pi_e(a^{(n)}_{t+1}|s^{(n)}_{t})}{\pi_b(a^{(n)}_{t+1}|s^{(n)}_{t})} = \frac{\pi_e(a^{(n)}_{1}|s^{(n)}_{0}) \cdot \pi_e(a^{(n)}_{2}|s^{(n)}_{1}) \ ... \ \pi_e(a^{(n)}_{T+1}|s^{(n)}_{T})}{\pi_b(a^{(n)}_{1}|s^{(n)}_{0}) \cdot \pi_b(a^{(n)}_{2}|s^{(n)}_{1}) \ ... \ \pi_b(a^{(n)}_{T+1}|s^{(n)}_{T})}
\end{equation}

We may apply Equation \ref{eq:policy-factorisation} to the numerator and denominator to write:

\begin{equation}
    \label{rho-expansion}
    \rho_{0:T}^{(n)} = \prod_{t=0}^{T} \frac{\pi_e(a^{(n)}_{t+1}|s^{(n)}_{t})}{\pi_b(a^{(n)}_{t+1}|s^{(n)}_{t})} =
    \prod_{t=0}^{T} \frac{\prod_{d=1}^{D} \pi_{e}^{d}(a^{(n),d}_{T+1}|z^{(n),d}_{T})}{\prod_{d=1}^{D} \pi_{b}^{d}(a^{(n),d}_{T+1}|z^{(n),d}_{T})}
\end{equation}

$$
= \frac{\pi_{e}^{1}(a^{(n),1}_{1}|z^{(n),1}_{0}) ... \pi_{e}^{D}(a^{(n),D}_{1}|z^{(n),D}_{0}) ... \pi_{e}^{1}(a^{(n), 1}_{T+1}|z^{(n),1}_{T}) ... \pi_{e}^{D}(a^{(n),D}_{T+1}|z^{(n),D}_{T})}{\pi_{b}^{1}(a^{(n), 1}_{1}|z^{(n), 1}_{0}) ...  \pi_{b}^{D}(a^{(n),D}_{1}|z^{(n),D}_{0}) ... \pi_{b}^{1}(a^{(n), 1}_{T+1}|z^{(n), 1}_{T}) ... \pi_{b}^{D}(a^{(n),D}_{T+1}|z^{(n),D}_{T})}
$$

The continued products in the numerator and denominator contain terms generated by the policies $\pi_{d}^e$ and $\pi_{d}^b$ corresponding to each action space where $\mathcal{A}^d$. 

In Equation \ref{sub-IS-estimator}, we see that $\hat{Q}^{IS,d}$ contains only reward terms $r^d(z_{t}^{(n),d},a_{t}^{(n),d})$ corresponding to the $d^{th}$ action space. The purpose of importance sampling is to account for the fact that each sampled reward term is generated from a behaviour policy, but used to evaluate an evaluation policy. In the context of $\hat{Q}^{IS,d}$, the sampled reward term is $r^d(z_{t}^{(n),d},a_{t}^{(n),d})$, the behaviour policy is $\pi_{b}^{d}$ and the evaluation policy is $\pi_{e}^{d}$. Thus we propose $\rho_{0:T}^{(n), d}$ such that:

\begin{equation}
    \label{better-rho}
    \rho_{0:T}^{(n), d} = \prod_{t=0}^{T} \frac{\pi_{e}^{d}(a^{(n),d}_{t+1}|z^{(n),d}_{t})}{\pi_{b}^{d}(a^{(n),d}_{t+1}|z^{(n),d}_{t})} = 
    \frac{\pi_{e}^{d}(a^{(n),d}_{1}|z^{(n),d}_{0}) \cdot \pi_{e}^{d}(a^{(n),d}_{2}|z^{(n),d}_{1}) \ ... \ \pi_{e}^{d}(a^{(n),d}_{T+1}|z^{(n),d}_{T})}{\pi_{b}^{d}(a^{(n),d}_{1}|z^{(n),d}_{0}) \cdot \pi_{b}^{d}(a^{(n),d}_{2}|z^{(n),d}_{1}) \ ... \ \pi_{b}^{d}(a^{(n),d}_{T+1}|z^{(n),d}_{T})}
\end{equation}

\hspace{0.5cm}

This only contains the terms relevant to the $d^{th}$ action space. We note that we have essentially decomposed $\rho_{0:T}^{(n)}$:

\begin{equation}
    \rho_{0:T}^{(n)} = \prod_{d=1}^{D} \rho_{0:T}^{(n), d}
\end{equation}

We thus propose a new version of $\hat{Q}_{\pi_e}^{IS,d}$ as:

\begin{equation}
    \label{better-sub-IS-estimator}
    \hat{Q}_{\pi_e}^{DecIS, \ d} = \frac{1}{N} \sum_{n=1}^{N} \rho_{0:T}^{(n),d} \sum_{t=0}^{T} \gamma^t \cdot r^d(z_{t}^{(n),d},a_{t}^{(n),d})
\end{equation}

and the overall decomposed IS estimator would be given by:

\begin{equation}
    \label{overall-decomposed-IS}
    \hat{Q}_{\pi_e}^{DecIS} = 
    \sum_{d=1}^{D} \hat{Q}_{\pi_e}^{DecIS, \ d} = \sum_{d=1}^{D} \frac{1}{N} \sum_{n=1}^{N} \rho_{0:T}^{(n),d} \sum_{t=0}^{T} \gamma^t \cdot r^d(z_{t}^{(n),d},a_{t}^{(n),d}) 
\end{equation}

The decomposed Per-Decision IS (PDIS) estimator is derived in the same way as above. The expression for the PDIS estimator from \citet{precup2000} is given by:

\begin{equation}
    \label{PDIS-raw-notation}
    \hat{Q}_{\pi_e}^{PDIS} = \frac{1}{N} \sum_{n=1}^{N}  \sum_{t=0}^{T} \rho_{0:t}^{(n)} \cdot \gamma^t \cdot r(s^{(n)}_t, a^{(n)}_t)
\end{equation}

where

\begin{equation}
    \rho_{0:t}^{(n)} = \prod_{t'=0}^{t} \frac{\pi_e(a^{(n)}_{t'+1}|s^{(n)}_{t'})}{\pi_b(a^{(n)}_{t'+1}|s^{(n)}_{t'})}
\end{equation}

We would apply equations \ref{eq:factored-Q-decomposition}, \ref{eq:reward-factorisation} and \ref{eq:policy-factorisation} in the same way to obtain the decomposed PDIS estimator.

\section{Derivation of the Decomposed PDWIS Estimator}
\label{sec:decPDWIS-expression}

The Per-Decision Weighted IS (PDWIS) estimator \citep{precup2000} is defined as:

\begin{equation}
\label{PDWIS}
    \hat{Q}^{IS} = \frac{\sum_{n=1}^{N} \sum_{t=0}^{T} \gamma^t \cdot \rho_{0:t}^{(n)} \cdot r(s^{(n)}_t, a^{(n)}_t)}{\sum_{n=1}^{N} \sum_{t=0}^{T} \gamma^t \cdot \rho_{0:t}^{(n)}}
\end{equation}

Observing the expression for the PDIS estimator in Equation \ref{PDIS-raw-notation}, we see that instead of dividing by $N$, we have divided by the discounted sum of IS weights $\sum_{n=1}^{N} \sum_{t=0}^{T} \gamma^t \cdot \rho_{0:t}^{(n)}$ in Equation \ref{PDWIS}. This is how we convert the PDIS estimator to the PDWIS estimator.

The decomposed PDIS estimator is given by:

\begin{equation}
    \label{overall-decomposed-PDIS}
    \hat{Q}_{\pi_e}^{DecPDIS} = 
    \sum_{d=1}^{D} \hat{Q}_{\pi_e}^{DecPDIS, \ d} = \sum_{d=1}^{D} \frac{1}{N} \sum_{n=1}^{N} \sum_{t=0}^{T} \gamma^t \cdot \rho_{0:t}^{(n),d} \cdot r^d(z_{t}^{(n),d},a_{t}^{(n),d}) 
\end{equation}

We can convert each $\hat{Q}_{\pi_e}^{DecPDIS, \ d}$ into a $\hat{Q}_{\pi_e}^{DecPDWIS, \ d}$ given by:

$$
\hat{Q}_{\pi_e}^{DecPDWIS, \ d} = \frac{\sum_{n=1}^{N} \sum_{t=0}^{T} \gamma^t \cdot \rho_{0:t}^{(n),d} \cdot r^d(z_{t}^{(n),d},a_{t}^{(n),d})}{\sum_{n=1}^{N} \sum_{t=0}^{T} \gamma^t \cdot \rho_{0:t}^{(n),d}}
$$

Then, the decomposed PDWIS estimator is given by:

\begin{equation}
    \label{overall-decomposed-PDWIS}
    \hat{Q}_{\pi_e}^{DecPDWIS} = 
    \sum_{d=1}^{D} \hat{Q}_{\pi_e}^{DecPDWIS, \ d} = \sum_{d=1}^{D} \frac{\sum_{n=1}^{N} \sum_{t=0}^{T} \gamma^t \cdot \rho_{0:t}^{(n),d} \cdot r^d(z_{t}^{(n),d},a_{t}^{(n),d})}{\sum_{n=1}^{N} \sum_{t=0}^{T} \gamma^t \cdot \rho_{0:t}^{(n),d}}
\end{equation}

\section{Bias of the Decomposed Estimators}
\label{sec:dec-bias}

We are finding the bias with respect to the true $Q_{\pi_e}$, which is given by: 

\begin{equation}
    Q_{\pi_e} = \mathbb{E}_{\pi_e} [\sum_{t=0}^{T} \gamma^t \cdot r(s^{(n)}_t, a^{(n)}_{t}) ]
\end{equation}

For each estimator, we calulate the bias using Equation \ref{eq:bias-definition}. For all subsequent derivations, we define $\mathcal{T}(T)$ as the set of all possible trajectories $\tau$, each of length $T$, that can be generated by $\pi_b$ (where a trajectory is as defined in part \ref{sec:problem-setup}).

\subsection{Decomposed IS Estimator}

We first write:

$$
  \mathbb{E}_{\pi_b}[\hat{Q}_{\pi_e}^{DecIS}] = \mathbb{E}_{\pi_b}[\sum_{d=1}^{D} \frac{1}{N} \sum_{n=1}^{N} \rho_{0:T}^{(n),d} \sum_{t=0}^{T} \gamma^t \cdot r^d(z^{(n),d}_t, a^{(n),d}_t)] = \mathbb{E}_{\pi_b}[\sum_{d=1}^{D} \rho_{0:T}^{(n),d} \sum_{t=0}^{T} \gamma^t \cdot r^d(z_{t}^{(n),d},a_{t}^{(n),d})]
$$

where we make use of the fact that $\mathbb{E}[\frac{1}{N} \sum_{i = 1}^{N} X_i] = \mathbb{E}[X_i]$. We then say:

$$
\mathbb{E}_{\pi_b}[\sum_{d=1}^{D} \rho_{0:T}^{(n),d} \sum_{t=0}^{T} \gamma^t \cdot r^d(z^{(n),d}_t, a^{(n),d}_t)] = \sum_{\tau \ \in \ \mathcal{T}(T)} \  (\prod_{t=0}^{T} \pi_b(a^{(n)}_t|s^{(n)}_t)) \cdot \sum_{d=1}^{D} \rho_{0:T}^{(n),d} \sum_{t=0}^{T} \gamma^t \cdot r^d(z^{(n),d}_t, a^{(n),d}_{t})
$$

Here, $(\prod_{t=0}^{T} \pi_b(s^{(n)}_t, a^{(n)}_t))$ denotes the probability of this entire trajectory occurring under $\pi_b$. Given its definition in Equation \ref{better-rho}, the denominator of the $\rho_{0:T}^{(n),d}$ cancels with $(\prod_{t=0}^{T} \pi_b(s^{(n)}_t, a^{(n)}_t))$ to give:

$$
\sum_{\tau \ \in \ \mathcal{T}(T)} \   \sum_{d=1}^{D} (\prod_{t=0}^{T} \prod_{d^{\prime} \ne d} \pi^{d^{\prime}}_{b}(a^{(n), d^{\prime}}_t | z^{(n), d^{\prime}}_t)) \cdot (\prod_{t=0}^{T} \pi^{d}_{e}(a^{(n), d}_t|z^{(n), d}_t)) \sum_{t=0}^{T} \gamma^t \cdot r^d(z^{(n),d}_t, a^{(n),d}_{t}) 
$$

The term $(\prod_{t=0}^{T} \prod_{d^{\prime} \ne d} \pi^{d^{\prime}}_{b}(a^{(n), d^{\prime}}_t|z^{(n), d^{\prime}}_t))$ is the probability under $\pi_b$ of all trajectories occurring in all factored action spaces except $d$. This probability is independent of the other terms that are related to $d$, and the probabilities sum to $1$ over all trajectories, hence we get:

$$
\sum_{\tau \ \in \ \mathcal{T}(T)} \  1 \cdot  \sum_{d=1}^{D} (\prod_{t=0}^{T} \pi^{d}_{e}(a^{(n), d}_t|z^{(n), d}_t)) \sum_{t=0}^{T} \gamma^t \cdot r^d(z^{(n),d}_t, a^{(n),d}_{t}) 
$$

Since $(\prod_{t=0}^{T} \prod_{d^{\prime} \ne d} \pi^{d^{\prime}}_{e}(a^{(n), d^{\prime}}_t|z^{(n), d^{\prime}}_t))$ would also sum to $1$ over all trajectories as it is independent of $d$, we can write:

$$
 = \sum_{\tau \ \in \ \mathcal{T}(T)} \    \sum_{d=1}^{D} (\prod_{t=0}^{T} \prod_{d^{\prime} \ne d} \pi^{d^{\prime}}_{e}(a^{(n), d^{\prime}}_t|z^{(n), d^{\prime}}_t)) \cdot (\prod_{t=0}^{T} \pi^{d}_{e}( a^{(n), d}_t|z^{(n), d}_t)) \sum_{t=0}^{T} \gamma^t \cdot r^d(z^{(n),d}_t, a^{(n),d}_{t}) 
$$

$$
 = \sum_{\tau \ \in \ \mathcal{T}(T)} \  (\prod_{t=0}^{T}  \pi_{e}(a^{(n)}_t|s^{(n)}_t)) \cdot  \sum_{d=1}^{D}  \sum_{t=0}^{T} \gamma^t \cdot r^d(z^{(n),d}_t, a^{(n),d}_{t}) 
$$

$$
= \mathbb{E}_{\pi_e}[ \sum_{d=1}^{D}  \sum_{t=0}^{T} \gamma^t \cdot r^d(z^{(n),d}_t, a^{(n),d}_{t}) ] = \mathbb{E}_{\pi_e}[  \sum_{t=0}^{T} \gamma^t \cdot r(s^{(n)}_t, a^{(n)}_{t}) ] 
$$

Where we have applied Equation \ref{eq:reward-factorisation} i.e. $r(s,a) = \sum_{d=1}^{D} r^d(z^d,a^d)$. Thus it is clear that:

$$
\mathbb{E}_{\pi_b}[\hat{Q}^{DecIS}_{\pi_e}] = \mathbb{E}_{\pi_b}[\sum_{d=1}^{D} \frac{1}{N} \sum_{n=1}^{N} \rho_{0:T}^{(n),d} \sum_{t=0}^{T} \gamma^t \cdot r^d(z^{(n),d}_t, a^{(n),d}_t)] =  \mathbb{E}_{\pi_e} [\sum_{t=0}^{T} \gamma^t \cdot r(s^{(n)}_t, a^{(n)}_{t}) ] = Q_{\pi_e}
$$

That is:

$$
Bias[\tilde{Q}^{IS}] = \mathbb{E}_{\pi_b}[\hat{Q}^{DecIS}_{\pi_e}] -  Q_{\pi_e} = 0
$$

This means that the decomposed IS estimator is unbiased with respect to $Q_{\pi_e}$.

\subsection{Decomposed PDIS Estimator}

The proof is similar to that of the decomposed IS estimator.

$$
\mathbb{E}_{\pi_b}[\hat{Q}_{\pi_e}^{DecPDIS}] = \mathbb{E}_{\pi_b}\Big[\sum_{d=0}^{D} \frac{1}{N} \sum_{n=1}^{N} \sum_{t=0}^{T} \gamma^t \cdot  \rho_{0:t}^{(n),d} \cdot r^d(z_t^{d}, a_{t+1}^d)\Big] = \mathbb{E}_{\pi_b}\Big[\sum_{d=0}^{D} \sum_{t=0}^{T} \gamma^t \cdot  \rho_{0:t}^{(n),d} \cdot r^d(z_t^{d}, a_{t+1}^d)\Big]
$$

$$
= \sum_{\tau \ \in \ \mathcal{T}(T)} \  (\prod_{t=0}^{T} \pi_b(a^{(n)}_t|s^{(n)}_t)) \sum_{t=0}^{T} \gamma^t \cdot \sum_{d=0}^{D} \rho_{0:t}^{(n),d} \cdot r^d(z_t^{d}, a_{t+1}^d)
$$

$$
= \sum_{\tau \ \in \ \mathcal{T}(T)} \  \sum_{t=0}^{T} \gamma^t \cdot \sum_{d=0}^{D} \Big( \prod_{t=0}^{T} \prod_{d^{\prime} \ne d} \pi^{d^{\prime}}_{b}(a^{(n), d^{\prime}}_t|z^{(n), d^{\prime}}_t) \Big) \cdot \Big( \prod_{t=t+1}^{T} \pi^{d}_{b}(a^{(n), d}_t|z^{(n), d}_t) \Big)
$$
$$
\cdot \Big( \prod_{t=0}^{t} \pi^{d}_{e}(a^{(n), d}_t|z^{(n), d}_t) \Big) \cdot r^d(z_t^{d}, a_{t+1}^d)
$$

The term $\Big( \prod_{t=0}^{T} \prod_{d^{\prime} \ne d} \pi^{d^{\prime}}_{b}(a^{(n), d^{\prime}}_t|z^{(n), d^{\prime}}_t) \Big) \cdot \Big( \prod_{t=t+1}^{T} \pi^{d}_{b}(a^{(n), d}_t|z^{(n), d}_t) \Big)$ is a probability value that varies independently of the reward $r^d(z_t^{d}, a_{t+1}^d)$ because it corresponds to unrelated factored action spaces and timesteps. Hence, we can expect that it will sum to $1$ under the nested summations to give:

$$
= \sum_{\tau \ \in \ \mathcal{T}(T)} \  \sum_{t=0}^{T} \gamma^t \cdot \sum_{d=0}^{D} 1 \cdot \Big( \prod_{t=0}^{t} \pi^{d}_{e}(a^{(n), d}_t|z^{(n), d}_t) \Big) \cdot r^d(z_t^{d}, a_{t+1}^d)
$$
$$
=  \sum_{t=0}^{T} \gamma^t \cdot \sum_{d=0}^{D} \sum_{\tau \ \in \ \mathcal{T}(T)} \  \Big( \prod_{t=0}^{t} \pi^{d}_{e}(a^{(n), d}_t|z^{(n), d}_t) \Big) \cdot r^d(z_t^{d}, a_{t+1}^d) =  \sum_{t=0}^{T} \gamma^t \cdot \sum_{d=0}^{D} \mathbb{E}_{\pi_e} \Big[r^d(z_t^{d}, a_{t+1}^d) \Big]
$$

Then, since expectation is a linear operator, we get:

$$
\mathbb{E}_{\pi_e} \Big[ \sum_{t=0}^{T} \gamma^t \cdot \sum_{d=0}^{D} \cdot r^d(z_t^{d}, a_{t+1}^d) \Big] =  \mathbb{E}_{\pi_e} \Big[ \sum_{t=0}^{T} \gamma^t \cdot r(s^{(n)}_t, a^{(n)}_{t}) \Big] = Q_{\pi_e}
$$
 i.e. the true value of the $Q$ function. Thus, the decomposed PDIS estimator is unbiased with respect to the true $Q$-value.

 \subsection{Decomposed PDWIS Estimator}

Here, we find that the PDWIS estimator has non-zero bias:

$$
\mathbb{E}_{\pi_b}[\hat{Q}_{\pi_e}^{DecPDIS}] = \mathbb{E}_{\pi_b} \Big[\sum_{d=0}^{D} \frac{\sum_{n=1}^{N} \sum_{t=0}^{T} \gamma^t \cdot   \rho_{0:t}^{(n),d} \cdot r^d(z_t^{d}, a_{t+1}^d)}{\sum_{n=1}^{N} \sum_{t=0}^{T} \gamma^t \cdot \rho_{0:t}^{(n),d}} \Big] 
$$

$$
= \sum_{\tau \ \in \ \mathcal{T}(T)} \   (\prod_{t=0}^{T} \pi_b(a^{(n)}_t|s^{(n)}_t)) \sum_{d=0}^{D} \frac{\sum_{n=1}^{N} \sum_{t=0}^{T} \gamma^t \cdot   \rho_{0:t}^{(n),d} \cdot r^d(z_t^{d}, a_{t+1}^d)}{\sum_{n=1}^{N} \sum_{t=0}^{T} \gamma^t \cdot \rho_{0:t}^{(n),d}}
$$

$$
= \sum_{\tau \ \in \ \mathcal{T}(T)} \  \sum_{d=0}^{D} \frac{\sum_{n=1}^{N} \sum_{t=0}^{T} \gamma^t \cdot  \Big( \prod_{t=0}^{T} \prod_{d^{\prime} \ne d} \pi^{d^{\prime}}_{b}(z^{(n), d^{\prime}}_t, a^{(n), d^{\prime}}_t) \Big) \cdot \Big( \prod_{t=t+1}^{T} \pi^{d}_{b}(z^{(n), d}_t, a^{(n), d}_t) \Big)}{\sum_{n=1}^{N} \sum_{t=0}^{T} \gamma^t \cdot \rho_{0:t}^{(n),d}}
$$

$$
\times \frac{\Big( \prod_{t=0}^{t} \pi^{d}_{e}(z^{(n), d}_t, a^{(n), d}_t) \Big) \cdot r^d(z_t^{d}, a_{t+1}^d)}{1}
$$

We can no longer say that $\Big( \prod_{t=0}^{T} \prod_{d^{\prime} \ne d} \pi^{d^{\prime}}_{b}(z^{(n), d^{\prime}}_t, a^{(n), d^{\prime}}_t) \Big) \cdot \Big( \prod_{t=t+1}^{T} \pi^{d}_{b}(z^{(n), d}_t, a^{(n), d}_t) \Big)$ will add up to $1$ over $\tau \ \in \ \mathcal{T}(T)$ because each term is weighted differently due to the denominator $\sum_{n=1}^{N} \sum_{t=0}^{T} \gamma^t \cdot \rho_{0:t}^{(n),d}$ varying based on the trajectory. There is then no way to simplify the above expression, hence it is clear that the bias of the decomposed PDWIS estimator is non-zero. The bias would approach zero if the denominator term $\sum_{n=1}^{N} \sum_{t=0}^{T} \gamma^t \cdot \rho_{0:t}^{(n),d}$ is consistently $\approx 1$.

\section{Variance Guarantees for Decomposed Estimators}
\label{sec:dec-var}

In this section, we compare the variances of the decomposed IS, PDIS and PDWIS estimators.

\subsection{Decomposed IS Estimator}
\label{sec:vardev-decomposed-IS}

We utilise Equation \ref{eq:variance-definition} to write out for the decomposed IS estimator:

$$
\mathbb{V}_{\pi_b}[\hat{Q}_{\pi_e}^{DecIS}] = \sum_{d=1}^{D} \sum_{n=1}^{N} \sum_{t=0}^{T} \sum_{t^{\prime}=0}^{T} \frac{\gamma^{t + t^{\prime}}}{N^2} \Big(\mathbb{E}_{\pi_b}\Big[\Big(\rho_{0:T}^{(n),d}\Big)^2 \cdot r^d(z_{t}^{(n),d},a_{t}^{(n),d}) \cdot r^{d}(z_{t^{\prime}}^{(n),d},a_{t^{\prime}}^{(n),d})\Big]
$$

$$
- \mathbb{E}_{\pi_b}\Big[\rho_{0:T}^{(n),d} \cdot r^d(z_{t}^{(n),d},a_{t}^{(n),d}) \Big] \cdot \mathbb{E}_{\pi_b}\Big[\rho_{0:T}^{(n),d} \cdot r^{d}(z_{t^{\prime}}^{(n),d},a_{t^{\prime}}^{(n),d}) \Big] \Big)
$$

For the IS estimator, we have:

$$
\mathbb{V}_{\pi_b}[\hat{Q}_{\pi_e}^{IS}] = \sum_{n=1}^{N} \sum_{t=0}^{T} \sum_{t^{\prime}=0}^{T} \frac{\gamma^{t + t^{\prime}}}{N^2} \Big(\mathbb{E}_{\pi_b}\Big[\Big(\rho_{0:T}^{(n)}\Big)^2 \cdot r(s_{t}^{(n)},a_{t}^{(n)}) \cdot r(s_{t^{\prime}}^{(n)},a_{t^{\prime}}^{(n)})\Big]
$$

$$
- \mathbb{E}_{\pi_b}\Big[\rho_{0:T}^{(n)} \cdot r(s_{t}^{(n)},a_{t}^{(n)}) \Big] \cdot \mathbb{E}_{\pi_b}\Big[\rho_{0:T}^{(n)} \cdot r(s_{t^{\prime}}^{(n)},a_{t^{\prime}}^{(n)}) \Big] \Big)
$$

Applying equations \ref{eq:policy-factorisation} (decomposition of policy) and \ref{eq:reward-factorisation} (decomposition of reward), we get:

$$
\mathbb{V}_{\pi_b}[\hat{Q}^{IS}] = \sum_{n=1}^{N} \sum_{t=0}^{T} \sum_{t^{\prime}=0}^{T} \frac{\gamma^{t + t^{\prime}}}{N^2} \Big(\mathbb{E}_{\pi_b}\Big[\Big(\prod_{d=0}^{D} \rho_{0:T}^{(n), d}\Big)^2 \cdot \Big( \sum_{d=0}^{D} r^d(z_{t}^{(n),d},a_{t}^{(n),d}) \Big) \cdot \Big( \sum_{d=0}^{D} r^d(z_{t^{\prime}}^{(n),d},a_{t^{\prime}}^{(n),d}) \Big)\Big]
$$

$$
- \mathbb{E}_{\pi_b}\Big[\Big(\prod_{d=0}^{D} \rho_{0:T}^{(n), d}\Big) \cdot \Big( \sum_{d=0}^{D} r^d(z_{t}^{(n),d},a_{t}^{(n),d}) \Big) \Big] \cdot \mathbb{E}_{\pi_b}\Big[\Big(\prod_{d=0}^{D} \rho_{0:T}^{(n), d}\Big) \cdot \Big( \sum_{d=0}^{D} r^d(z_{t^{\prime}}^{(n),d},a_{t^{\prime}}^{(n),d}) \Big) \Big]
$$

\subsubsection{Simplified Case of D=2}
\label{subsec-simplified}

We consider a simplified case where the number of factored action spaces $D=2$; we let the labels $d = \{0,1\}$. In this case:

$$
\mathbb{V}_{\pi_b}[\hat{Q}_{\pi_e}^{IS}] = \sum_{n=1}^{N} \sum_{t=0}^{T} \sum_{t^{\prime}=0}^{T} \frac{\gamma^{t + t^{\prime}}}{N^2} \Big(\mathbb{E}_{\pi_b}\Big[(\rho_{0:T}^{(n), 0})^2 \cdot (\rho_{0:T}^{(n), 1})^2 \cdot r^0(z_{t}^{(n),0},a_{t}^{(n),0}) \cdot r^{0}(z_{t^{\prime}}^{(n),0},a_{t^{\prime}}^{(n),0})  \Big] +
$$

$$
2 \cdot \mathbb{E}_{\pi_b}\Big[(\rho_{0:T}^{(n), 0})^2 \cdot (\rho_{0:T}^{(n), 1})^2 \cdot r^0(z_{t}^{(n),0},a_{t}^{(n),0}) \cdot r^{1}(z_{t^{\prime}}^{(n),1},a_{t^{\prime}}^{(n),1})  \Big] +
$$

$$
\mathbb{E}_{\pi_b}\Big[(\rho_{0:T}^{(n), 0})^2 \cdot (\rho_{0:T}^{(n), 1})^2 \cdot r^1(z_{t}^{(n),1},a_{t}^{(n),1}) \cdot r^{1}(z_{t^{\prime}}^{(n),1},a_{t^{\prime}}^{(n),1})  \Big] -
$$

$$
\mathbb{E}_{\pi_b}\Big[\rho_{0:T}^{(n),0} \cdot \rho_{0:T}^{(n),1} \cdot r^0(z_{t}^{(n),0},a_{t}^{(n),0}) \Big] \cdot \mathbb{E}_{\pi_b}\Big[\rho_{0:T}^{(n),0} \cdot \rho_{0:T}^{(n),1} \cdot r^{0}(z_{t^{\prime}}^{(n),0},a_{t^{\prime}}^{(n),0}) \Big] -
$$

$$
2 \cdot \mathbb{E}_{\pi_b}\Big[\rho_{0:T}^{(n),0} \cdot \rho_{0:T}^{(n),1} \cdot r^0(z_{t}^{(n),0},a_{t}^{(n),0}) \Big] \cdot \mathbb{E}_{\pi_b}\Big[\rho_{0:T}^{(n),0} \cdot \rho_{0:T}^{(n),1} \cdot r^{1}(z_{t^{\prime}}^{(n),1},a_{t^{\prime}}^{(n),1}) \Big] -
$$

$$
\mathbb{E}_{\pi_b}\Big[\rho_{0:T}^{(n),0} \cdot \rho_{0:T}^{(n),1} \cdot r^1(z_{t}^{(n),1},a_{t}^{(n),0}) \Big] \cdot \mathbb{E}_{\pi_b}\Big[\rho_{0:T}^{(n),0} \cdot \rho_{0:T}^{(n),1} \cdot r^{1}(z_{t^{\prime}}^{(n),1},a_{t^{\prime}}^{(n),1}) \Big] \Big)
$$

And for the decomposed IS estimator $\hat{Q}_{\pi_e}^{DecIS}$, we get:

$$
\mathbb{V}_{\pi_b}[\hat{Q}_{\pi_e}^{DecIS}] = \sum_{n=1}^{N} \sum_{t=0}^{T} \sum_{t^{\prime}=0}^{T} \frac{\gamma^{t + t^{\prime}}}{N^2} \Big(\mathbb{E}_{\pi_b}\Big[\Big(\rho_{0:T}^{(n),0}\Big)^2 \cdot r^0(z_{t}^{(n),0},a_{t}^{(n),0}) \cdot r^{0}(z_{t^{\prime}}^{(n),0},a_{t^{\prime}}^{(n),0})\Big]
$$

$$
- \mathbb{E}_{\pi_b}\Big[\rho_{0:T}^{(n),0} \cdot r^0(z_{t}^{(n),0},a_{t}^{(n),0}) \Big] \cdot \mathbb{E}_{\pi_b}\Big[\rho_{0:T}^{(n),0} \cdot r^{0}(z_{t^{\prime}}^{(n),0},a_{t^{\prime}}^{(n),0}) \Big]
$$

$$
+ \mathbb{E}_{\pi_b}\Big[\Big(\rho_{0:T}^{(n),1}\Big)^2 \cdot r^1(z_{t}^{(n),1},a_{t}^{(n),1}) \cdot r^{1}(z_{t^{\prime}}^{(n),1},a_{t^{\prime}}^{(n),1})\Big]
$$

$$
- \mathbb{E}_{\pi_b}\Big[\rho_{0:T}^{(n),1} \cdot r^1(z_{t}^{(n),1},a_{t}^{(n),1}) \Big] \cdot \mathbb{E}_{\pi_b}\Big[\rho_{0:T}^{(n),1} \cdot r^{1}(z_{t^{\prime}}^{(n),1},a_{t^{\prime}}^{(n),1}) \Big] \Big)
$$

The following terms in $\mathbb{V}_{\pi_b}[\hat{Q}_{\pi_e}^{IS}]$ and $\mathbb{V}_{\pi_b}[\hat{Q}_{\pi_e}^{DecIS}]$ correspond to each other and are similar but slightly different:

\begin{itemize}
    \item $\mathbb{E}_{\pi_b}\Big[(\rho_{0:T}^{(n), 0})^2 \cdot (\rho_{0:T}^{(n), 1})^2 \cdot r^0(z_{t}^{(n),0},a_{t}^{(n),0}) \cdot r^{0}(z_{t^{\prime}}^{(n),0},a_{t^{\prime}}^{(n),0})  \Big]$ and $\mathbb{E}_{\pi_b}\Big[\Big(\rho_{0:T}^{(n),0}\Big)^2 \cdot r^0(z_{t}^{(n),0},a_{t}^{(n),0}) \cdot r^{0}(z_{t^{\prime}}^{(n),0},a_{t^{\prime}}^{(n),0})\Big]$
    \item $\mathbb{E}_{\pi_b}\Big[(\rho_{0:T}^{(n), 0})^2 \cdot (\rho_{0:T}^{(n), 1})^2 \cdot r^1(z_{t}^{(n),1},a_{t}^{(n),1}) \cdot r^{1}(z_{t^{\prime}}^{(n),1},a_{t^{\prime}}^{(n),1})  \Big]$ and $ \mathbb{E}_{\pi_b}\Big[\Big(\rho_{0:T}^{(n),1}\Big)^2 \cdot r^1(z_{t}^{(n),1},a_{t}^{(n),1}) \cdot r^{1}(z_{t^{\prime}}^{(n),1},a_{t^{\prime}}^{(n),1})\Big]$
    \item $\mathbb{E}_{\pi_b}\Big[\rho_{0:T}^{(n),0} \cdot \rho_{0:T}^{(n),1} \cdot r^0(z_{t}^{(n),0},a_{t}^{(n),0}) \Big] \cdot \mathbb{E}_{\pi_b}\Big[\rho_{0:T}^{(n),0} \cdot \rho_{0:T}^{(n),1} \cdot r^{0}(z_{t^{\prime}}^{(n),0},a_{t^{\prime}}^{(n),0}) \Big]$ and $\mathbb{E}_{\pi_b}\Big[\rho_{0:T}^{(n),0} \cdot r^0(z_{t}^{(n),0},a_{t}^{(n),0}) \Big] \cdot \mathbb{E}_{\pi_b}\Big[\rho_{0:T}^{(n),0} \cdot r^{0}(z_{t^{\prime}}^{(n),0},a_{t^{\prime}}^{(n),0}) \Big]$
    \item $\mathbb{E}_{\pi_b}\Big[\rho_{0:T}^{(n),0} \cdot \rho_{0:T}^{(n),1} \cdot r^1(z_{t}^{(n),1},a_{t}^{(n),1}) \Big] \cdot \mathbb{E}_{\pi_b}\Big[\rho_{0:T}^{(n),0} \cdot \rho_{0:T}^{(n),1} \cdot r^{1}(z_{t^{\prime}}^{(n),1},a_{t^{\prime}}^{(n),1})\Big]$ and $\mathbb{E}_{\pi_b}\Big[\rho_{0:T}^{(n),1} \cdot r^1(z_{t}^{(n),1},a_{t}^{(n),1}) \Big] \cdot \mathbb{E}_{\pi_b}\Big[\rho_{0:T}^{(n),1} \cdot r^{1}(z_{t^{\prime}}^{(n),1},a_{t^{\prime}}^{(n),1}) \Big]$
\end{itemize}

Additionally, The expression for $\mathbb{V}_{\pi_b}[\hat{Q}_{\pi_e}^{IS}]$ has the following extra terms:

\begin{itemize}
    \item $2 \cdot \mathbb{E}_{\pi_b}\Big[(\rho_{0:T}^{(n), 0})^2 \cdot (\rho_{0:T}^{(n), 1})^2 \cdot r^0(z_{t}^{(n),0},a_{t}^{(n),0}) \cdot r^{1}(z_{t^{\prime}}^{(n),1},a_{t^{\prime}}^{(n),1})  \Big]$
    \item $-2 \cdot \mathbb{E}_{\pi_b}\Big[\rho_{0:T}^{(n),0} \cdot \rho_{0:T}^{(n),1} \cdot r^0(z_{t}^{(n),0},a_{t}^{(n),0}) \Big] \cdot \mathbb{E}_{\pi_b}\Big[\rho_{0:T}^{(n),0} \cdot \rho_{0:T}^{(n),1} \cdot r^{1}(z_{t^{\prime}}^{(n),1},a_{t^{\prime}}^{(n),1}) \Big]$
\end{itemize}

First, we would like to conclude something about whether the sum of the extra terms is greater than or less than zero:

$$
2 \cdot \mathbb{E}_{\pi_b}\Big[(\rho_{0:T}^{(n), 0})^2 \cdot (\rho_{0:T}^{(n), 1})^2 \cdot r^0(z_{t}^{(n),0},a_{t}^{(n),0}) \cdot r^{1}(z_{t^{\prime}}^{(n),1},a_{t^{\prime}}^{(n),1})  \Big] - 
$$
$$
2 \cdot \mathbb{E}_{\pi_b}\Big[\rho_{0:T}^{(n),0} \cdot \rho_{0:T}^{(n),1} \cdot r^0(z_{t}^{(n),0},a_{t}^{(n),0}) \Big] \cdot \mathbb{E}_{\pi_b}\Big[\rho_{0:T}^{(n),0} \cdot \rho_{0:T}^{(n),1} \cdot r^{1}(z_{t^{\prime}}^{(n),1},a_{t^{\prime}}^{(n),1}) \Big]
$$

The above expression is a covariance expression i.e. it is equal to $Cov(\rho_{0:T}^{(n),0} \cdot \rho_{0:T}^{(n),1} \cdot r^0(z_{t}^{(n),0},a_{t}^{(n),0}), \ \rho_{0:T}^{(n),0} \cdot \rho_{0:T}^{(n),1} \cdot r^{1}(z_{t^{\prime}}^{(n),1},a_{t^{\prime}}^{(n),1}))$. This is the covariance between the IS weighted rewards in dimension $0$ and dimension $1$ at times $t$ and $t^{\prime}$ respectively. If this covariance is negative, it would mean that increasing the weighted reward in one action space would cause a (possibly delayed) decrease in weighted reward in another action space, which means that optimising these rewards in different action spaces is like trying to achieve contrasting goals.

Here, we will assume that we have factored the action spaces in such a way that all of them are "aimed" towards a common goal i.e. optimising reward in one action space will either not affect the reward in another action space or increase the reward in this action space. In this case, we would have $Cov(r^0(z_{t}^{(n),0},a_{t}^{(n),0}), \  r^{1}(z_{t^{\prime}}^{(n),1},a_{t^{\prime}}^{(n),1})) \ge 0$. This relates to condition \ref{eq:reward-correlation} from Assumption \ref{ass:variance-bounds}. The product of importance sampling weights i.e. $\rho_{0:T}^{(n),0} \cdot \rho_{0:T}^{(n),1}$ is common to each expression, hence the overall covariance depends only on the reward terms. Therefore, we can assume that $Cov(\rho_{0:T}^{(n),0} \cdot \rho_{0:T}^{(n),1} \cdot r^0(z_{t}^{(n),0},a_{t}^{(n),0}), \ \rho_{0:T}^{(n),0} \cdot \rho_{0:T}^{(n),1} \cdot r^{1}(z_{t^{\prime}}^{(n),1},a_{t^{\prime}}^{(n),1})) \ge 0$ ie. \textit{the sum of the extra terms is never less than zero}. We hold on to this result.

Second, we want to compare $\mathbb{E}_{\pi_b}\Big[(\rho_{0:T}^{(n), d})^2 \cdot (\rho_{0:T}^{(n), d^{\prime}})^2 \cdot r^d(z_{t}^{(n),d},a_{t}^{(n),d}) \cdot r^{d}(z_{t^{\prime}}^{(n),d},a_{t^{\prime}}^{(n),d})  \Big]$ and $\mathbb{E}_{\pi_b}\Big[\Big(\rho_{0:T}^{(n),d}\Big)^2 \cdot r^d(z_{t}^{(n),d},a_{t}^{(n),d}) \cdot r^{d}(z_{t^{\prime}}^{(n),d},a_{t^{\prime}}^{(n),d})\Big]$ for $(d, d^{\prime}) = (0,1), (1,0)$. Let $X = (\rho_{0:T}^{(n), d})^2 \cdot r^d(z_{t}^{(n),d},a_{t}^{(n),d}) \cdot r^{d}(z_{t^{\prime}}^{(n),d},a_{t^{\prime}}^{(n),d})$ and $A = (\rho_{0:T}^{(n), d^{\prime}})^2$. Here, we assume that $Cov(X,A) = 0$ due to the two terms being in different factored action spaces $d$ and $d'$ - this is condition \ref{eq:ratio-correlation} from Assumption \ref{ass:variance-bounds}. However, $Cov(X,A) = E(XA) - E(X)E(A)$, which implies $E(XA) = E(X)E(A) + Cov(X,A)$. Let us consider $E(A) = \mathbb{E}_{\pi_b}[(\rho_{0:T}^{(n), d})^2]$. As mentioned in \citet{NanJiang2022}, we know that the expected value of an IS weight is $1$ i.e. $\mathbb{E}_{\pi_b}[(\rho_{0:T}^{(n), d})] = 1$. Since we know that $Var(\rho_{0:T}^{(n)}) \ge 0$ (true for all variances), we have $\mathbb{E}_{\pi_b}[(\rho_{0:T}^{(n), d})^2] - \mathbb{E}_{\pi_b}[(\rho_{0:T}^{(n), d})]^2 \ge 0$. Thus, $E((\rho_{0:T}^{(n), d})^2) - 1^2 \ge 0$, which implies $E((\rho_{0:T}^{(n), d})^2) \ge 1$, and thus $E(A) \ge 1$. This gives us $E(XA) = \lambda \cdot E(X) + 0$, where $\lambda \ge 1$. Clearly, $E(XA) \ge E(X)$ i.e. $\mathbb{E}_{\pi_b}\Big[(\rho_{0:T}^{(n), d})^2 \cdot (\rho_{0:T}^{(n), d^{\prime}})^2 \cdot r^0(z_{t}^{(n),d},a_{t}^{(n),0}) \cdot r^{d}(z_{t^{\prime}}^{(n),d},a_{t^{\prime}}^{(n),d})  \Big] \ \ge \ \mathbb{E}_{\pi_b}\Big[\Big(\rho_{0:T}^{(n),d}\Big)^2 \cdot r^d(z_{t}^{(n),d},a_{t}^{(n),d}) \cdot r^{d}(z_{t^{\prime}}^{(n),d},a_{t^{\prime}}^{(n),d})\Big]$ for $(d, d^{\prime}) = (0,1), (1,0)$.

\vspace{0.5cm}

Third, we want to compare $\mathbb{E}_{\pi_b}\Big[\rho_{0:T}^{(n),d} \cdot \rho_{0:T}^{(n),d^{\prime}} \cdot r^d(z_{t}^{(n),d},a_{t}^{(n),d}) \Big] \cdot \mathbb{E}_{\pi_b}\Big[\rho_{0:T}^{(n),d} \cdot \rho_{0:T}^{(n),d^{\prime}} \cdot r^{d}(z_{t^{\prime}}^{(n),d},a_{t^{\prime}}^{(n),d})\Big]$ and $\mathbb{E}_{\pi_b}\Big[\rho_{0:T}^{(n),d} \cdot r^d(z_{t}^{(n),d},a_{t}^{(n),d}) \Big] \cdot \mathbb{E}_{\pi_b}\Big[\rho_{0:T}^{(n),d} \cdot r^{d}(z_{t^{\prime}}^{(n),d},a_{t^{\prime}}^{(n),d}) \Big]$. Here, we assume that $Cov(\rho_{0:T}^{(n),d} \cdot r^d(z_{t}^{(n),d},a_{t}^{(n),d}), \ \rho_{0:T}^{(n),d^{\prime}}) = 0$ due to the difference in the factored action space - this utilises conditions \ref{eq:ratio-correlation} and \ref{eq:reward-ratio-correlation} from assumption \ref{ass:variance-bounds}. As a result, $\mathbb{E}_{\pi_b}[\rho_{0:T}^{(n),d} \cdot \rho_{0:T}^{(n),d^{\prime}} \cdot r^d(z_{t}^{(n),d},a_{t}^{(n),d})] - \mathbb{E}_{\pi_b}[\rho_{0:T}^{(n),d^{\prime}}] \cdot \mathbb{E}_{\pi_b}[\rho_{0:T}^{(n),d} \cdot r^{d}(z_{t^{\prime}}^{(n),d},a_{t^{\prime}}^{(n),d})] = 0$. Knowing that $\mathbb{E}_{\pi_b}[\rho_{0:T}^{(n),d^{\prime}}] = 1$ gives us: $\mathbb{E}_{\pi_b}[\rho_{0:T}^{(n),d} \cdot \rho_{0:T}^{(n),d^{\prime}} \cdot r^d(z_{t}^{(n),d},a_{t}^{(n),d})] = \mathbb{E}_{\pi_b}[\rho_{0:T}^{(n),d} \cdot r^{d}(z_{t^{\prime}}^{(n),d},a_{t^{\prime}}^{(n),d})]$. This means that we can say $\mathbb{E}_{\pi_b}\Big[\rho_{0:T}^{(n),d} \cdot \rho_{0:T}^{(n),d^{\prime}} \cdot r^d(z_{t}^{(n),d},a_{t}^{(n),d}) \Big] \cdot \mathbb{E}_{\pi_b}\Big[\rho_{0:T}^{(n),d} \cdot \rho_{0:T}^{(n),d^{\prime}} \cdot r^{d}(z_{t^{\prime}}^{(n),d},a_{t^{\prime}}^{(n),d})\Big] \ = \ \mathbb{E}_{\pi_b}\Big[\rho_{0:T}^{(n),d} \cdot r^d(z_{t}^{(n),d},a_{t}^{(n),d}) \Big] \cdot \mathbb{E}_{\pi_b}\Big[\rho_{0:T}^{(n),d} \cdot r^{d}(z_{t^{\prime}}^{(n),d},a_{t^{\prime}}^{(n),d}) \Big]$.

Overall, the corresponding terms in $\mathbb{V}_{\pi_b}[\hat{Q}_{\pi_e}^{IS}]$ are larger or equal to those in $ \mathbb{V}_{\pi_b}[\hat{Q}_{\pi_b}^{DecIS}]$ and the extra terms in the former also add up to a value $\ge 0$. Thus, clearly in this case, $\mathbb{V}_{\pi_b}[\hat{Q}_{\pi_e}^{IS}] \ge \mathbb{V}_{\pi_b}[\hat{Q}_{\pi_b}^{DecIS}]$.

\subsubsection{Extension to $D > 2$}

We extend the findings from the previous section to the general case. Consider the general expressions for $\mathbb{V}_{\pi_b}[\hat{Q}_{\pi_e}^{DecIS}]$ and $ \mathbb{V}_{\pi_b}[\hat{Q}_{\pi_e}^{IS}]$:

$$
\mathbb{V}_{\pi_b}[\hat{Q}_{\pi_e}^{DecIS}] = \sum_{d=1}^{D} \sum_{n=1}^{N} \sum_{t=0}^{T} \sum_{t^{\prime}=0}^{T} \frac{\gamma^{t + t^{\prime}}}{N^2} \Big(\mathbb{E}_{\pi_b}\Big[\Big(\rho_{0:T}^{(n),d}\Big)^2 \cdot r^d(z_{t}^{(n),d},a_{t}^{(n),d}) \cdot r^{d}(z_{t^{\prime}}^{(n),d},a_{t^{\prime}}^{(n),d})\Big]
$$

$$
- \mathbb{E}_{\pi_b}\Big[\rho_{0:T}^{(n),d} \cdot r^d(z_{t}^{(n),d},a_{t}^{(n),d}) \Big] \cdot \mathbb{E}_{\pi_b}\Big[\rho_{0:T}^{(n),d} \cdot r^{d}(z_{t^{\prime}}^{(n),d},a_{t^{\prime}}^{(n),d}) \Big] \Big)
$$

$$
\mathbb{V}_{\pi_b}[\hat{Q}_{\pi_e}^{IS}] = \sum_{n=1}^{N} \sum_{t=0}^{T} \sum_{t^{\prime}=0}^{T} \frac{\gamma^{t + t^{\prime}}}{N^2} \Big(\mathbb{E}_{\pi_b}\Big[\Big(\prod_{d=0}^{D} \rho_{0:T}^{(n), d}\Big)^2 \cdot \Big( \sum_{d=0}^{D} r^d(z_{t}^{(n),d},a_{t}^{(n),d}) \Big) \cdot \Big( \sum_{d=0}^{D} r^d(z_{t^{\prime}}^{(n),d},a_{t^{\prime}}^{(n),d}) \Big)\Big]
$$

$$
- \mathbb{E}_{\pi_b}\Big[\Big(\prod_{d=0}^{D} \rho_{0:T}^{(n), d}\Big) \cdot \Big( \sum_{d=0}^{D} r^d(z_{t}^{(n),d},a_{t}^{(n),d}) \Big) \Big] \cdot \mathbb{E}_{\pi_b}\Big[\Big(\prod_{d=0}^{D} \rho_{0:T}^{(n), d}\Big) \cdot \Big( \sum_{d=0}^{D} r^d(z_{t^{\prime}}^{(n),d},a_{t^{\prime}}^{(n),d}) \Big) \Big]
$$

A general positive term in $\mathbb{V}_{\pi_b}[\hat{Q}_{\pi_e}^{DecIS}]$ is $PT_1 = \frac{\gamma^{t + t^{\prime}}}{N^2} \mathbb{E}_{\pi_b}\Big[\Big(\rho_{0:T}^{(n),d}\Big)^2 \cdot r^d(z_{t}^{(n),d},a_{t}^{(n),d}) \cdot r^{d}(z_{t^{\prime}}^{(n),d},a_{t^{\prime}}^{(n),d})\Big]$, while a general positive term in $\mathbb{V}_{\pi_b}[\hat{Q}_{\pi_e}^{IS}]$ is $PT_2 = \frac{\gamma^{t + t^{\prime}}}{N^2} \mathbb{E}_{\pi_b}\Big[\Big(\prod_{d=0}^{D} \rho_{0:T}^{(n), d}\Big)^2 \cdot r^{d_1}(z_{t}^{(n),d_1},a_{t}^{(n),d_1}) \cdot r^{d_2}(z_{t^{\prime}}^{(n),d_2},a_{t^{\prime}}^{(n),d_2})\Big]$. Looking at $PT_2$, when $d_1 = d_2 = d$, $PT_2$ corresponds to $PT_1$ in that the terms are similar. When $d_1 \ne d_2$, $PT_2$ is extra i.e. has no corresponding term in  $\mathbb{V}_{\pi_b}[\hat{Q}_{\pi_e}^{DecIS}]$.

A general negative term in $\mathbb{V}_{\pi_b}[\hat{Q}_{\pi_e}^{DecIS}]$ is $NT_1 = -\frac{\gamma^{t + t^{\prime}}}{N^2} \mathbb{E}_{\pi_b}\Big[\rho_{0:T}^{(n),d} \cdot r^d(z_{t}^{(n),d},a_{t}^{(n),d}) \Big] \cdot \mathbb{E}_{\pi_b}\Big[\rho_{0:T}^{(n),d} \cdot r^{d}(z_{t^{\prime}}^{(n),d},a_{t^{\prime}}^{(n),d}) \Big]$, while a general negative term in $\mathbb{V}_{\pi_b}[\hat{Q}_{\pi_e}^{IS}]$ is $NT_2 = -\frac{\gamma^{t + t^{\prime}}}{N^2} \mathbb{E}_{\pi_b}\Big[\Big(\prod_{d=0}^{D} \rho_{0:T}^{(n), d}\Big) \cdot r^{d_1}(z_{t}^{(n),d_1},a_{t}^{(n),d_1}) \Big] \cdot \mathbb{E}_{\pi_b}\Big[\Big(\prod_{d=0}^{D} \rho_{0:T}^{(n), d} \Big) \cdot r^{d_2}(z_{t^{\prime}}^{(n),d_2},a_{t^{\prime}}^{(n),d_2}) \Big]$. Again, when $d_1 = d_2 = d$, then $NT_2$ corresponds to $NT_1$ and when $d_1 \ne d_2$, $NT_2$ is extra.

Combining positive and negative terms, a general extra term in $\mathbb{V}_{\pi_b}[\hat{Q}_{\pi_e}^{DecIS}]$ that has no counterpart in $\mathbb{V}_{\pi_b}[\hat{Q}_{\pi_e}^{IS}]$ is  $\frac{\gamma^{t + t^{\prime}}}{N^2} \Big( \mathbb{E}_{\pi_b}\Big[\Big(\prod_{d=0}^{D} \rho_{0:T}^{(n), d}\Big)^2 \cdot r^{d_1}(z_{t}^{(n),d_1},a_{t}^{(n),d_1}) \cdot r^{d_2}(z_{t^{\prime}}^{(n),d_2},a_{t^{\prime}}^{(n),d_2})\Big] - \mathbb{E}_{\pi_b}\Big[\Big(\prod_{d=0}^{D} \rho_{0:T}^{(n), d}\Big) \cdot r^{d_1}(z_{t}^{(n),d_1},a_{t}^{(n),d_1}) \Big] \cdot \mathbb{E}_{\pi_b}\Big[\Big(\prod_{d=0}^{D} \rho_{0:T}^{(n), d} \Big) \cdot r^{d_2}(z_{t^{\prime}}^{(n),d_2},a_{t^{\prime}}^{(n),d_2}) \Big] \Big)$, which is equal to $\frac{\gamma^{t + t^{\prime}}}{N^2} \cdot Cov\Big(\Big(\prod_{d=0}^{D} \rho_{0:T}^{(n), d}\Big) \cdot r^{d_1}(z_{t}^{(n),d_1},a_{t}^{(n),d_1}), \Big(\prod_{d=0}^{D} \rho_{0:T}^{(n), d} \Big) \cdot r^{d_2}(z_{t^{\prime}}^{(n),d_2},a_{t^{\prime}}^{(n),d_2}) \Big)$. In part \ref{subsec-simplified}, we argued that $Cov(\rho_{0:T}^{(n),0} \cdot \rho_{0:T}^{(n),1} \cdot r^0(z_{t}^{(n),0},a_{t}^{(n),0}), \ \rho_{0:T}^{(n),0} \cdot \rho_{0:T}^{(n),1} \cdot r^{1}(z_{t^{\prime}}^{(n),1},a_{t^{\prime}}^{(n),1})) \ge 0$., based on the assumption that $Cov(r^{d_1}(z_{t}^{(n),d_1},a_{t}^{(n),d_1}), \  r^{d_2}(z_{t^{\prime}}^{(n),d_2},a_{t^{\prime}}^{(n),d_2})) \ge 0$. We can use this assumption to argue that $Cov\Big(\Big(\prod_{d=0}^{D} \rho_{0:T}^{(n), d}\Big) \cdot r^{d_1}(z_{t}^{(n),d_1},a_{t}^{(n),d_1}), \Big(\prod_{d=0}^{D} \rho_{0:T}^{(n), d} \Big) \cdot r^{d_2}(z_{t^{\prime}}^{(n),d_2},a_{t^{\prime}}^{(n),d_2}) \Big) \ge 0$.

We now consider the positive corresponding terms i.e. $PT_1$ and $PT_2$ where $d_1 = d_2$. These terms are $PT_1 = \frac{\gamma^{t + t^{\prime}}}{N^2} \mathbb{E}_{\pi_b}\Big[\Big(\rho_{0:T}^{(n),d}\Big)^2 \cdot r^d(z_{t}^{(n),d},a_{t}^{(n),d}) \cdot r^{d}(z_{t^{\prime}}^{(n),d},a_{t^{\prime}}^{(n),d})\Big]$ and $PT_2 = \frac{\gamma^{t + t^{\prime}}}{N^2} \mathbb{E}_{\pi_b}\Big[\Big(\prod_{d=0}^{D} \rho_{0:T}^{(n), d}\Big)^2 \cdot r^{d}(z_{t}^{(n),d},a_{t}^{(n),d}) \cdot r^{d}(z_{t^{\prime}}^{(n),d},a_{t^{\prime}}^{(n),d})\Big]$. In part \cref{subsec-simplified}, we argued that $\mathbb{E}_{\pi_b}\Big[(\rho_{0:T}^{(n), d})^2 \cdot (\rho_{0:T}^{(n), d^{\prime}})^2 \cdot r^0(z_{t}^{(n),d},a_{t}^{(n),0}) \cdot r^{d}(z_{t^{\prime}}^{(n),d},a_{t^{\prime}}^{(n),d})  \Big] \ \ge \ \mathbb{E}_{\pi_b}\Big[\Big(\rho_{0:T}^{(n),d}\Big)^2 \cdot r^d(z_{t}^{(n),d},a_{t}^{(n),d}) \cdot r^{d}(z_{t^{\prime}}^{(n),d},a_{t^{\prime}}^{(n),d})\Big]$ for $(d, d') = (0,1), (1,0)$, using the assumption that an importance sampling ratio in any factored action space is independent of the reward terms and IS ratios in other factored action spaces. Using this assumption, we can say that $\Big(\rho_{0:T}^{(n),d}\Big)^2 \cdot r^d(z_{t}^{(n),d},a_{t}^{(n),d}) \cdot r^{d}(z_{t^{\prime}}^{(n),d},a_{t^{\prime}}^{(n),d})$ and $\Big(\prod_{d^{\prime} \ne d} \rho_{0:T}^{(n), d^{\prime}}\Big)^2$ are uncorrelated. We can also use this assumption to say that $\mathbb{E}_{\pi_b}\Big[\Big(\prod_{d^{\prime} \ne d} \rho_{0:T}^{(n), d^{\prime}}\Big)^2\Big] \ge 1$ because $Var\Big[\prod_{d^{\prime} \ne d} \rho_{0:T}^{(n), d^{\prime}}\Big] \ge 0$ and $\mathbb{E}_{\pi_b}\Big[\prod_{d^{\prime} \ne d} \rho_{0:T}^{(n), d^{\prime}}\Big]^2 = \Big(\prod_{d^{\prime} \ne d} \ \mathbb{E}_{\pi_b}\Big[\rho_{0:T}^{(n), d^{\prime}}\Big]\Big)^2 = \prod_{d^{\prime} \ne d} \ 1$. Here, we have also used the result that $\mathbb{E}_{\pi_b}\Big[\rho_{0:T}^{(n), d^{\prime}}\Big] = 1$ \citep{NanJiang2022}. Therefore, $\mathbb{E}_{\pi_b}\Big[\Big(\prod_{d=0}^{D} \rho_{0:T}^{(n), d}\Big)^2 \cdot r^{d}(z_{t}^{(n),d},a_{t}^{(n),d}) \cdot r^{d}(z_{t^{\prime}}^{(n),d},a_{t^{\prime}}^{(n),d})\Big] \ge \mathbb{E}_{\pi_b}\Big[\Big(\rho_{0:T}^{(n),d}\Big)^2 \cdot r^d(z_{t}^{(n),d},a_{t}^{(n),d}) \cdot r^{d}(z_{t^{\prime}}^{(n),d},a_{t^{\prime}}^{(n),d})\Big]$.

Finally we consider the negative corresponding terms i.e. $NT_1$ and $NT_2$ where $d_1 = d_2$. These terms are $NT_1 = -\frac{\gamma^{t + t^{\prime}}}{N^2} \mathbb{E}_{\pi_b}\Big[\rho_{0:T}^{(n),d} \cdot r^d(z_{t}^{(n),d},a_{t}^{(n),d}) \Big] \cdot \mathbb{E}_{\pi_b}\Big[\rho_{0:T}^{(n),d} \cdot r^{d}(z_{t^{\prime}}^{(n),d},a_{t^{\prime}}^{(n),d}) \Big]$ and  $NT_2 = -\frac{\gamma^{t + t^{\prime}}}{N^2} \mathbb{E}_{\pi_b}\Big[\Big(\prod_{d=0}^{D} \rho_{0:T}^{(n), d}\Big) \cdot r^{d}(z_{t}^{(n),d},a_{t}^{(n),d}) \Big] \cdot \mathbb{E}_{\pi_b}\Big[\Big(\prod_{d=0}^{D} \rho_{0:T}^{(n), d} \Big) \cdot r^{d}(z_{t^{\prime}}^{(n),d},a_{t^{\prime}}^{(n),d}) \Big]$. In part \ref{subsec-simplified}, we argued that $\mathbb{E}_{\pi_b}\Big[\rho_{0:T}^{(n),d} \cdot \rho_{0:T}^{(n),d^{\prime}} \cdot r^d(z_{t}^{(n),d},a_{t}^{(n),d}) \Big] \cdot \mathbb{E}_{\pi_b}\Big[\rho_{0:T}^{(n),d} \cdot \rho_{0:T}^{(n),d^{\prime}} \cdot r^{d}(z_{t^{\prime}}^{(n),d},a_{t^{\prime}}^{(n),d})\Big] \ = \ \mathbb{E}_{\pi_b}\Big[\rho_{0:T}^{(n),d} \cdot r^d(z_{t}^{(n),d},a_{t}^{(n),d}) \Big] \cdot \mathbb{E}_{\pi_b}\Big[\rho_{0:T}^{(n),d} \cdot r^{d}(z_{t^{\prime}}^{(n),d},a_{t^{\prime}}^{(n),d}) \Big]$, using the assumption that an importance sampling ratio in any factored action space is independent of the reward terms and IS ratios in other factored action spaces. Based on this assumption, we can say that $Cov(\rho_{0:T}^{(n),d} \cdot r^d(z_{t}^{(n),d},a_{t}^{(n),d}), \ \prod_{d^{\prime} \ne d} \rho_{0:T}^{(n), d^{\prime}}) = 0$. Furthermore, assuming that $\mathbb{E}_{\pi_b}[ \rho_{0:T}^{(n), d}] = 1$, we can say that $\mathbb{E}_{\pi_b}\Big[\prod_{d^{\prime} \ne d} \rho_{0:T}^{(n), d^{\prime}}\Big] = \prod_{d^{\prime} \ne d} \mathbb{E}_{\pi_b}\Big[\rho_{0:T}^{(n), d^{\prime}}\Big] = \prod_{d^{\prime} \ne d} 1 = 1$. Based on this, we can conclude that ${E}_{\pi_b}\Big[\rho_{0:T}^{(n),d} \cdot r^d(z_{t}^{(n),d},a_{t}^{(n),d}) \Big] = \mathbb{E}_{\pi_b}\Big[\Big(\prod_{d=0}^{D} \rho_{0:T}^{(n), d} \Big) \cdot r^{d}(z_{t}^{(n),d},a_{t}^{(n),d}) \Big]$, and thus $NT_1 = NT_2$.

Since the negative terms are equal, the positive terms in the variance of the original IS estimator are larger and the extra terms are all no less than zero, it is clear that even for $D > 2$, we have that $\mathbb{V}_{\pi_b}[\hat{Q}_{\pi_e}^{DecIS}] \le \mathbb{V}_{\pi_b}[\hat{Q}_{\pi_e}^{IS}]$.

\subsection{Decomposed PDIS}

Given the highly similar natures of the IS and PDIS estimators, the Theorem \ref{theorem-lower-variance-factored-estimator} variance guarantee for the decomposed PDIS estimator may be proven in the exact same way as for decomposed IS in part \ref{sec:vardev-decomposed-IS}. The only difference, is that at all steps, we would replace every occurrence of an episode-based IS ratio of the form $\rho_{0:T}^{(n),d}$ with a per-decision IS ratio of the form $\rho_{0:t}^{(n),d}$.

\subsection{Decomposed PDWIS}

Using Equation \ref{eq:variance-definition}, the variance of the decomposed PDWIS estimator can be derived as below. Note that for brevity, we have replaced $r^{d}(z_{t}^{(n),d},a_{t}^{(n),d})$ with $r^{(n),d}_t$:

$$
\mathbb{V}[\hat{Q}^{DecPDWIS}_{\pi_b}] = \sum_{n=1}^{N} \sum_{d=1}^{D} \sum_{t=0}^{T} \sum_{t^{\prime}=0}^{T} \frac{\gamma^{t+t^{\prime}}}{N^2} \Bigg( \frac{\mathbb{E}_{\pi_b} \Big[ \rho_{0:t}^{(n),d} \cdot \rho_{0:t^{\prime}}^{(n),d} \cdot r^{(n),d}_{t} \cdot r^{(n),d}_{t^{\prime}} \Big]}{\sum_{n=1}^{N} \sum_{t=0}^{T} \sum_{t^{\prime}=0}^{T} \gamma^{t + t^{\prime}} \cdot \mathbb{E}_{\pi_b} \Big[ \rho_{0:t}^{(n),d} \cdot \rho_{0:t^{\prime}}^{(n),d} \Big]} 
$$

$$
- \ \frac{\mathbb{E}_{\pi_b} \Big[ \rho_{0:t}^{(n),d} \cdot r^{(n),d}_{t} \Big] \cdot \mathbb{E}_{\pi_b} \Big[ \rho_{0:t^{\prime}}^{(n),d} \cdot r^{(n),d}_{t^{\prime}} \Big]}{\sum_{n=1}^{N} \sum_{t=0}^{T} \sum_{t^{\prime}=0}^{T} \gamma^{t + t^{\prime}} \cdot \mathbb{E}_{\pi_b} \Big[ \rho_{0:t}^{(n),d} \Big] \cdot \mathbb{E}_{\pi_b} \Big[ \rho_{0:t^{\prime}}^{(n),d} \Big]} \Bigg)
$$

This is similar to the expression for the decomposed IS estimator - the difference is the presence of denominator terms. The expression for the variance of the PDWIS estimator can be similarly obtained as:

$$
Var[\hat{Q}^{DecPDWIS}_{\pi_b}] = \sum_{n=1}^{N} \sum_{t=0}^{T} \sum_{t^{\prime}=0}^{T} \frac{\gamma^{t+t^{\prime}}}{N^2} \Bigg( \frac{\mathbb{E}_{\pi_b} \Big[ \rho_{0:t}^{(n)} \cdot \rho_{0:t^{\prime}}^{(n)} \cdot r(s_{t}^{(n)},a_{t}^{(n)}) \cdot r(s_{t^{\prime}}^{(n)},a_{t^{\prime}}^{(n)}) \Big]}{\sum_{n=1}^{N} \sum_{t=0}^{T} \sum_{t^{\prime}=0}^{T} \gamma^{t + t^{\prime}} \cdot \mathbb{E}_{\pi_b} \Big[ \rho_{0:t}^{(n)} \cdot \rho_{0:t^{\prime}}^{(n)} \Big]} 
$$

$$
- \ \frac{\mathbb{E}_{\pi_b} \Big[ \rho_{0:t}^{(n)} \cdot r(s_{t}^{(n)},a_{t}^{(n)}) \Big] \mathbb{E}_{\pi_b} \Big[ \rho_{0:t^{\prime}}^{(n)} \cdot r(s_{t^{\prime}}^{(n)},a_{t^{\prime}}^{(n)}) \Big]}{\sum_{n=1}^{N} \sum_{t=0}^{T} \sum_{t^{\prime}=0}^{T} \gamma^{t + t^{\prime}} \cdot \mathbb{E}_{\pi_b} \Big[ \rho_{0:t}^{(n)} \Big] \cdot \mathbb{E}_{\pi_b} \Big[ \rho_{0:t^{\prime}}^{(n)} \Big]} \Bigg)
$$

We have already shown in the variance comparison between decomposed and non-decomposed IS in part \ref{sec:vardev-decomposed-IS} that, since the expected value of any IS ratio is 1, then $\mathbb{E}_{\pi_b} \Big[ \rho_{0:t}^{(n)} \cdot \rho_{0:t^{\prime}}^{(n)} \cdot r(s_{t}^{(n)},a_{t}^{(n)}) \cdot r(s_{t^{\prime}}^{(n)},a_{t^{\prime}}^{(n)}) \Big] \ge \sum_{d=1}^{D} \mathbb{E}_{\pi_b} \Big[ \rho_{0:t}^{(n),d} \cdot \rho_{0:t^{\prime}}^{(n),d} \cdot r^d(z_{t}^{(n),d},a_{t}^{(n),d}) \cdot r^{d}(z_{t^{\prime}}^{(n), d},a_{t^{\prime}}^{(n),d}) \Big]$. We also showed that $\mathbb{E}_{\pi_b} \Big[ \rho_{0:t}^{(n)} \cdot r(s_{t}^{(n)},a_{t}^{(n)}) \Big] \mathbb{E}_{\pi_b} \Big[ \rho_{0:t^{\prime}}^{(n)} \cdot r(s_{t^{\prime}}^{(n)},a_{t^{\prime}}^{(n)}) \Big] = \sum_{d=1}^{D} \mathbb{E}_{\pi_b} \Big[ \rho_{0:t}^{(n),d} \cdot r^d(z_{t}^{(n),d},a_{t}^{(n),d}) \Big] \mathbb{E}_{\pi_b} \Big[ \rho_{0:t^{\prime}}^{(n),d} \cdot r^{d}(z_{t^{\prime}}^{(n), d},a_{t^{\prime}}^{(n),d}) \Big]$. This, so far, has allowed us to compare the numerators of the corresponding pairs of terms in the variance expressions of the PDWIS estimators. 

What remains is the denominators. Since we have assumed $\mathbb{E}_{\pi_b} \Big[ \rho_{0:t}^{(n)} \Big] = \mathbb{E}_{\pi_b} \Big[ \rho_{0:t}^{(n),d} \Big] = 1$, this means that $\sum_{n=1}^{N} \sum_{t=0}^{T} \sum_{t^{\prime}=0}^{T} \gamma^{t + t^{\prime}} \cdot \mathbb{E}_{\pi_b} \Big[ \rho_{0:t}^{(n)} \Big] \cdot \mathbb{E}_{\pi_b} \Big[ \rho_{0:t^{\prime}}^{(n)} \Big] = \sum_{n=1}^{N} \sum_{t=0}^{T} \sum_{t^{\prime}=0}^{T} \gamma^{t + t^{\prime}} \cdot \mathbb{E}_{\pi_b} \Big[ \rho_{0:t}^{(n),d} \Big] \cdot \mathbb{E}_{\pi_b} \Big[ \rho_{0:t^{\prime}}^{(n),d} \Big]$ i.e. the denominators of the second terms in each variance expression are equal. This only leaves the denominators of the first terms to compare.

We can rewrite $\sum_{n=1}^{N} \sum_{t=0}^{T} \sum_{t^{\prime}=0}^{T} \gamma^{t + t^{\prime}} \cdot \mathbb{E}_{\pi_b} \Big[ \rho_{0:t}^{(n)} \cdot \rho_{0:t^{\prime}}^{(n)} \Big]$ as $\sum_{n=1}^{N} \sum_{t=0}^{T} \sum_{t^{\prime}=0}^{T} \gamma^{t + t^{\prime}} \cdot \mathbb{E}_{\pi_b} \Big[ \big( \prod_{d=1}^D \rho_{0:t}^{(n),d} \big) \cdot \big( \prod_{d=1}^D \rho_{0:t^{\prime}}^{(n),d} \big) \Big]$. Thus, we guarantee that the decomposed PDWIS estimator has at most the variance of the PDWIS estimator if $\mathbb{E}_{\pi_b} \Big[ \big( \prod_{d=1}^D \rho_{0:t}^{(n),d} \big) \cdot \big( \prod_{d=1}^D \rho_{0:t^{\prime}}^{(n),d} \big) \Big] \le \mathbb{E}_{\pi_b} \Big[ \rho_{0:t}^{(n),d} \cdot \rho_{0:t^{\prime}}^{(n),d} \Big]$. This could be the case if $Cov \Big(\big( \prod_{d^{\prime} \ne d} \rho_{0:t}^{(n),d^{\prime}} \big) \cdot \big( \prod_{d^{\prime} \ne d} \rho_{0:t^{\prime}}^{(n),d^{\prime}} \big) , \ \rho_{0:t}^{(n),d} \cdot \rho_{0:t^{\prime}}^{(n),d}\Big) = \mathbb{E}_{\pi_b} \Big[ \big( \prod_{d=1}^D \rho_{0:t}^{(n),d} \big) \cdot \big( \prod_{d=1}^D \rho_{0:t^{\prime}}^{(n),d} \big) \Big] - \mathbb{E} \Big[ \prod_{d^{\prime} \ne d} \rho_{0:t}^{(n),d^{\prime}} \big) \cdot \big( \prod_{d^{\prime} \ne d} \rho_{0:t^{\prime}}^{(n),d^{\prime}} \big) \Big] \cdot \mathbb{E} \Big[ \rho_{0:t}^{(n),d} \cdot \rho_{0:t^{\prime}}^{(n),d} \Big] \le 0$. This is in fact related to the condition in Equation \ref{eq:ratio-correlation} in Assumption \ref{ass:variance-bounds}. We additionally require that $Cov \Big(\big( \prod_{d^{\prime} \ne d} \rho_{0:t}^{(n),d^{\prime}} \big), \ \big( \prod_{d^{\prime} \ne d} \rho_{0:t^{\prime}}^{(n),d^{\prime}} \big)\Big) \ge 0$, which is related to condition \ref{eq:same-ratio-correlation} in Assumption \ref{ass:PDWIS-variance-bounds} to ensure that $\mathbb{E} \Big[ \prod_{d^{\prime} \ne d} \rho_{0:t}^{(n),d^{\prime}} \big) \cdot \big( \prod_{d^{\prime} \ne d} \rho_{0:t^{\prime}}^{(n),d^{\prime}} \big) \Big] \ge 1$. This is because  $Cov \Big(\big( \prod_{d^{\prime} \ne d} \rho_{0:t}^{(n),d^{\prime}} \big), \ \big( \prod_{d^{\prime} \ne d} \rho_{0:t^{\prime}}^{(n),d^{\prime}} \big)\Big) = \mathbb{E} \Big[ \prod_{d^{\prime} \ne d} \rho_{0:t}^{(n),d^{\prime}} \big) \cdot \big( \prod_{d^{\prime} \ne d} \rho_{0:t^{\prime}}^{(n),d^{\prime}} \big) \Big] - \mathbb{E} \Big[ \prod_{d^{\prime} \ne d} \rho_{0:t}^{(n),d^{\prime}} \Big] \cdot \Big[ \prod_{d^{\prime} \ne d} \rho_{0:t^{\prime}}^{(n),d^{\prime}} \Big] = \mathbb{E} \Big[ \prod_{d^{\prime} \ne d} \rho_{0:t}^{(n),d^{\prime}} \big) \cdot \big( \prod_{d^{\prime} \ne d} \rho_{0:t^{\prime}}^{(n),d^{\prime}} \big) \Big] - 1$.

Thus, when necessary conditions are satisfied, we can show that the decomposed PDWIS estimator has lower or equal variance, compared to non-decomposed PDWIS.

\section{Grouping Action Spaces Together to Create Unbiased Estimators}
\label{sec:grouping-action-spaces}

Here we illustrate how actions may be grouped in cases where equations \ref{eq:policy-factorisation} and \ref{eq:reward-factorisation} do not apply. As a running example, we take the decomposed IS estimator, but the same transformations can be applied to the decomposed PDIS and PDWIS estimators.

When Equation \ref{eq:policy-factorisation} does not apply i.e. the policy probabilities cannot be product-wise separated for the current action space factorisation, we need to adjust the IS ratios. We group the action space indices $d \in \{1,2...D\}$ into sets $S_{k,e}$ for policy $\pi_e$ , such that $k \in \{1, 2 ... K_e\}$, and $S_{k,b}$ for policy $\pi_b$, such that $k \in \{1, 2 ... K_b\}$. We can then express Equation \ref{eq:policy-factorisation} as:

\begin{equation}
\label{eq:new-policy-factorisation}
     \pi(a|s) = \prod_{k=1}^{K} \pi^k(a^k|z^k) 
 \end{equation}

where $a^k$ is a composite action from the union of action spaces in set $S_k$, $\pi^k$ is a policy on actions from this composite space and $z^k$ is a a corresponding state abstraction. The expression for the decomposed IS estimator is similar to before:

\begin{equation}
    \tilde{Q}^{DecIS}_{\pi_e} = \sum_{d=1}^{D} \frac{1}{N} \sum_{n=1}^{N} \rho_{0:T}^{(n),d} \sum_{t=0}^{T} \gamma^t \cdot r^d(z_{t}^{(n),d},a_{t}^{(n),d})
\end{equation}

However $\rho_{0:T}^{(n),d}$ is defined as follows if $d \in S_{k_1, e}$ and $d \in S_{k_2, b}$:

$$
    \rho_{0:T}^{(n), d} = \prod_{t=0}^{T} \frac{\pi^{k_1}_e(a^{k_1}_t|z^{k_1}_t)}{\pi^{k_2}_b(a^{k_2}_t|z^{k_2}_t)}
$$

This version of the decomposed IS estimator is unbiased provided that $S_{k,e} = S_{k,b}, \ \forall k$ i.e. we have the exact same groupings of action spaces for $\pi_b$ and $\pi_e$. This is because in this case, we can also group the rewards as:

$$
r^k(z^k, a^k) = \sum_{d \in S_{k,e}} r^d(z^d, a^d)
$$

and thus we are satisfying Theorem \ref{theorem-1-shengpu paper} with $K_b = K_e$ factored action spaces. The expression for the estimator would be:

\begin{equation}
    \tilde{Q}^{DecIS}_{\pi_e} = \sum_{k=1}^{K_e} \frac{1}{N} \sum_{n=1}^{N} \rho_{0:T}^{(n),d} \sum_{t=0}^{T} \gamma^t \cdot r^k(z_{t}^{(n),k},a_{t}^{(n),k})
\end{equation}

Since the estimator is a decomposed estimator with theorem \ref{theorem-1-shengpu paper} satisfied, by theorem \ref{theorem-lower-variance-factored-estimator} it has at most the variance of the non-decomposed IS estimator. Even if the condition $S_{k,e} = S_{k,b}, \ \forall k$ is not met, and thus theorem \ref{theorem-1-shengpu paper} is not satisfied, the decomposed nature of of the estimator still allows us to guarantee that it will have at most the variance of the corresponding non-decomposed estimator. The proof for this is similar to part \ref{sec:dec-var} and is omitted.

When Equation \ref{eq:reward-factorisation} does not apply either i.e. the rewards cannot be decomposed as a summation, we would again group the the action spaces $d \in \{1,2...D\}$ into sets $S_{k,r}$ such that $k \in \{1, 2 ... K_r\}$. This allows Equation \ref{eq:reward-factorisation} to be expressed as:

\begin{equation}
 \label{eq:new-reward-factorisation}
     r(s,a) = \sum_{k=1}^{K_r} r^k(z^k, a^k), 
 \end{equation}

The new expression for the decomposed IS estimator would be:

\begin{equation}
    \tilde{Q}^{DecIS}_{\pi_e} = \sum_{k=1}^{K_r} \frac{1}{N} \sum_{n=1}^{N} \rho_{0:T}^{(n),d} \sum_{t=0}^{T} \gamma^t \cdot r^k(z_{t}^{(n),k},a_{t}^{(n),k})
\end{equation}

where $\rho_{0:T}^{(n),d}$ is as defined above. It can again be shown that if $S_{k,r} = S_{k,e} = S_{k,b}, \ \forall k$, then the above estimator is unbiased, because we can satisfy Theorem \ref{theorem-1-shengpu paper} with $K_b = K_e = K_r$ factored action spaces. Again, since it is a decomposed estimator, we can show that it has at most the variance of the IS estimator, regardless of whether the condition $S_{k,r} = S_{k,e} = S_{k,b}, \ \forall k$ is met. 

We can view the grouping sets $S_{k,r}, S_{k,e}, S_{k,b}$ as the merging of action spaces to form new larger ones that satisfy Theorem \ref{theorem-1-shengpu paper} - we can define how exactly this grouping is done. In the most general case that $S_{1,r} = S_{1,e} = S_{1,b} = \mathcal{A}$ i.e. we group all the spaces together into one space, we get the original IS estimator back, and then the bias and variance of our "decomposed estimator" is equal to that of the original non-decomposed IS estimator. 

\newpage

\section{Diagram of MDP 1 and its Factorisation}
\label{sec:diag-MDP-1}

In Figure \ref{fig:diag-MDP-1}, states are represented as labelled circles, while the actions are represented by directed arrows. For example, an arrow pointing up and right represents the action $up\_right$ and an arrow pointing down and right represents $down\_right$. There is only one transition possible, from $state$ to $terminal$, and only the action can vary.

\begin{figure}[h!]
    \centering
    \begin{tikzpicture}[>={Stealth[inset=0pt,length=6pt,angle'=40,round]}, scale=0.01]
\node[state] (s0) {state};
\node[state, accepting, left of=s0, xshift=-1.8cm, yshift=-1.8cm] (s1) {terminal};
\node[state, accepting, right of=s0, xshift=1.8cm, yshift=-1.8cm] (s2) {terminal};
\node[state, accepting, right of=s0, xshift=1.8cm, yshift=1.8cm] (s3) {terminal};
\node[state, accepting, left of=s0, xshift=-1.8cm, yshift=1.8cm] (s4) {terminal};

\draw[->] (s0) edge[below] node{$1 + \alpha + \beta$} (s3)
(s0) edge[above] node{1} (s2)
(s0) edge[above] node{$\alpha$} (s4)
(s0) edge[above] node{0} (s1);
\end{tikzpicture}
\caption{Factored MDP-1 in first factored action space (horizontal).}
\label{fig:diag-MDP-1}
\end{figure}

As discussed, throughout experiments $\alpha = 1$. Furthermore, unless explicitly stated otherwise $\beta = 0$. The only time $\beta$ is varied was in a set of experiments explicitly designed to observe the effect of varying $\beta$, as $\beta \ne 0$ results in Theorem \ref{theorem-1-shengpu paper} no longer applying to the MDP, due to violation of the condition in Equation \ref{eq:reward-factorisation}. Thus, in the default configuration, $r(state, \ up\_right) = 2$, $r(state, \ up\_left) = 1$, $r(state, \ down\_right) = 1$ and $r(state, \ down\_left) = 0$. The MDP action space can be factored into $\mathcal{A}^1 = \{left, right\}$ and $\mathcal{A}^2 = \{up, down\}$. The MDP based on $\mathcal{A}^1$, assuming $\beta = 0$, is in Figure \ref{fig:diag-MDP-1-factored-1}. Here, there is no need to abstract the states according to the action as every action fully affects the state.

\begin{figure}[h!]
    \centering
    \begin{tikzpicture}[>={Stealth[inset=0pt,length=6pt,angle'=40,round]}, scale=0.01]
\node[state] (s0) {state};
\node[state, accepting, right of=s0, xshift=1cm] (s1) {terminal};
\node[state, accepting, left of=s0, xshift=-1cm] (s2) {terminal};

\draw[->] (s0) edge[above] node{1} (s1)
(s0) edge[above] node{0} (s2);
\end{tikzpicture}
\caption{Factored MDP-1 in first factored action space (horizontal).}
\label{fig:diag-MDP-1-factored-1}
\end{figure}

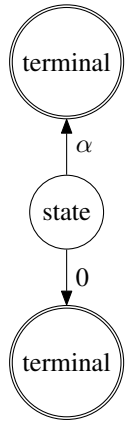
\begin{figure}[h!]
    \centering
    \begin{tikzpicture}[>={Stealth[inset=0pt,length=6pt,angle'=40,round]}, scale=0.01]
\node[state] (s0) {state};
\node[state, accepting, above of=s0, yshift=1cm] (s1) {terminal};
\node[state, accepting, below of=s0, yshift=-1cm] (s2) {terminal};

\draw[->] (s0) edge[right] node{$\alpha$} (s1)
(s0) edge[right] node{0} (s2);
\end{tikzpicture}
\caption{Factored MDP-1 in second factored action space (vertical).}
\label{fig:diag-MDP-1-factored-2}
\end{figure}

The MDP based on $\mathcal{A}^2$, assuming $\beta = 0$, is in Figure \ref{fig:diag-MDP-1-factored-2}. It is clear that putting the actions from corresponding states together from the factored MDPs together and summing the coinciding rewards would give the overall MDP - this is possible because Theorem \ref{theorem-1-shengpu paper} is satisfied. This would not be the case if $\beta \ne 0$ - a situation which we investigate in the main paper.

\newpage

\section{Diagram of MDP 2 and its Factorisation}
\label{sec:diag-MDP-2}

\begin{figure}[h!]
    \centering
    \begin{tikzpicture}[>={Stealth[inset=0pt,length=6pt,angle'=40,round]}, scale=1]
\node[state] (s0) {0,0};
\node[state, right of=s0, xshift=1.5cm] (s2) {1,0};
\node[state, above of=s2, yshift=1.5cm] (s3) {1,1};
\node[state, above of=s0, yshift=1.5cm] (s1) {0,1};

\draw[->] (s0) edge[bend angle=45, bend left, left] node{1} (s1)
(s0) edge[bend angle=45, bend right, below] node{1} (s2)
(s0) edge[bend angle=30, bend left, above] node{2} (s3)
(s2) edge[bend left, above] node{0} (s0)
(s1) edge[bend right, right] node{0} (s0)
(s1) edge[bend angle=45, bend left, above] node{1} (s3)
(s1) edge[bend angle=30, bend left, above] node{1} (s2)
(s2) edge[bend angle=30, bend left, above] node{1} (s1)
(s3) edge[bend right, below] node{0} (s1)
(s2) edge[bend angle=45, bend right, right] node{1} (s3)
(s3) edge[bend left, left] node{0} (s2)
(s3) edge[bend angle=30, bend left, below] node{0} (s0);
\draw[->] (s0) edge[min distance=7mm, loop left, out=-135, in=-110] node{0} (s0);
\draw[->] (s1) edge[min distance=7mm, loop left, out=135, in=160] node{-1} (s1);
\draw[->] (s3) edge[min distance=7mm, loop right, out=45, in=70] node{-2} (s3);
\draw[->] (s2) edge[min distance=7mm, loop right, out=-45, in=-70] node{-1} (s2);
\end{tikzpicture}

    \caption{MDP-2 with corresponding rewards used in experiments.}
    \label{fig:diag-MDP-2}
\end{figure}
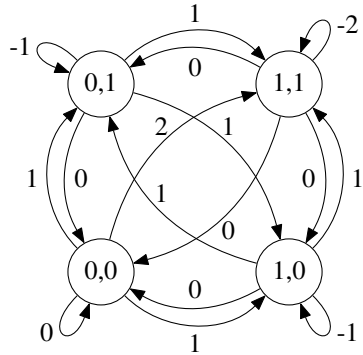

In Figure \ref{fig:diag-MDP-2}, the states are labelled circles, while actions from states are arrows that leave the state in a particular direction. For example, an arrow leaving a state in the upward-right direction represents the action $up\_right$ from that state. Each arrow is annotated with a number representing the reward of the \textit{(state, action)} pair it represents i.e. $r(s,a)$. To clarify in case there is ambiguity for the reader the rewards are: $r(0,0, \ up\_right) = 2$, $r(0,0, \ up\_left) = 1$, $r(0,0, \ down\_right) = 1$, $r(0,0, \ down\_left) = 0$, $r(0,1, \ up\_right) = 1$, $r(0,1, \ up\_left) = -1$, $r(0,1, \ down\_right) = 1$, $r(0,1, \ down\_left) = 0$, $r(1,0, \ up\_right) = 1$, $r(1,0, \ up\_left) = 1$, $r(1,0, \ down\_right) = -1$, $r(1,0, \ down\_left) = 0$, $r(1,1, \ up\_right) = -2$, $r(1,1, \ up\_left) = 0$, $r(1,1, \ down\_right) = 0$, $r(1,1, \ down\_left) = 0$.

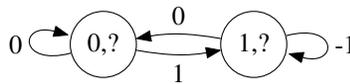
\begin{figure}[h!]
    \centering
    \begin{tikzpicture}[>={Stealth[inset=0pt,length=6pt,angle'=40,round]}, scale=0.3]
\node[state] (s0) {0,?};
\node[state, right of=s0, xshift=1cm] (s1) {1,?};

\draw[->] (s0) edge[bend angle=10, bend right, below] node{1} (s1)
(s0) edge[loop left] node{0} (s0)
(s1) edge[bend angle=10, bend right, above] node{0} (s0)
(s1) edge[loop right] node{-1} (s1);
\end{tikzpicture}

\caption{Factored MDP-2 in first factored action space (horizontal).}
\label{fig:diag-MDP-2-factored-1}
\end{figure}

This MDP has no terminal states, which enables the agent to follow trajectories of unlimited length. The MDP action space may be factored into two spaces: $\mathcal{A}^1 = \{left, \ right\}$ and $\mathcal{A}^2 = \{up, \ down\}$. The MDP based on $\mathcal{A}^1$ is in Figure \ref{fig:diag-MDP-2-factored-1}. If Figure \ref{fig:diag-MDP-2} is observed, it is seen that only the first number in the state label is relevant to the actions in $\mathcal{A}^1$; the other number is independent. Thus every state can be abstracted to only its first number e.g. $0,0$ becomes $0,?$ and $1,0$ becomes $1,?$.

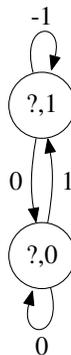
\begin{figure}[h!]
    \centering

\begin{tikzpicture}[>={Stealth[inset=0pt,length=6pt,angle'=40,round]}, scale=0.3]
\node[state] (s0) {?,0};
\node[state, above of=s0, yshift=1cm] (s1) {?,1};

\draw[->] (s0) edge[bend angle=10, bend right, right] node{1} (s1)
(s0) edge[loop below] node{0} (s0)
(s1) edge[bend angle=10, bend right, left] node{0} (s0)
(s1) edge[loop above] node{-1} (s1);
\end{tikzpicture}

\caption{Factored MDP-2 in second factored action space (vertical).}
\label{fig:diag-MDP-2-factored-2}
\end{figure}

The MDP based on $\mathcal{A}^2$ is in Figure \ref{fig:diag-MDP-2-factored-2}. Here, only the second number in the state label is relevant to the actions being taken. Thus every state is abstracted to its second number e.g. $0,0$ becomes $?,0$ and $0,1$ becomes $?,1$. It is clear to see that combinatorially putting the actions and state abstractions of the factored MDPs together and summing the coinciding rewards would give the overall MDP - this is possible because Theorem \ref{theorem-1-shengpu paper} is satisfied.

\newpage

\section{Policies Used in Experiments}

Throughout experiments, the policy divergence, as defined in Equation \ref{eq:policy-divergence-definition}, is varied by specifically designing the behaviour policy $\pi_b$ and evaluation policy $\pi_e$. In this section, we discuss the exact policies utilised for each MDP, in order to generate each recorded policy divergence value. Note that all policy divergence values used throughout this paper have been rounded to 2 decimal places.

\subsection{In MDP 1}
\label{subsec:policies-MDP1}

The states of MDP 1 are $\mathcal{S} = \{ state, terminal \}$. The action space is $\mathcal{A} = \{ up\_right, \ up\_left, \ down\_right, \ down\_left\}$. There is only one transition, hence any policy would only be defined at $state$.

The first factored action space $\mathcal{A}^1 = \{ right, \ left \}$, while if $\phi_d: \mathcal{S} \to \mathcal{Z}^d$ maps a state to its abstraction with respect to the $d^{th}$ factored action space, then $\phi^1(state) = state$ and $\phi^1(terminal) = terminal$.

The second factored action space $\mathcal{A}^2 = \{ up, \ down \}$, while $\phi^2(state) = state$ and $\phi^2(terminal) = terminal$.

\subsubsection{Policy Divergence = 1.44}

\textbf{Behaviour Policy:}

$\pi_b(up\_right|state) = \pi_b(up\_left|state) = \pi_b(down\_right|state) = \pi_b(down\_left|state) = 0.25$.

With respect to $\mathcal{A}^1$, $\pi_b^1(right|state) = \pi_b^1(left|state) = 0.5$.

With respect to $\mathcal{A}^2$, $\pi_b^2(up|state) = \pi_b^2(down|state) = 0.5$.

\textbf{Evaluation Policy:}

$\pi_e(up\_right|state) = 0.36, \ \pi_e(up\_left|state) = 0.24, \ \pi_e(down\_right|state) = 0.24, \ \pi_e(down\_right|state) = 0.16$.

With respect to $\mathcal{A}^1$, $\pi^1_e(right|state) = 0.6, \ \pi^1_e(left|state) = 0.4$.

With respect to $\mathcal{A}^2$, $\pi^2_e(up|state) = 0.6, \ \pi^2_e(down|state) = 0.4$.

\subsubsection{Policy Divergence = 2.56}

\textbf{Behaviour Policy:}

$\pi_b(up\_right|state) = \pi_b(up\_left|state) = \pi_b(down\_right|state) = \pi_b(down\_left|state) = 0.25$.

With respect to $\mathcal{A}^1$, $\pi_b^1(right|state) = \pi_b^1(left|state) = 0.5$.

With respect to $\mathcal{A}^2$, $\pi_b^2(up|state) = \pi_b^2(down|state) = 0.5$.

\textbf{Evaluation Policy:}

$\pi_e(up\_right|state) = 0.64, \ \pi_e(up\_left|state) = 0.16, \ \pi_e(down\_right|state) = 0.16, \ \pi_e(down\_right|state) = 0.04$.

With respect to $\mathcal{A}^1$, $\pi^1_e(right|state) = 0.8, \ \pi^1_e(left|state) = 0.2$.

With respect to $\mathcal{A}^2$, $\pi^2_e(up|state) = 0.8, \ \pi^2_e(down|state) = 0.2$.

\subsubsection{Policy Divergence = 3.61}

\textbf{Behaviour Policy:}

$\pi_b(up\_right|state) = \pi_b(up\_left|state) = \pi_b(down\_right|state) = \pi_b(down\_left|state) = 0.25$.

With respect to $\mathcal{A}^1$, $\pi_b^1(right|state) = \pi_b^1(left|state) = 0.5$.

With respect to $\mathcal{A}^2$, $\pi_b^2(up|state) = \pi_b^2(down|state) = 0.5$.

\textbf{Evaluation Policy:}

$\pi_e(up\_right|state) = 0.9025, \ \pi_e(up\_left|state) = 0.0475, \ \pi_e(down\_right|state) = 0.0475, \ \pi_e(down\_right|state) = 0.0025$.

With respect to $\mathcal{A}^1$, $\pi^1_e(right|state) = 0.95, \ \pi^1_e(left|state) = 0.05$.

With respect to $\mathcal{A}^2$, $\pi^2_e(up|state) = 0.95, \ \pi^2_e(down|state) = 0.05$.

\subsubsection{Policy Divergence = 4.46}

\textbf{Behaviour Policy:}

$\pi_b(up\_right|state) = 0.2025, \ \pi_b(up\_left|state) = 0.2475, \ \pi_b(down\_right|state) = 0.2475, \ \pi_b(down\_left|state) = 0.3025$.

With respect to $\mathcal{A}^1$, $\pi_b^1(right|state) = 0.45, \ \pi_b^1(left|state) = 0.55$.

With respect to $\mathcal{A}^2$, $\pi_b^2(up|state) = 0.45, \ \pi_b^2(down|state) = 0.55$.

\textbf{Evaluation Policy:}

$\pi_e(up\_right|state) = 0.9025, \ \pi_e(up\_left|state) = 0.0475, \ \pi_e(down\_right|state) = 0.0475, \ \pi_e(down\_right|state) = 0.0025$.

With respect to $\mathcal{A}^1$, $\pi^1_e(right|state) = 0.95, \ \pi^1_e(left|state) = 0.05$.

With respect to $\mathcal{A}^2$, $\pi^2_e(up|state) = 0.95, \ \pi^2_e(down|state) = 0.05$.

\subsubsection{Policy Divergence = 5.64}

\textbf{Behaviour Policy:}

$\pi_b(up\_right|state) = 0.16, \ \pi_b(up\_left|state) = 0.24, \ \pi_b(down\_right|state) = 0.24, \ \pi_b(down\_left|state) = 0.36$.

With respect to $\mathcal{A}^1$, $\pi_b^1(right|state) = 0.4, \ \pi_b^1(left|state) = 0.6$.

With respect to $\mathcal{A}^2$, $\pi_b^2(up|state) = 0.4, \ \pi_b^2(down|state) = 0.6$.

\textbf{Evaluation Policy:}

$\pi_e(up\_right|state) = 0.9025, \ \pi_e(up\_left|state) = 0.0475, \ \pi_e(down\_right|state) = 0.0475, \ \pi_e(down\_right|state) = 0.0025$.

With respect to $\mathcal{A}^1$, $\pi^1_e(right|state) = 0.95, \ \pi^1_e(left|state) = 0.05$.

With respect to $\mathcal{A}^2$, $\pi^2_e(up|state) = 0.95, \ \pi^2_e(down|state) = 0.05$.

\subsubsection{Policy Divergence = 10.03}

\textbf{Behaviour Policy:}

$\pi_b(up\_right|state) = 0.09, \ \pi_b(up\_left|state) = 0.21, \ \pi_b(down\_right|state) = 0.21, \ \pi_b(down\_left|state) = 0.49$.

With respect to $\mathcal{A}^1$, $\pi_b^1(right|state) = 0.3, \ \pi_b^1(left|state) = 0.7$.

With respect to $\mathcal{A}^2$, $\pi_b^2(up|state) = 0.3, \ \pi_b^2(down|state) = 0.7$.

\textbf{Evaluation Policy:}

$\pi_e(up\_right|state) = 0.9025, \ \pi_e(up\_left|state) = 0.0475, \ \pi_e(down\_right|state) = 0.0475, \ \pi_e(down\_right|state) = 0.0025$.

With respect to $\mathcal{A}^1$, $\pi^1_e(right|state) = 0.95, \ \pi^1_e(left|state) = 0.05$.

With respect to $\mathcal{A}^2$, $\pi^2_e(up|state) = 0.95, \ \pi^2_e(down|state) = 0.05$.

\subsubsection{Policy Divergence = 22.56}

\textbf{Behaviour Policy:}

$\pi_b(up\_right|state) = 0.04, \ \pi_b(up\_left|state) = 0.16, \ \pi_b(down\_right|state) = 0.16, \ \pi_b(down\_left|state) = 0.64$.

With respect to $\mathcal{A}^1$, $\pi_b^1(right|state) = 0.2, \ \pi_b^1(left|state) = 0.8$.

With respect to $\mathcal{A}^2$, $\pi_b^2(up|state) = 0.2, \ \pi_b^2(down|state) = 0.8$.

\textbf{Evaluation Policy:}

$\pi_e(up\_right|state) = 0.9025, \ \pi_e(up\_left|state) = 0.0475, \ \pi_e(down\_right|state) = 0.0475, \ \pi_e(down\_right|state) = 0.0025$.

With respect to $\mathcal{A}^1$, $\pi^1_e(right|state) = 0.95, \ \pi^1_e(left|state) = 0.05$.

With respect to $\mathcal{A}^2$, $\pi^2_e(up|state) = 0.95, \ \pi^2_e(down|state) = 0.05$.

\subsubsection{Policy Divergence = 90.25}

\textbf{Behaviour Policy:}

$\pi_b(up\_right|state) = 0.01, \ \pi_b(up\_left|state) = 0.09, \ \pi_b(down\_right|state) = 0.09, \ \pi_b(down\_left|state) = 0.81$.

With respect to $\mathcal{A}^1$, $\pi_b^1(right|state) = 0.1, \ \pi_b^1(left|state) = 0.9$.

With respect to $\mathcal{A}^2$, $\pi_b^2(up|state) = 0.1, \ \pi_b^2(down|state) = 0.9$.

\textbf{Evaluation Policy:}

$\pi_e(up\_right|state) = 0.9025, \ \pi_e(up\_left|state) = 0.0475, \ \pi_e(down\_right|state) = 0.0475, \ \pi_e(down\_right|state) = 0.0025$.

With respect to $\mathcal{A}^1$, $\pi^1_e(right|state) = 0.95, \ \pi^1_e(left|state) = 0.05$.

With respect to $\mathcal{A}^2$, $\pi^2_e(up|state) = 0.95, \ \pi^2_e(down|state) = 0.05$.

\subsubsection{Policy Divergence = 361.0}

\textbf{Behaviour Policy:}

$\pi_b(up\_right|state) = 0.0025, \ \pi_b(up\_left|state) = 0.0475, \ \pi_b(down\_right|state) = 0.0475, \ \pi_b(down\_left|state) = 0.9025$.

With respect to $\mathcal{A}^1$, $\pi_b^1(right|state) = 0.05, \ \pi_b^1(left|state) = 0.95$.

With respect to $\mathcal{A}^2$, $\pi_b^2(up|state) = 0.05, \ \pi_b^2(down|state) = 0.95$.

\textbf{Evaluation Policy:}

$\pi_e(up\_right|state) = 0.9025, \ \pi_e(up\_left|state) = 0.0475, \ \pi_e(down\_right|state) = 0.0475, \ \pi_e(down\_right|state) = 0.0025$.

With respect to $\mathcal{A}^1$, $\pi^1_e(right|state) = 0.95, \ \pi^1_e(left|state) = 0.05$.

With respect to $\mathcal{A}^2$, $\pi^2_e(up|state) = 0.95, \ \pi^2_e(down|state) = 0.05$.

\newpage

\subsection{In MDP 2}
\label{subsec:policies-MDP2}

The states of MDP 1 are $\mathcal{S} = \{ 0,0, \ 0,1, \ 1,0, \ 1,1 \}$. The policy must be defined at all these states. The action space is $\mathcal{A} = \{ up\_right, \ up\_left, \ down\_right, \ down\_left\}$. There can be unlimited transitions, we denote the number of transitions i.e. trajectory length as $T$. The policy divergence is thus represented as $P^T$, where $P$ is the policy divergence for 1 transition.

The first factored action space $\mathcal{A}^1 = \{ right, \ left \}$, while if $\phi_d: \mathcal{S} \to \mathcal{Z}_d$ maps a state to its abstraction with respect to the $d^{th}$ factored action space, then $\phi^1(0,0) = \phi^1(0,1) = 0,?$ and $\phi^1(1,0) = \phi^1(1,1) = 1,?$.

The second factored action space $\mathcal{A}^2 = \{ up, \ down \}$, while $\phi^2(0,0) = \phi^2(1,0) = ?,0$ and $\phi^2(0,1) = \phi^2(1,1) = ?,1$.

\subsubsection{Policy Divergence = $1.44^T$}

\textbf{Behaviour Policy:}

$\pi_b(up\_right|s) = \pi_b(up\_left|s) = \pi_b(down\_right|s) = \pi_b(down\_left|s) = 0.25, \ \forall \ s \in \mathcal{S}$.

With respect to $\mathcal{A}^1$, $\pi_b^1(right|z^1) = \pi_b^1(left|z^1) = 0.5, \ \forall \ z^1 \in \mathcal{Z}^1$.

With respect to $\mathcal{A}^2$, $\pi_b^2(up|z^2) = \pi_b^2(down|z^2) = 0.5, \ \forall \ z^2 \in \mathcal{Z}^2$.

\textbf{Evaluation Policy:}

$\pi_e(up\_right|s) = 0.36, \ \pi_e(up\_left|s) = 0.24, \ \pi_e(down\_right|s) = 0.24, \ \pi_e(down\_right|s) = 0.16, \ \forall \ s \in \mathcal{S}$.

With respect to $\mathcal{A}^1$, $\pi^1_e(right|z^1) = 0.6, \ \pi^1_e(left|z^1) = 0.4, \ \forall \ z^1 \in \mathcal{Z}^1$.

With respect to $\mathcal{A}^2$, $\pi^2_e(up|z^2) = 0.6, \ \pi^2_e(down|z^2) = 0.4, \ \forall \ z^2 \in \mathcal{Z}^2$.

\subsubsection{Policy Divergence = $2.56^T$}

\textbf{Behaviour Policy:}

$\pi_b(up\_right|s) = \pi_b(up\_left|s) = \pi_b(down\_right|s) = \pi_b(down\_left|s) = 0.25, \ \forall \ s \in \mathcal{S}$.

With respect to $\mathcal{A}^1$, $\pi_b^1(right|z^1) = \pi_b^1(left|z^1) = 0.5, \ \forall \ z^1 \in \mathcal{Z}^1$.

With respect to $\mathcal{A}^2$, $\pi_b^2(up|z^2) = \pi_b^2(down|z^2) = 0.5, \ \forall \ z^2 \in \mathcal{Z}^2$.

\textbf{Evaluation Policy:}

$\pi_e(up\_right|s) = 0.64, \ \pi_e(up\_left|s) = 0.16, \ \pi_e(down\_right|s) = 0.16, \ \pi_e(down\_right|s) = 0.04, \ \forall \ s \in \mathcal{S}$.

With respect to $\mathcal{A}^1$, $\pi^1_e(right|z^1) = 0.8, \ \pi^1_e(left|z^1) = 0.2, \ \forall \ z^1 \in \mathcal{Z}^1$.

With respect to $\mathcal{A}^2$, $\pi^2_e(up|z^2) = 0.8, \ \pi^2_e(down|z^2) = 0.2, \ \forall \ z^2 \in \mathcal{Z}^2$.

\subsubsection{Policy Divergence = $3.61^T$}

\textbf{Behaviour Policy:}

$\pi_b(up\_right|s) = \pi_b(up\_left|s) = \pi_b(down\_right|s) = \pi_b(down\_left|s) = 0.25, \ \forall \ s \in \mathcal{S}$.

With respect to $\mathcal{A}^1$, $\pi_b^1(right|z^1) = \pi_b^1(left|z^1) = 0.5, \ \forall \ z^1 \in \mathcal{Z}^1$.

With respect to $\mathcal{A}^2$, $\pi_b^2(up|z^2) = \pi_b^2(down|z^2) = 0.5, \ \forall \ z^2 \in \mathcal{Z}^2$.

\textbf{Evaluation Policy:}

$\pi_e(up\_right|s) = 0.9025, \ \pi_e(up\_left|s) = 0.0475, \ \pi_e(down\_right|s) = 0.0475, \ \pi_e(down\_right|s) = 0.0025, \ \forall \ s \in \mathcal{S}$.

With respect to $\mathcal{A}^1$, $\pi^1_e(right|z^1) = 0.95, \ \pi^1_e(left|z^1) = 0.05, \ \forall \ z^1 \in \mathcal{Z}^1$.

With respect to $\mathcal{A}^2$, $\pi^2_e(up|z^2) = 0.95, \ \pi^2_e(down|z^2) = 0.05, \ \forall \ z^2 \in \mathcal{Z}^2$.

\subsubsection{Policy Divergence = $4.46^T$}

\textbf{Behaviour Policy:}

$\pi_b(up\_right|s) = 0.2025, \ \pi_b(up\_left|s) = 0.2475, \ \pi_b(down\_right|s) = 0.2475, \ \pi_b(down\_left|s) = 0.3025, \ \forall \ s \in \mathcal{S}$.

With respect to $\mathcal{A}^1$, $\pi_b^1(right|z^1) = 0.45, \ \pi_b^1(left|z^1) = 0.55, \ \forall \ z^1 \in \mathcal{Z}^1$.

With respect to $\mathcal{A}^2$, $\pi_b^2(up|z^2) = 0.45, \ \pi_b^2(down|z^2) = 0.55, \ \forall \ z^2 \in \mathcal{Z}^2$.

\textbf{Evaluation Policy:}

$\pi_e(up\_right|s) = 0.9025, \ \pi_e(up\_left|s) = 0.0475, \ \pi_e(down\_right|s) = 0.0475, \ \pi_e(down\_right|s) = 0.0025, \ \forall \ s \in \mathcal{S}$.

With respect to $\mathcal{A}^1$, $\pi^1_e(right|z^1) = 0.95, \ \pi^1_e(left|z^1) = 0.05, \ \forall \ z^1 \in \mathcal{Z}^1$.

With respect to $\mathcal{A}^2$, $\pi^2_e(up|z^2) = 0.95, \ \pi^2_e(down|z^2) = 0.05, \ \forall \ z^2 \in \mathcal{Z}^2$.

\subsubsection{Policy Divergence = $5.64^T$}

\textbf{Behaviour Policy:}

$\pi_b(up\_right|s) = 0.16, \ \pi_b(up\_left|s) = 0.24, \ \pi_b(down\_right|s) = 0.24, \ \pi_b(down\_left|s) = 0.36, \ \forall \ s \in \mathcal{S}$.

With respect to $\mathcal{A}^1$, $\pi_b^1(right|z^1) = 0.4, \ \pi_b^1(left|z^1) = 0.6, \ \forall \ z^1 \in \mathcal{Z}^1$.

With respect to $\mathcal{A}^2$, $\pi_b^2(up|z^2) = 0.4, \ \pi_b^2(down|z^2) = 0.6, \ \forall \ z^2 \in \mathcal{Z}^2$.

\textbf{Evaluation Policy:}

$\pi_e(up\_right|s) = 0.9025, \ \pi_e(up\_left|s) = 0.0475, \ \pi_e(down\_right|s) = 0.0475, \ \pi_e(down\_right|s) = 0.0025, \ \forall \ s \in \mathcal{S}$.

With respect to $\mathcal{A}^1$, $\pi^1_e(right|z^1) = 0.95, \ \pi^1_e(left|z^1) = 0.05, \ \forall \ z^1 \in \mathcal{Z}^1$.

With respect to $\mathcal{A}^2$, $\pi^2_e(up|z^2) = 0.95, \ \pi^2_e(down|z^2) = 0.05, \ \forall \ z^2 \in \mathcal{Z}^2$.

\subsubsection{Policy Divergence = $10.03^T$}

\textbf{Behaviour Policy:}

$\pi_b(up\_right|s) = 0.09, \ \pi_b(up\_left|s) = 0.21, \ \pi_b(down\_right|s) = 0.21, \ \pi_b(down\_left|s) = 0.49, \ \forall \ s \in \mathcal{S}$.

With respect to $\mathcal{A}^1$, $\pi_b^1(right|z^1) = 0.3, \ \pi_b^1(left|z^1) = 0.7, \ \forall \ z^1 \in \mathcal{Z}^1$.

With respect to $\mathcal{A}^2$, $\pi_b^2(up|z^2) = 0.3, \ \pi_b^2(down|z^2) = 0.7, \ \forall \ z^2 \in \mathcal{Z}^2$.

\textbf{Evaluation Policy:}

$\pi_e(up\_right|s) = 0.9025, \ \pi_e(up\_left|s) = 0.0475, \ \pi_e(down\_right|s) = 0.0475, \ \pi_e(down\_right|s) = 0.0025, \ \forall \ s \in \mathcal{S}$.

With respect to $\mathcal{A}^1$, $\pi^1_e(right|z^1) = 0.95, \ \pi^1_e(left|z^1) = 0.05, \ \forall \ z^1 \in \mathcal{Z}^1$.

With respect to $\mathcal{A}^2$, $\pi^2_e(up|z^2) = 0.95, \ \pi^2_e(down|z^2) = 0.05, \ \forall \ z^2 \in \mathcal{Z}^2$.

\subsubsection{Policy Divergence = $22.56^T$}

\textbf{Behaviour Policy:}

$\pi_b(up\_right|s) = 0.04, \ \pi_b(up\_left|s) = 0.16, \ \pi_b(down\_right|s) = 0.16, \ \pi_b(down\_left|s) = 0.64, \ \forall \ s \in \mathcal{S}$.

With respect to $\mathcal{A}^1$, $\pi_b^1(right|z^1) = 0.2, \ \pi_b^1(left|z^1) = 0.8, \ \forall \ z^1 \in \mathcal{Z}^1$.

With respect to $\mathcal{A}^2$, $\pi_b^2(up|z^2) = 0.2, \ \pi_b^2(down|z^2) = 0.8, \ \forall \ z^2 \in \mathcal{Z}^2$.

\textbf{Evaluation Policy:}

$\pi_e(up\_right|s) = 0.9025, \ \pi_e(up\_left|s) = 0.0475, \ \pi_e(down\_right|s) = 0.0475, \ \pi_e(down\_right|s) = 0.0025, \ \forall \ s \in \mathcal{S}$.

With respect to $\mathcal{A}^1$, $\pi^1_e(right|z^1) = 0.95, \ \pi^1_e(left|z^1) = 0.05, \ \forall \ z^1 \in \mathcal{Z}^1$.

With respect to $\mathcal{A}^2$, $\pi^2_e(up|z^2) = 0.95, \ \pi^2_e(down|z^2) = 0.05, \ \forall \ z^2 \in \mathcal{Z}^2$.

\subsubsection{Policy Divergence = $90.25^T$}

\textbf{Behaviour Policy:}

$\pi_b(up\_right|s) = 0.04, \ \pi_b(up\_left|s) = 0.16, \ \pi_b(down\_right|s) = 0.16, \ \pi_b(down\_left|s) = 0.81, \ \forall \ s \in \mathcal{S}$.

With respect to $\mathcal{A}^1$, $\pi_b^1(right|z^1) = 0.1, \ \pi_b^1(left|z^1) = 0.9, \ \forall \ z^1 \in \mathcal{Z}^1$.

With respect to $\mathcal{A}^2$, $\pi_b^2(up|z^2) = 0.1, \ \pi_b^2(down|z^2) = 0.9, \ \forall \ z^2 \in \mathcal{Z}^2$.

\textbf{Evaluation Policy:}

$\pi_e(up\_right|s) = 0.9025, \ \pi_e(up\_left|s) = 0.0475, \ \pi_e(down\_right|s) = 0.0475, \ \pi_e(down\_right|s) = 0.0025, \ \forall \ s \in \mathcal{S}$.

With respect to $\mathcal{A}^1$, $\pi^1_e(right|z^1) = 0.95, \ \pi^1_e(left|z^1) = 0.05, \ \forall \ z^1 \in \mathcal{Z}^1$.

With respect to $\mathcal{A}^2$, $\pi^2_e(up|z^2) = 0.95, \ \pi^2_e(down|z^2) = 0.05, \ \forall \ z^2 \in \mathcal{Z}^2$.

\subsubsection{Policy Divergence = $361.0^T$}

\textbf{Behaviour Policy:}

$\pi_b(up\_right|s) = 0.0025, \ \pi_b(up\_left|s) = 0.0475, \ \pi_b(down\_right|s) = 0.0475, \ \pi_b(down\_left|s) = 0.9025, \ \forall \ s \in \mathcal{S}$.

With respect to $\mathcal{A}^1$, $\pi_b^1(right|z^1) = 0.05, \ \pi_b^1(left|z^1) = 0.95, \ \forall \ z^1 \in \mathcal{Z}^1$.

With respect to $\mathcal{A}^2$, $\pi_b^2(up|z^2) = 0.05, \ \pi_b^2(down|z^2) = 0.95, \ \forall \ z^2 \in \mathcal{Z}^2$.

\textbf{Evaluation Policy:}

$\pi_e(up\_right|s) = 0.9025, \ \pi_e(up\_left|s) = 0.0475, \ \pi_e(down\_right|s) = 0.0475, \ \pi_e(down\_right|s) = 0.0025, \ \forall \ s \in \mathcal{S}$.

With respect to $\mathcal{A}^1$, $\pi^1_e(right|z^1) = 0.95, \ \pi^1_e(left|z^1) = 0.05, \ \forall \ z^1 \in \mathcal{Z}^1$.

With respect to $\mathcal{A}^2$, $\pi^2_e(up|z^2) = 0.95, \ \pi^2_e(down|z^2) = 0.05, \ \forall \ z^2 \in \mathcal{Z}^2$.

\newpage

\section{Additional Plots From Experiments}

\subsection{MDP 1}

\subsubsection{Varying the Number of Trajectories in MDP 1 With Policy Divergence 1.44}

\begin{figure}[h!]
    \centering
    \includegraphics[scale=0.5]{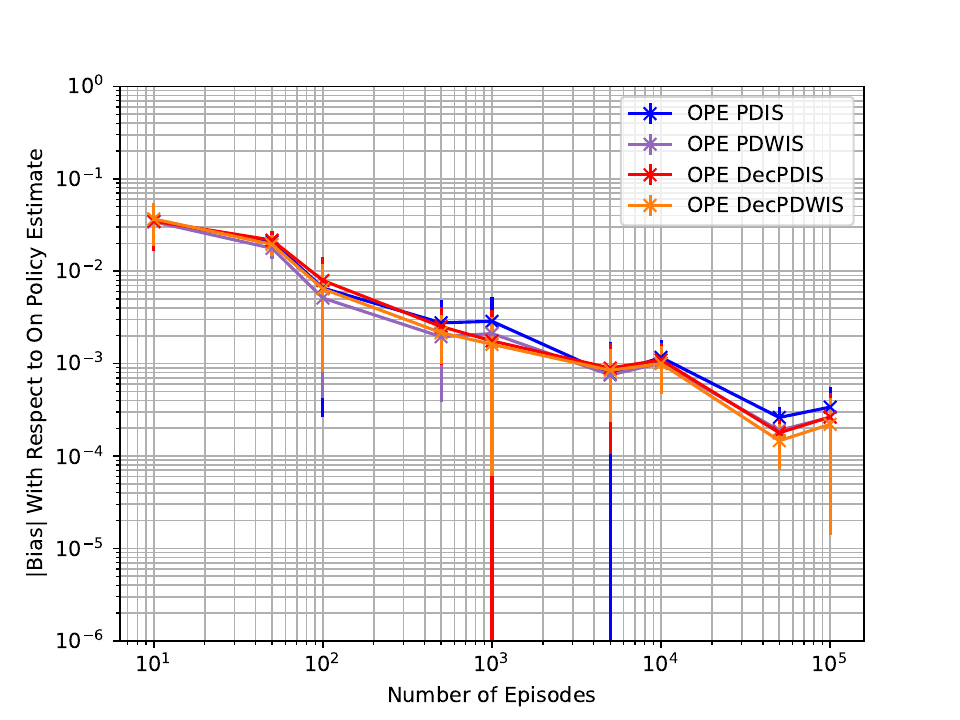}
    \includegraphics[scale=0.5]{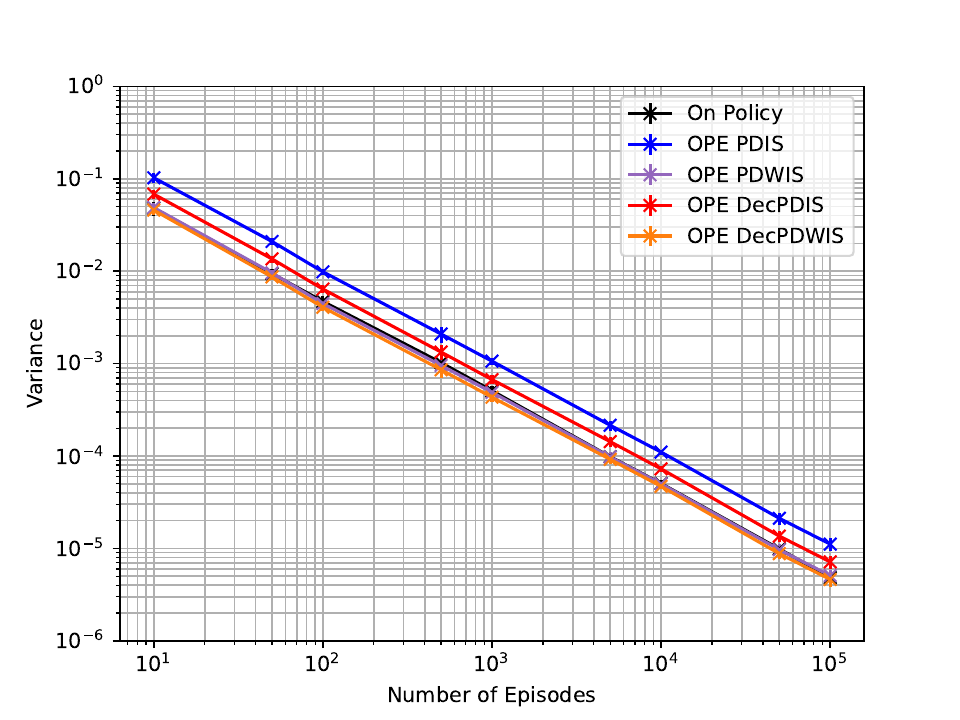}
    \includegraphics[scale=0.5]{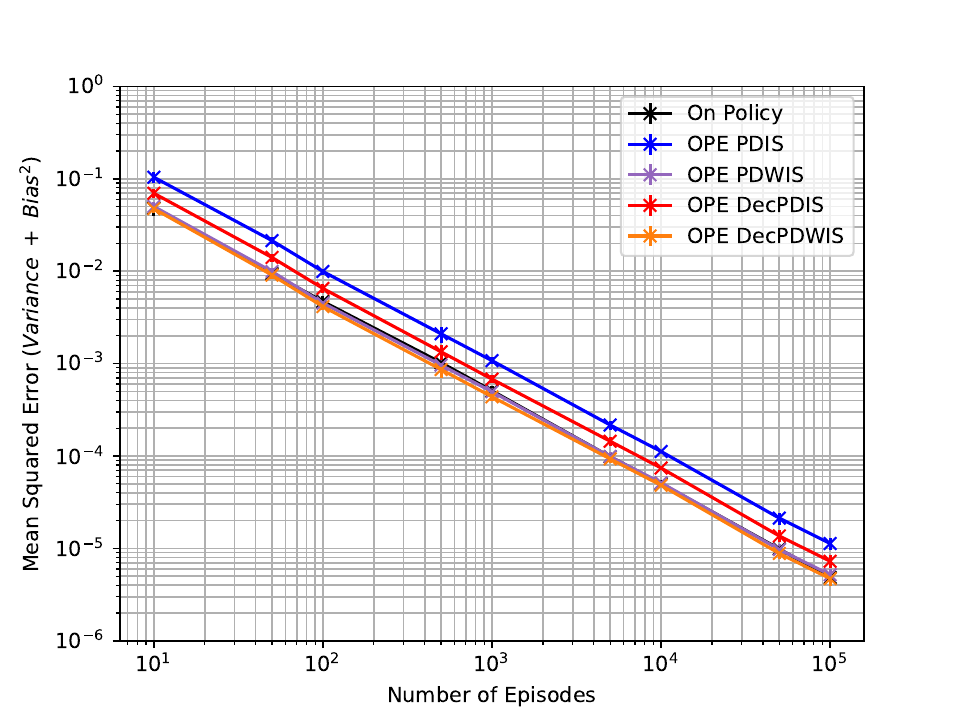}
    \includegraphics[scale=0.5]{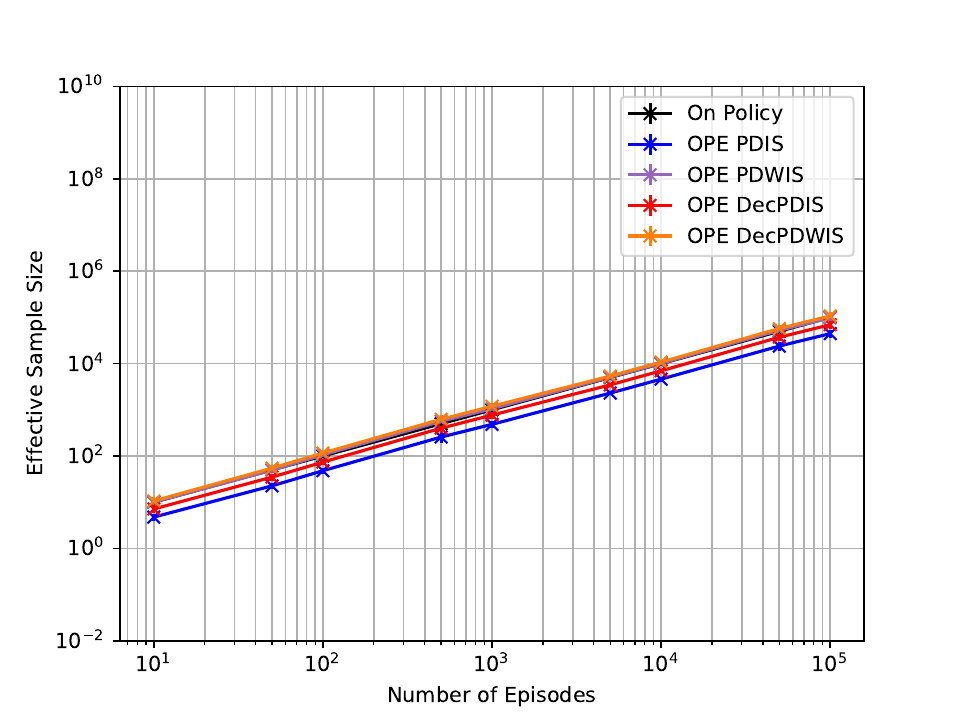}
    \caption{The biases of all estimators are very low and gradually converge to a lower limit of zero bias. Since this is a 1-step problem, the bias is low even for the PDWIS estimators. The variance shows similar behaviour to Figure \ref{fig:var-vs-trajectories-policy-divergence-2-56}, except that the graphs are closer together i.e. less difference between the variances of the policies due to lower policy divergence. MSE is dominated by the variance behaviour and is thus similar to Figure \ref{fig:var-vs-trajectories-policy-divergence-2-56}. The ESS graphs look similar to the variance graphs inverted about the x-axis, and have similar but inverted trends.}
    \label{fig:var-vs-trajectories-policy-divergence-1-44}
\end{figure}

\newpage

\subsubsection{Varying the Number of Trajectories in MDP 1 With Policy Divergence 2.56}

\begin{figure}[h!]
    \centering
    \includegraphics[scale=0.5]{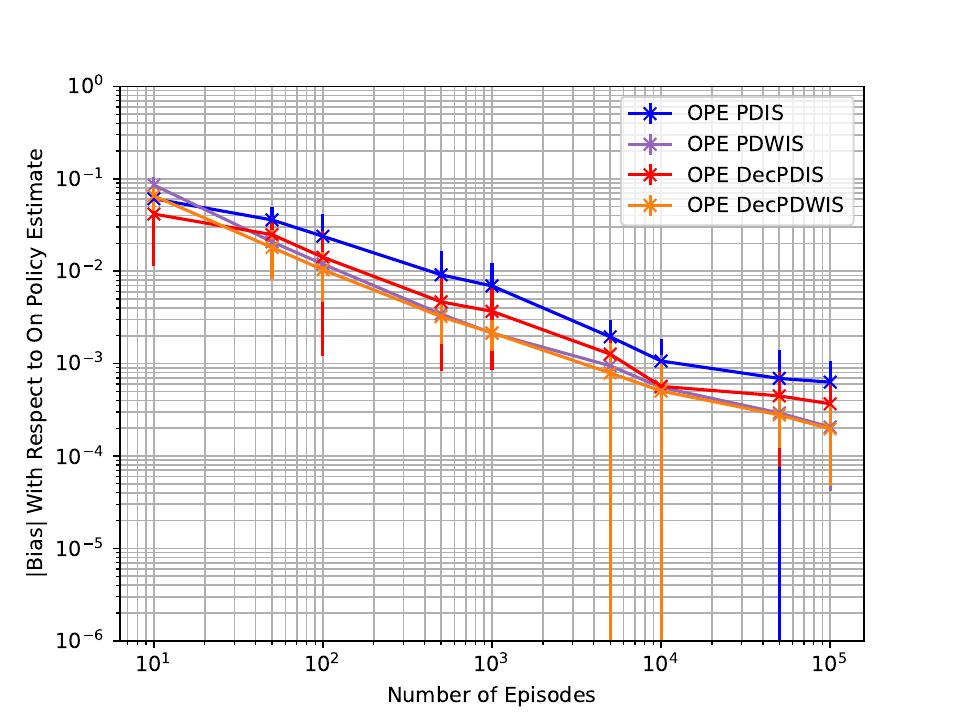}
    \includegraphics[scale=0.5]{images/1-step-MDP-experiments/var-vs-episodes-policy-divergence-2-56.pdf}
    \includegraphics[scale=0.5]{images/1-step-MDP-experiments/MSE-vs-episodes-policy-divergence-2-56.pdf}
    \includegraphics[scale=0.5]{images/1-step-MDP-experiments/ESS-vs-episodes-policy-divergence-2-56.pdf}
    \caption{The overall bias, variance, MSE and ESS behaviour of the estimators is highly similar to Figure \ref{fig:var-vs-trajectories-policy-divergence-1-44}. However, the graphs are more spread apart; this could  due to the increased policy divergence. This makes OPE estimation more difficult and results in the distinction between estimators becoming more clear. In the graphs, it is clear that the PDWIS estimators are performing better in this problem. Note that the variance and MSE graphs are the same as Figure \ref{fig:var-vs-trajectories-policy-divergence-2-56} in the main paper, while the ESS graph is the same as Figure \ref{fig:ESS-vs-trajectories-policy-divergence-2-56}.}
    \label{fig:bias-vs-trajectories-policy-divergence-2-56}
\end{figure}

\newpage

\subsubsection{Varying the Number of Trajectories in MDP 1 With Policy Divergence 3.61}

\begin{figure}[h!]
    \centering
    \includegraphics[scale=0.5]{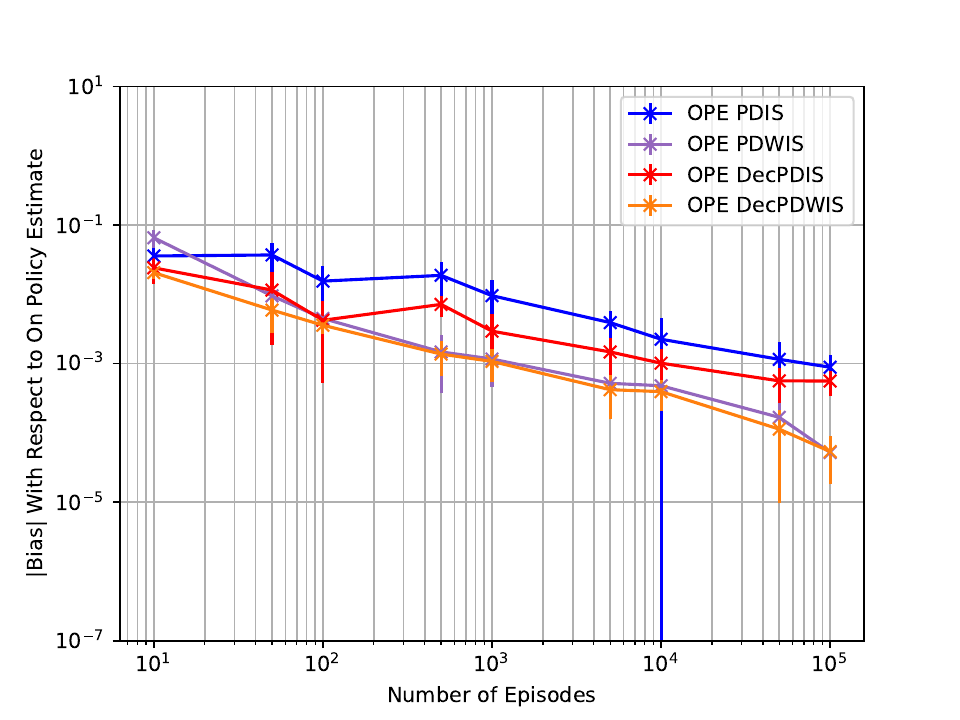}
    \includegraphics[scale=0.5]{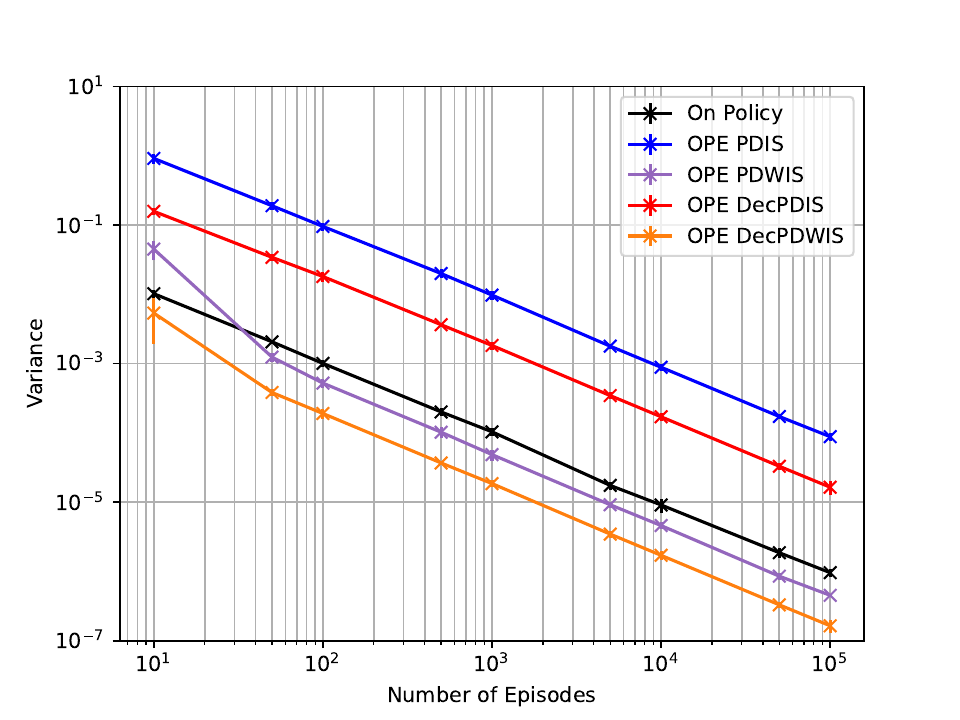}
    \includegraphics[scale=0.5]{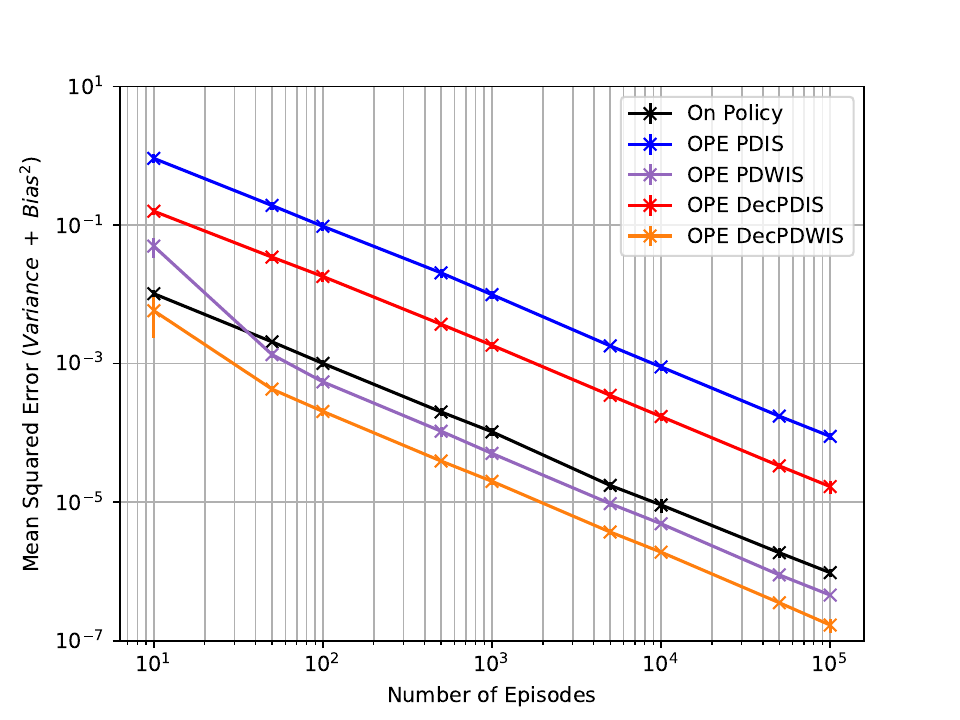}
    \includegraphics[scale=0.5]{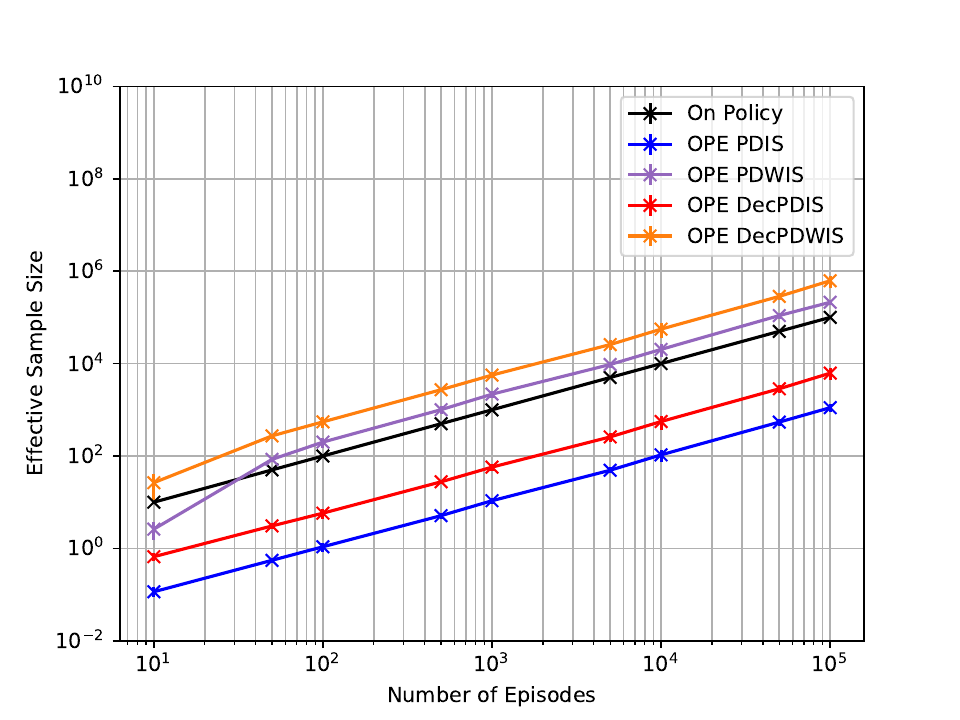}
    \caption{The bias, variance, MSE and ESS graphs again show almost identical trends to Figures \ref{fig:var-vs-trajectories-policy-divergence-1-44} and \ref{fig:bias-vs-trajectories-policy-divergence-2-56}, with the graphs being more spread out due to greater policy divergence. The superiority of PDWIS estimators over all other OPE methods, and over on-policy estimation, is clearer.}
    \label{fig:var-vs-trajectories-policy-divergence-3-61}
\end{figure}

\newpage

\subsubsection{Varying $\beta$ in MDP-1 With Policy Divergence 1.44 and 1000 Trajectories}

\begin{figure}[h!]
    \centering
    \includegraphics[scale=0.5]{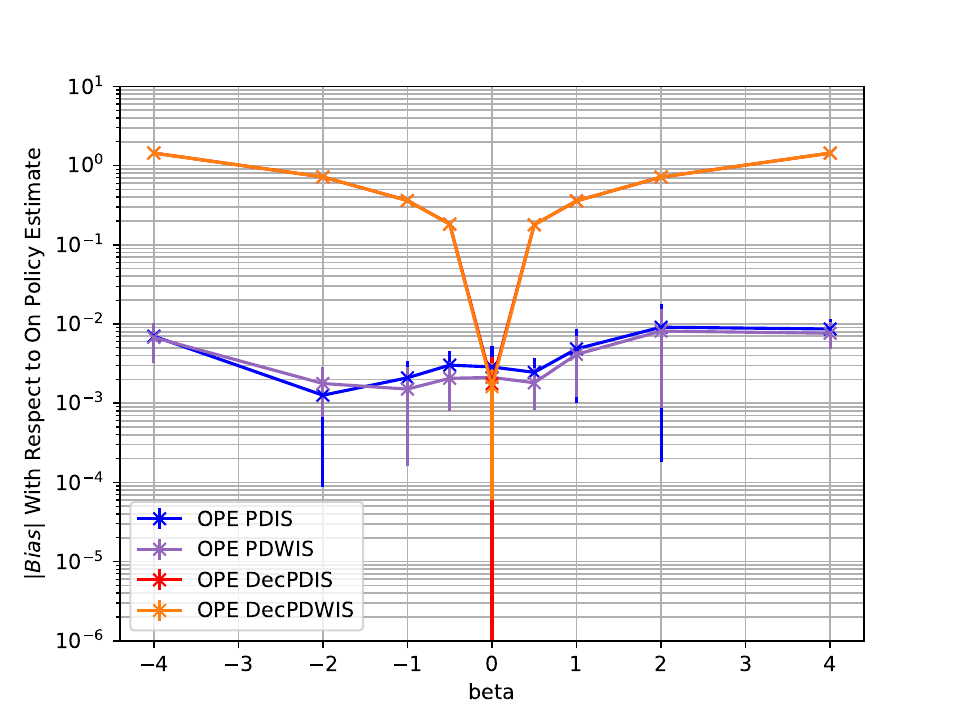}
    \includegraphics[scale=0.5]{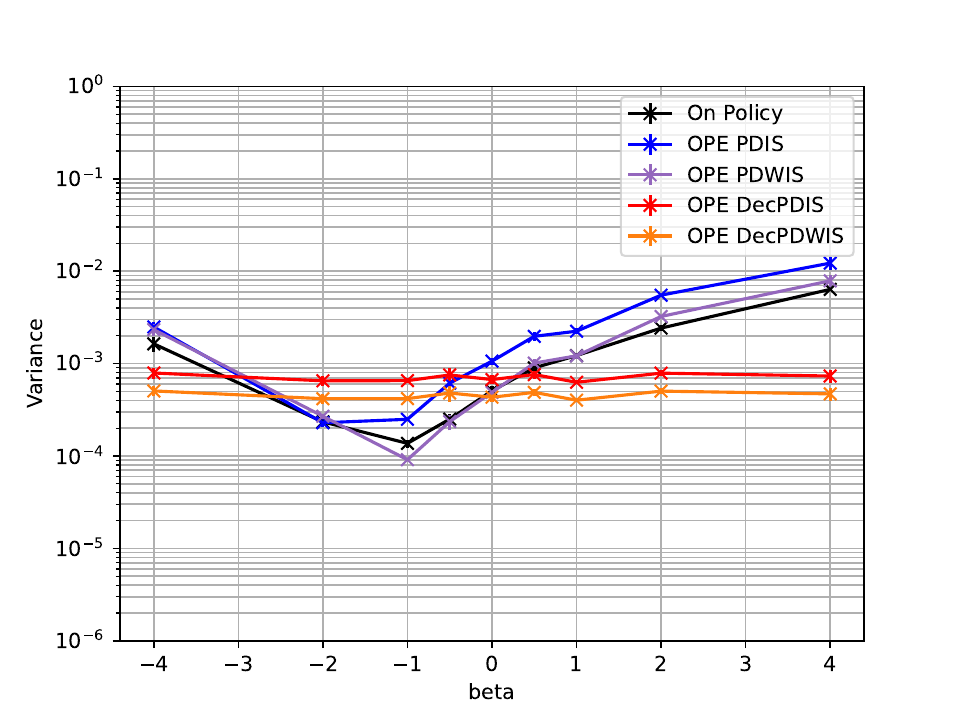}
    \includegraphics[scale=0.5]{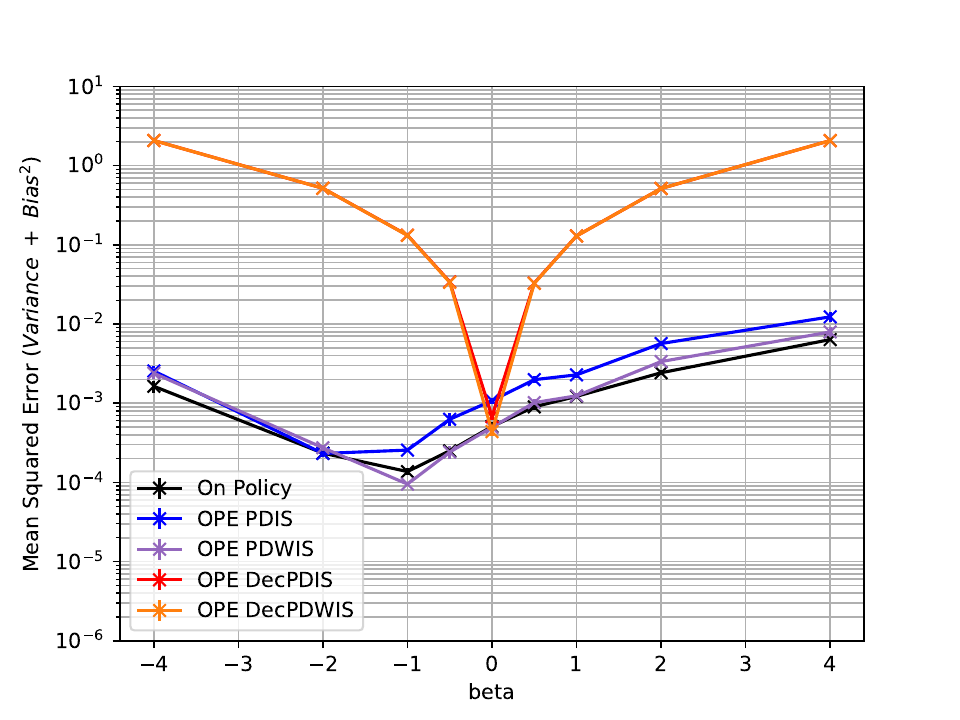}
    \includegraphics[scale=0.5]{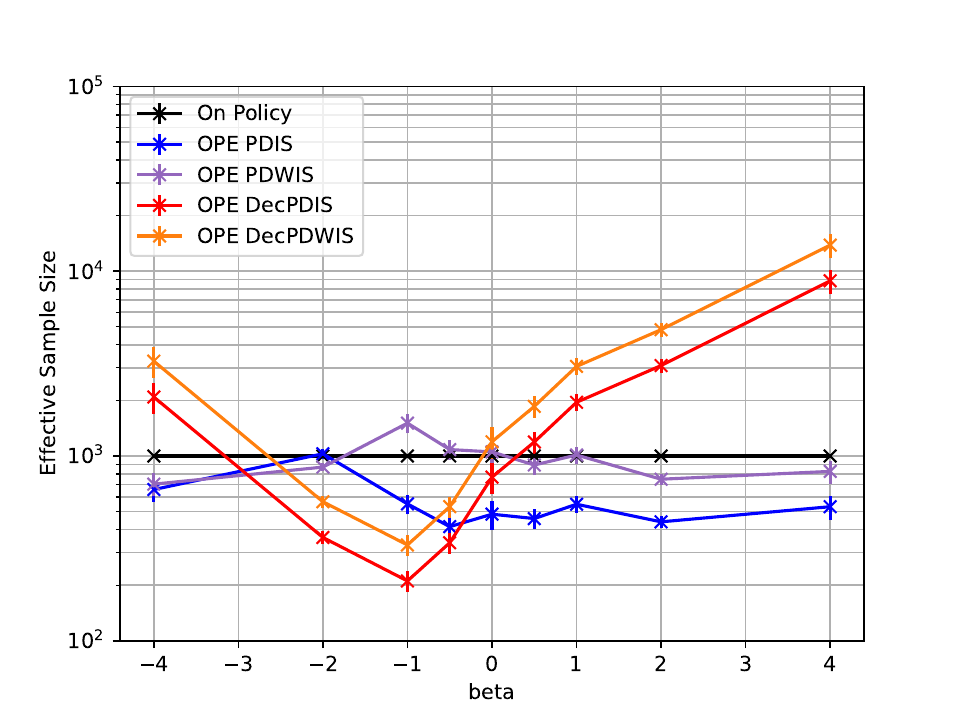}
    \caption{The bias graphs are similar to Figure \ref{fig:bias-100000-trajectories-vs-beta} in the main paper. Similar to that figure, we observe that for $\beta=0$, the bias of the decomposed estimators is comparable to that of the non-decomposed estimators. However, as we increase $|\beta|$, the bias of the decomposed estimators scales so dramatically that this behaviour dominates the MSE performance of these estimators. This is because for $|\beta| \ne 0$, Theorem \ref{theorem-1-shengpu paper} no longer applies and thus it is not assured that the MDP can be factored into sub-MDP's based on factored action spaces without introducing bias. It is notable that the MSE of non-decomposed estimators is dominated by variance, rather than bias. Meanwhile, $\beta$ affects the variance of the on-policy and non-decomposed estimators, and does not affect that of the decomposed estimators. The presence of a variance minimum at $\beta=-1$ is due to the equal reward in the MDP for three different actions - see Appendix \ref{sec:diag-MDP-1}. The further we depart from this equal-reward scenario, the more the variance of the on-policy and non-decomposed estimators increase. The relative trajectories of the OPE and on-policy estimator variance graphs determine the ESS graph.}
    \label{fig:var-vs-beta-policy-divergence-1-44-N-100}
\end{figure}

\newpage

\subsubsection{Varying $\beta$ in MDP-1 With Policy Divergence 1.44 and 100,000 Trajectories}

\begin{figure}[h!]
    \centering
    \includegraphics[scale=0.5]{images/1-step-MDP-experiments/bias-100000-episodes-vs-beta.pdf}
    \includegraphics[scale=0.5]{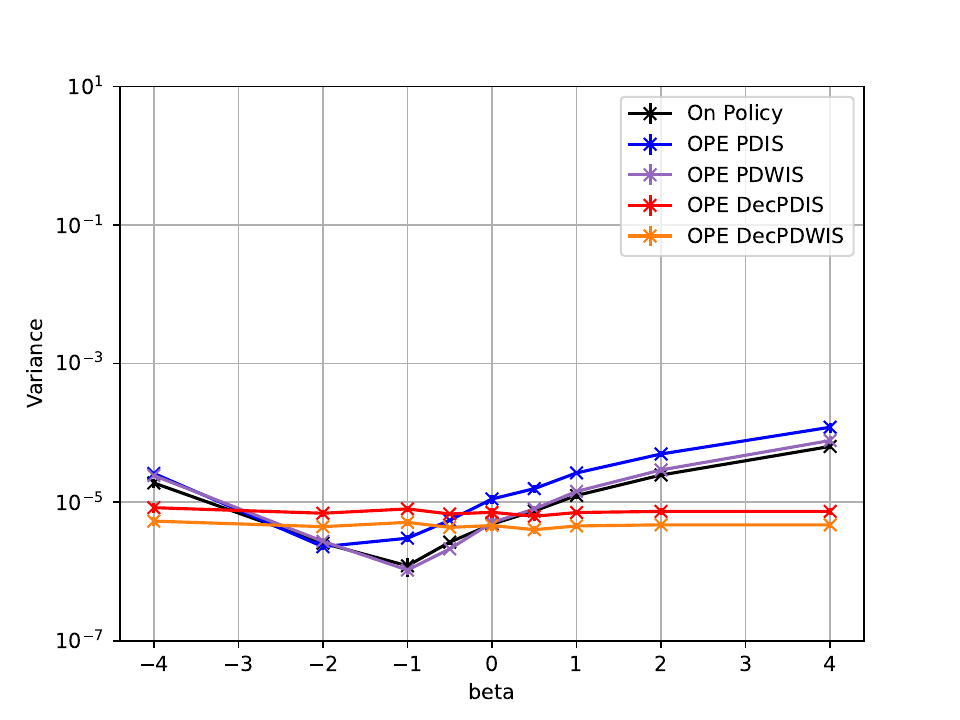}
    \includegraphics[scale=0.5]{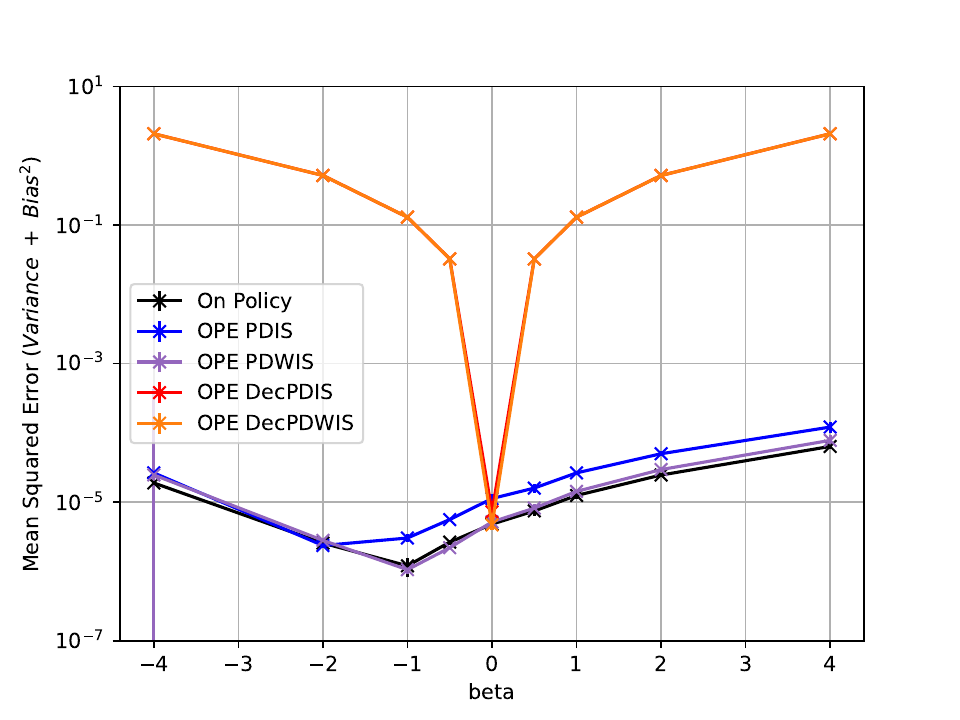}
    \includegraphics[scale=0.5]{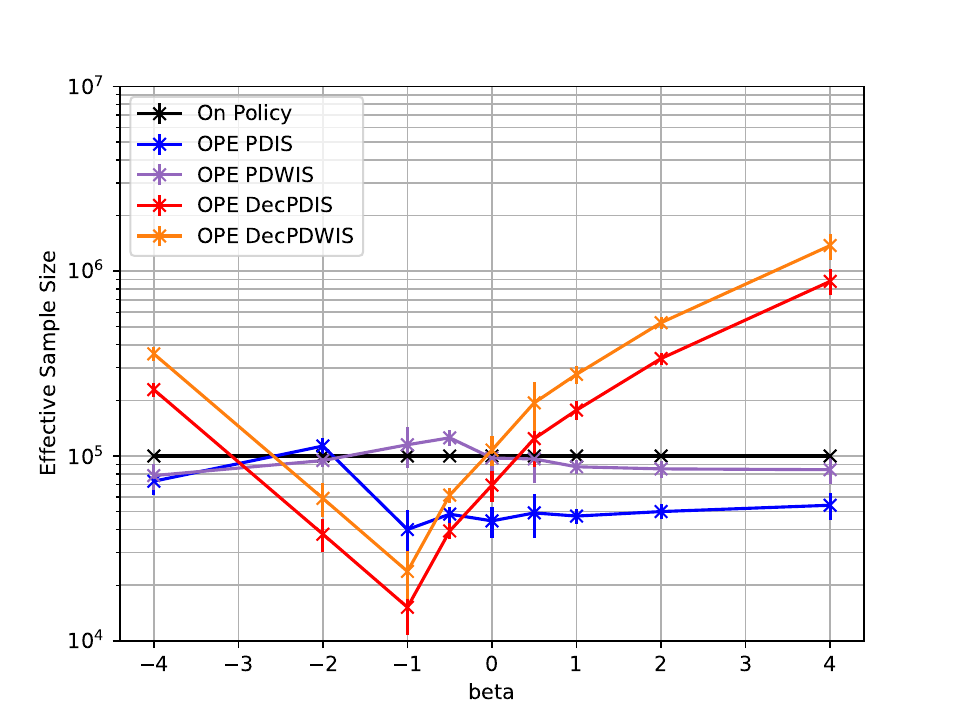}
    \caption{These graphs are highly similar to Figure \ref{fig:var-vs-beta-policy-divergence-1-44-N-100}, except that the variance and MSE are at lower scales and the ESS is at a higher scale.}
    \label{fig:var-vs-beta-policy-divergence-1-44-N-100000}
\end{figure}

\newpage

\subsubsection{Varying Policy Divergence in MDP-1 With 1000 Trajectories}

\begin{figure}[h!]
    \centering
    \includegraphics[scale=0.5]{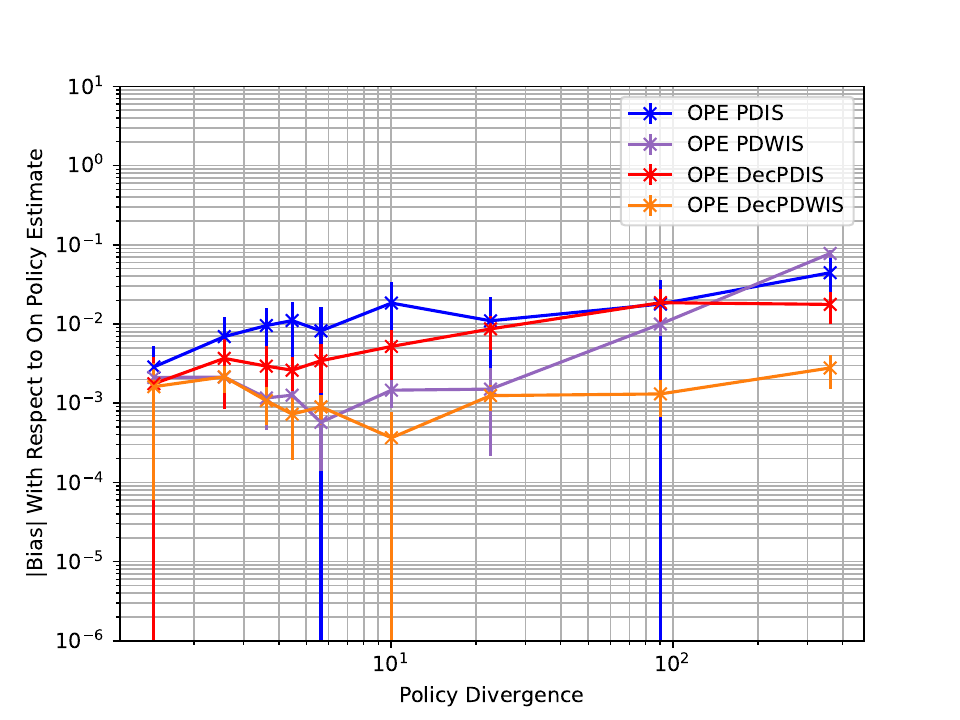}
    \includegraphics[scale=0.5]{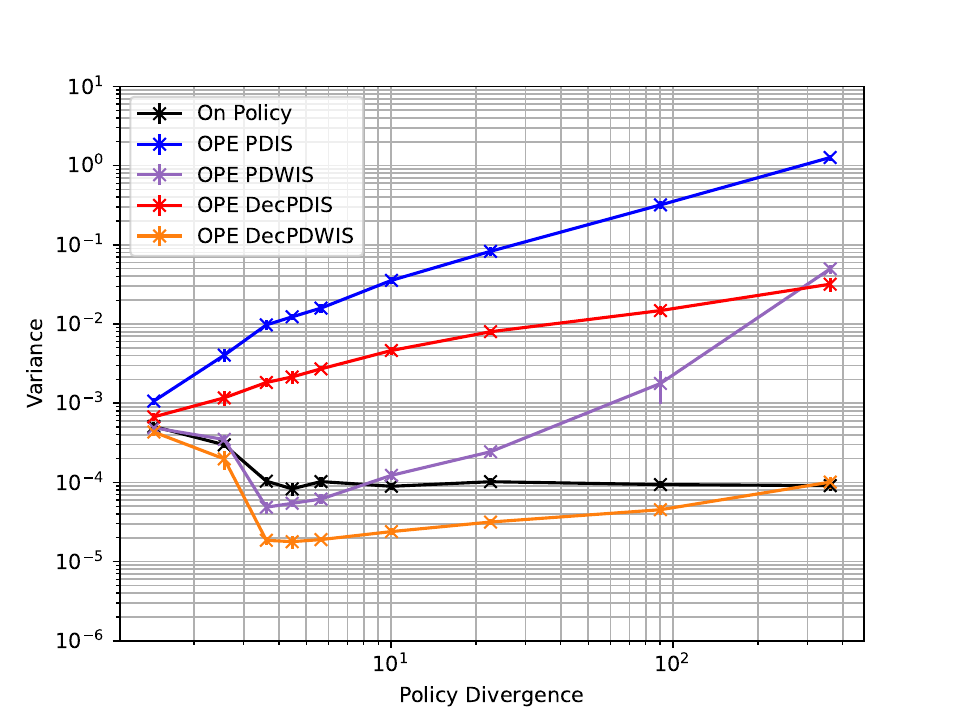}
    \includegraphics[scale=0.5]{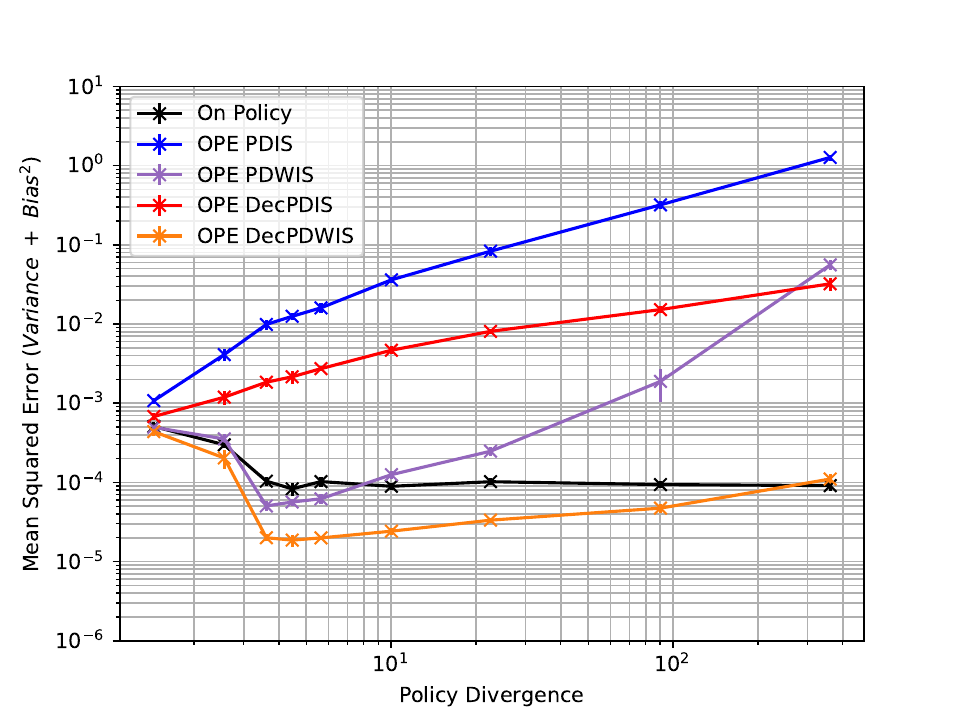}
    \includegraphics[scale=0.5]{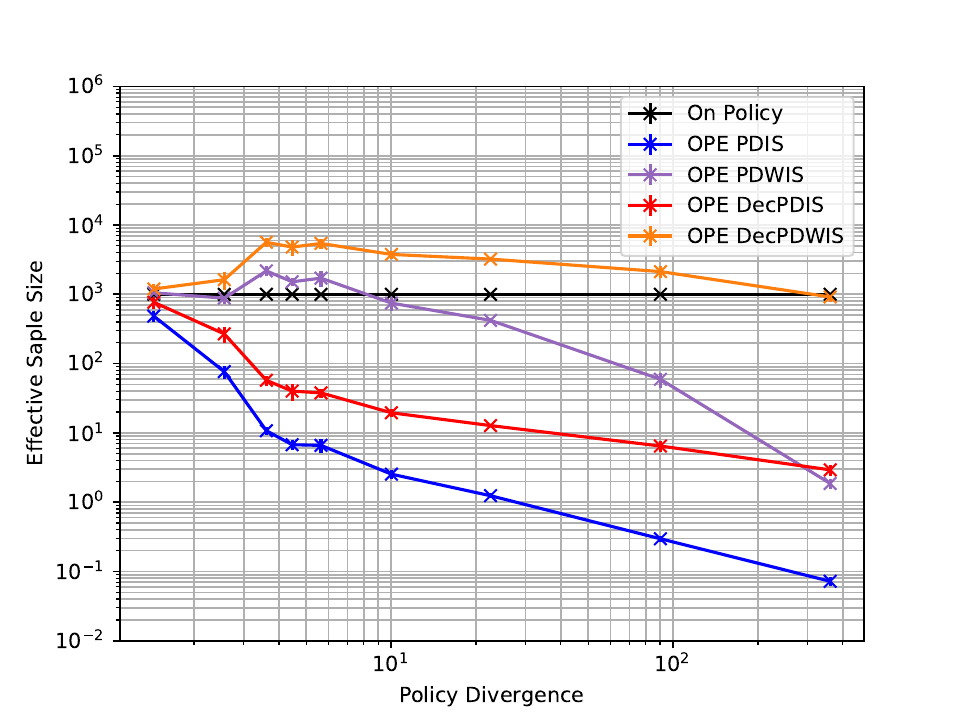}
    \caption{As the policy divergence increases, bias, with some random anomalies, increases due to decreasing coverage of the evaluation policy by the behaviour policy dataset. The variances of PDIS and decomposed PDIS also increase with policy divergence for the same reason, however decomposed PDIS always has lower variance and also scales slower than non-decomposed PDIS. Clearly, decomposed PDIS has ensured better coverage here. On the other hand, the variance of the PDWIS estimators first fall and then increase - this may be due to the sum of IS weights in the denominator initially becoming larger due to policy divergence then becoming smaller due to lower coverage. The ESS again inverts the variance graph, while the MSE is primarily affected by the variance.}
    \label{fig:var-vs-policy-divergence}
\end{figure}

\newpage

\subsubsection{Varying Policy Divergence in MDP-1 With 100,000 Trajectories}

\begin{figure}[h!]
    \centering
    \includegraphics[scale=0.5]{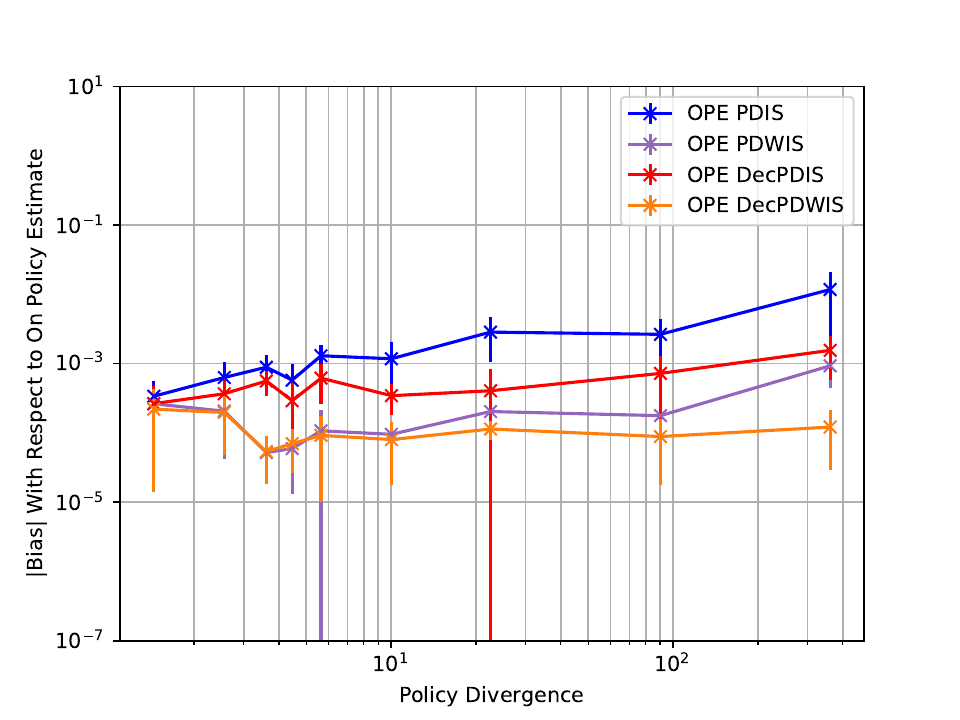}
    \includegraphics[scale=0.5]{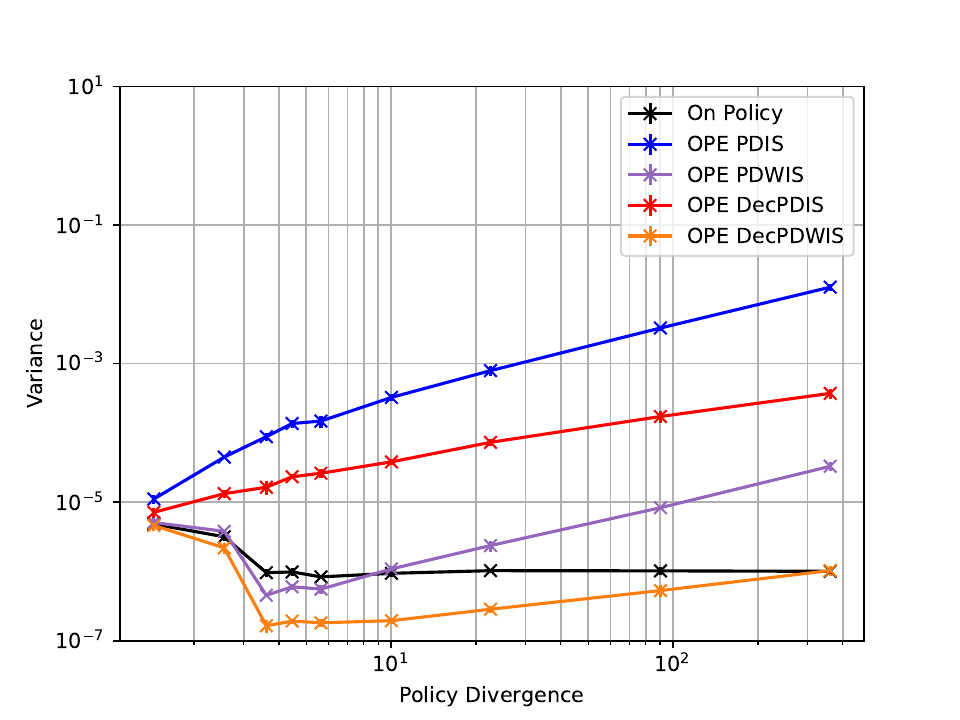}
    \includegraphics[scale=0.5]{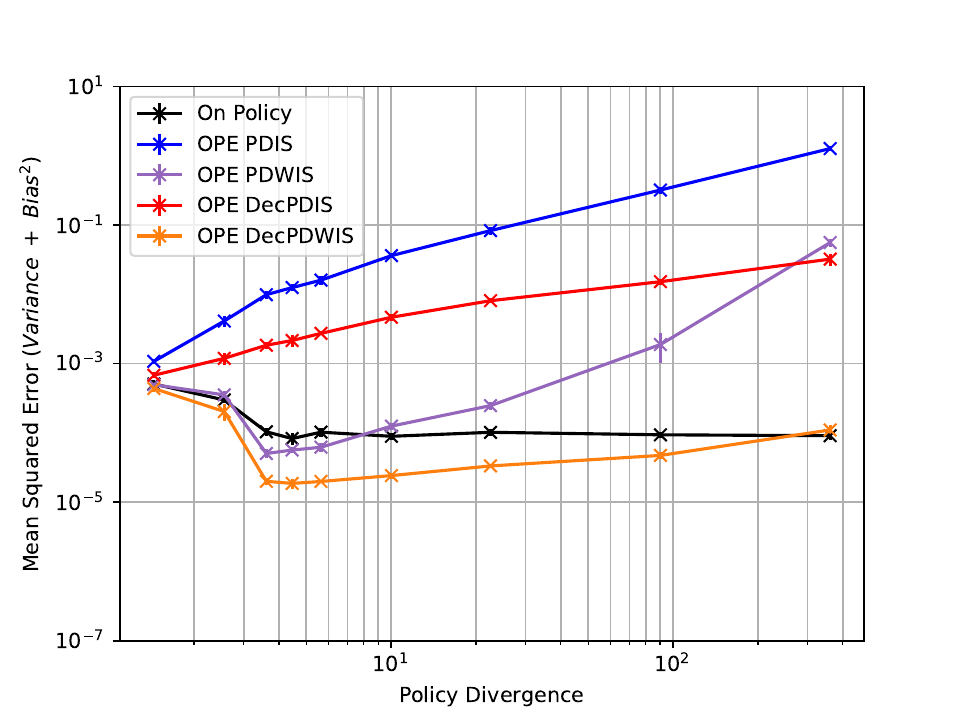}
    \includegraphics[scale=0.5]{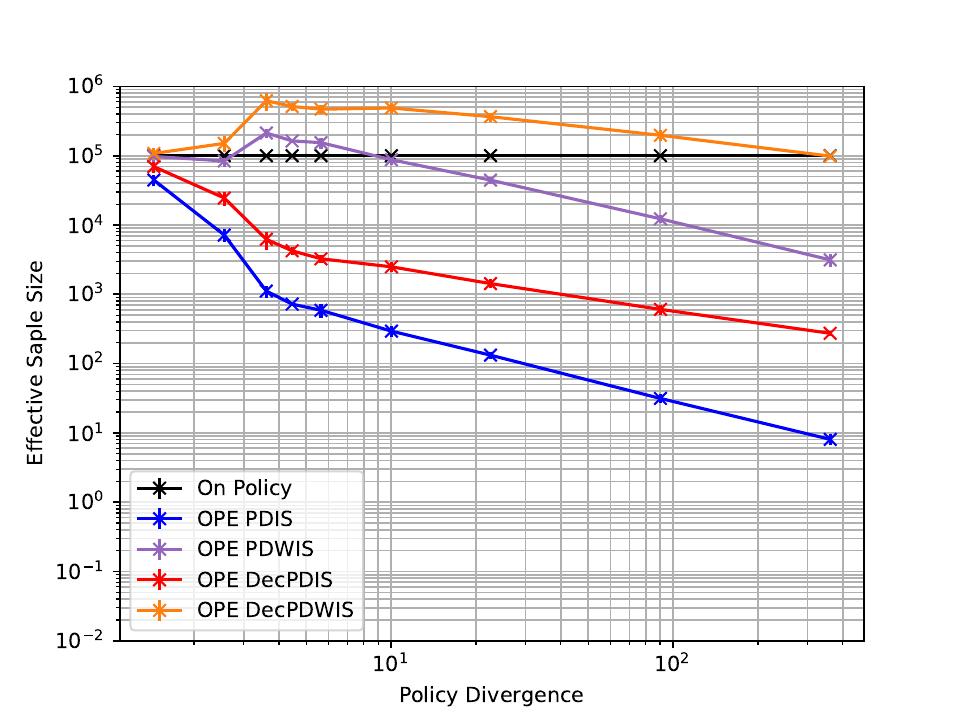}
    \caption{These graphs are highly similar to those in Figure \ref{fig:var-vs-policy-divergence}, except that the variance and ESS are on different scales i.e. variance is less and ESS is more.}
    \label{fig:var-vs-policy-divergence-10000-episodes}
\end{figure}

\newpage

\subsection{MDP 2}

\subsubsection{Varying the Trajectory Length $T$ in MDP 2 With Policy Divergence $1.44^T$, 1000 Trajectories and Discount Factor 0.7}
\label{images-MDP-2-gamma-0-7}

\begin{figure}[h!]
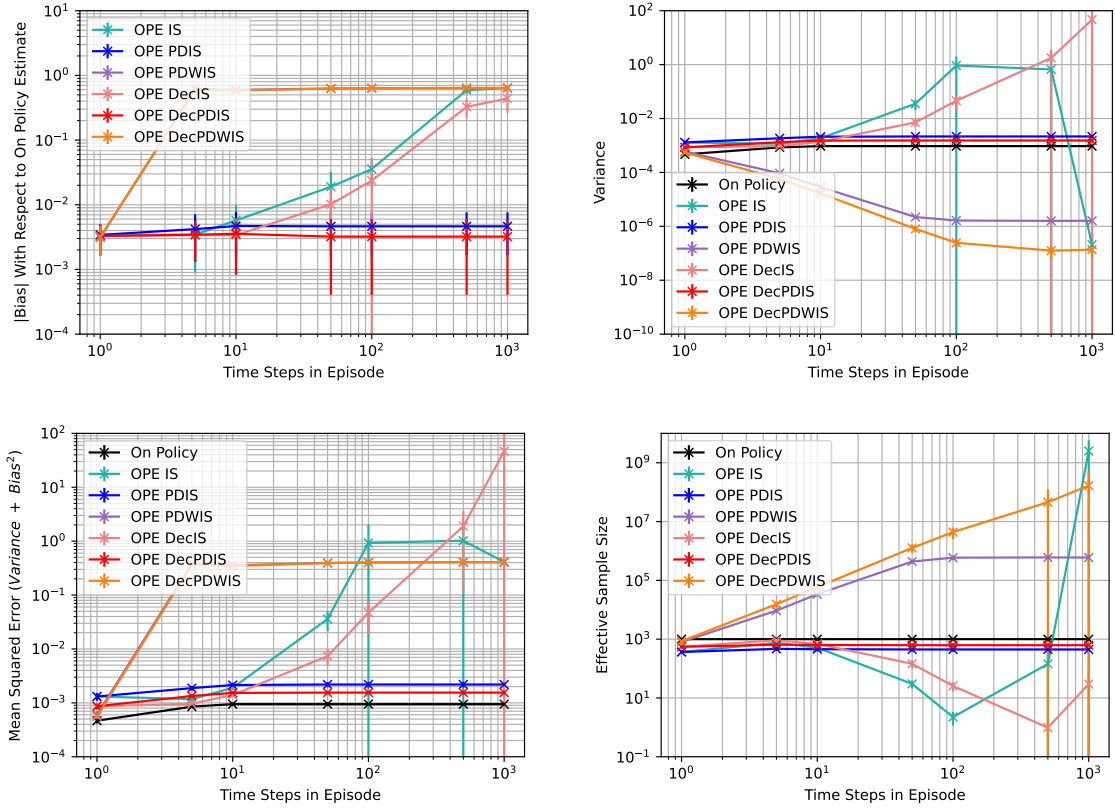

    \centering
    \includegraphics[scale=0.55]{images/4-state-MDP-experiments/bias-vs-episode-length-discount-0-7-policy-divergence-1-44.pdf}
    \includegraphics[scale=0.55]{images/4-state-MDP-experiments/var-vs-episode-length-discount-0-7-policy-divergence-1-44.pdf}
    \includegraphics[scale=0.55]{images/4-state-MDP-experiments/MSE-vs-episode-length-discount-0-7-policy-divergence-1-44.pdf}
    \includegraphics[scale=0.55]{images/4-state-MDP-experiments/ESS-vs-episode-length-discount-0-7-policy-divergence-1-44.pdf}
    \caption{These graphs are the same as those in figures \ref{fig:bias-vs-trajectory-length-discount-0-7-policy-divergence-1-44}, \ref{fig:var-vs-trajectory-length-discount-0-7-policy-divergence-1-44} and \ref{fig:ESS-vs-trajectory-length-discount-0-7-policy-divergence-1-44} in the main report. They are discussed at length in the main report. The decrease in variance of the PDWIS estimators with $T$ may be due to the IS weights in the denominator becoming larger in magnitude.}
    \label{fig:var-vs-trajectory-length-discount-0.7}
\end{figure}

\newpage

\subsubsection{Varying the Trajectory Length $T$ in MDP 2 With Policy Divergence $1.44^T$, 1000 Trajectories and Discount Factor 0.9}
\label{images-MDP-2-gamma-0-9}

\begin{figure}[h!]
    \centering
    \includegraphics[scale=0.55]{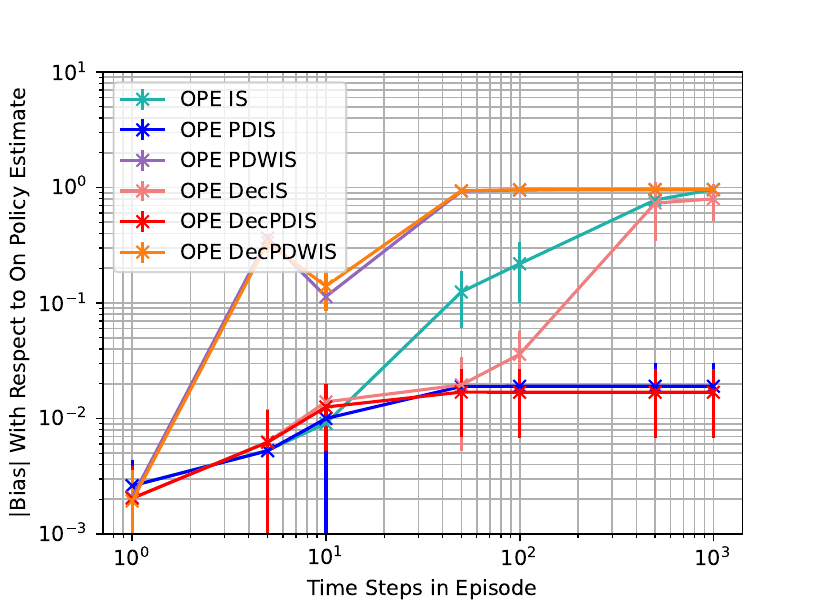}
    \includegraphics[scale=0.55]{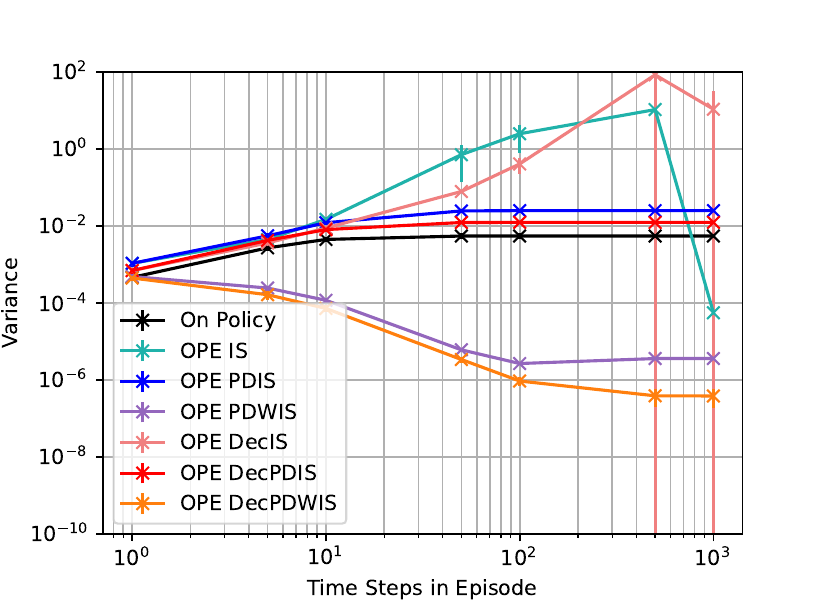}
    \includegraphics[scale=0.55]{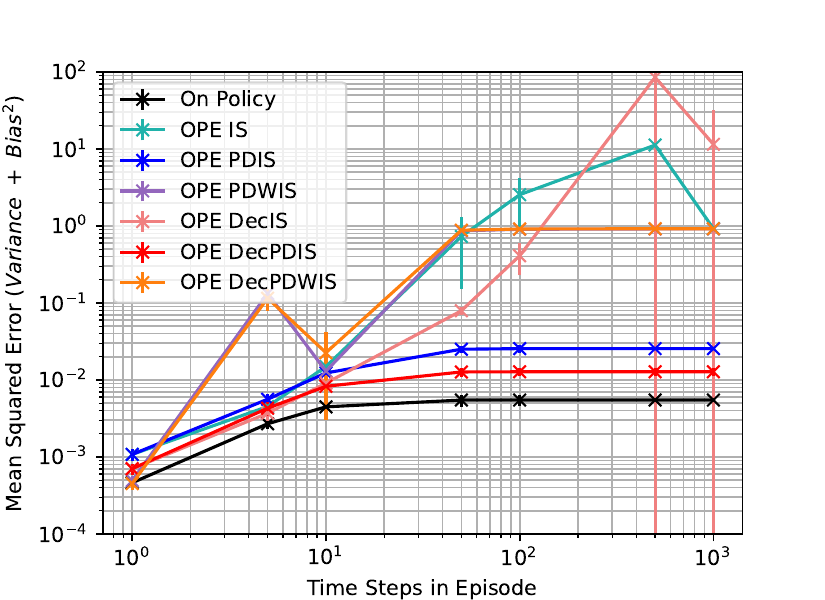}
    \includegraphics[scale=0.55]{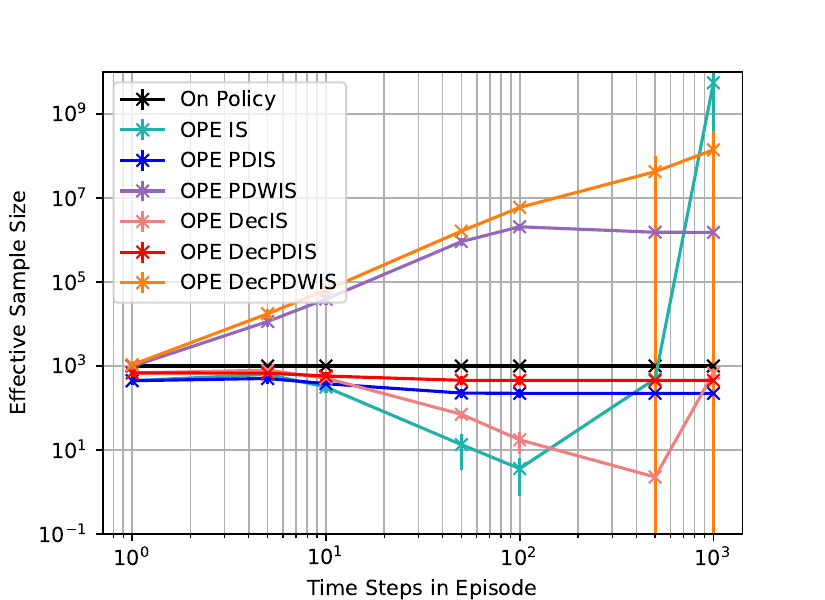}
    \caption{These graphs are similar to Figure \ref{fig:var-vs-trajectory-length-discount-0.7} with some differences. These differences are observed because the larger discount factor enables transitions further into the future to contribute to the Q-function estimate. The bias and variance of the PDIS estimators increase more before reaching their constant values. The variances of the IS estimators also scale faster and reach the point where the estimate is a low-variance, high bias value. The biases of the PDWIS estimators actually scale more slowly; this may be because the IS weight sum in the denominator is larger in magnitude for larger $\gamma$, which scales down the variance. Finally the gap between the MSE of the estimators and that of the on-policy estimate has increased.}
    \label{fig:var-vs-trajectories-discount-0-9}
\end{figure}

\newpage

\subsubsection{Varying the Trajectory Length $T$ in MDP 2 With Policy Divergence $1.44^T$, 1000 Trajectories and Discount Factor 0.9999}
\label{images-MDP-2-gamma-0-9999}

\begin{figure}[h!]
    \centering
    \includegraphics[scale=0.55]{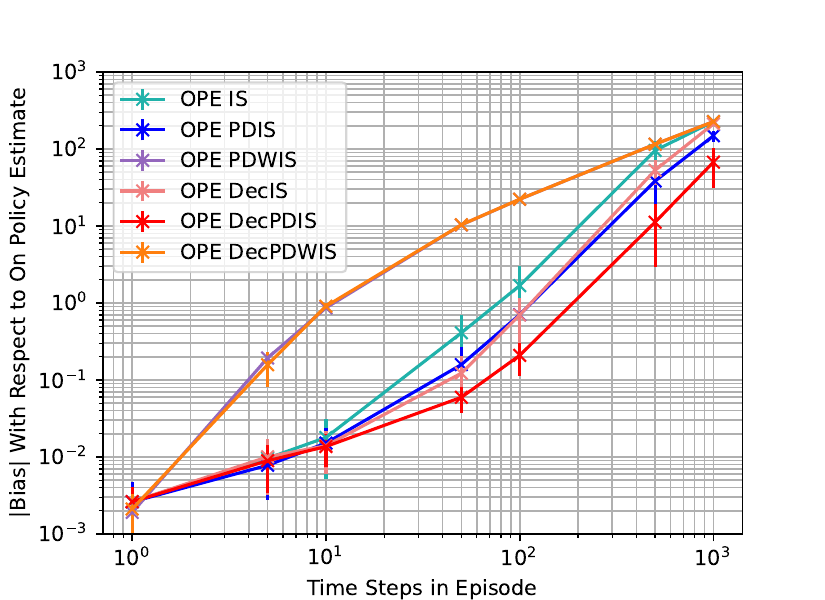}
    \includegraphics[scale=0.55]{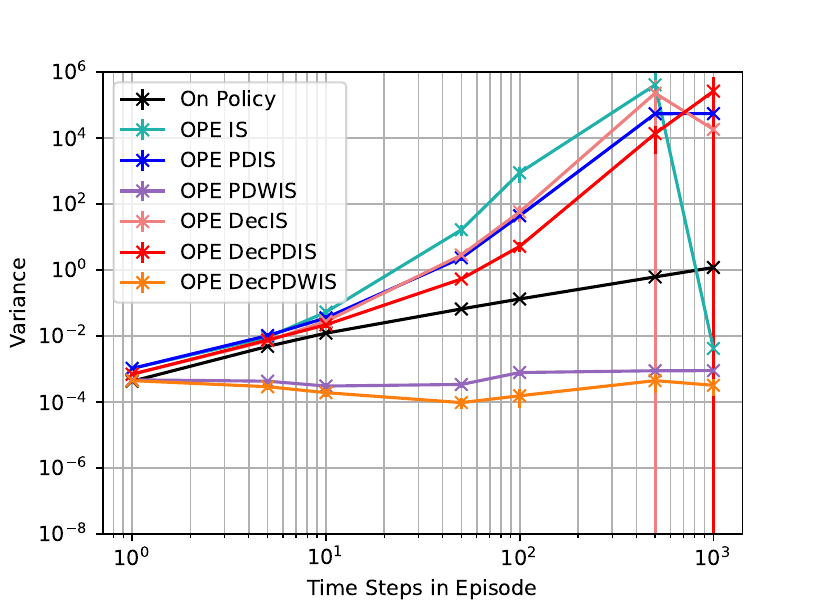}
    \includegraphics[scale=0.55]{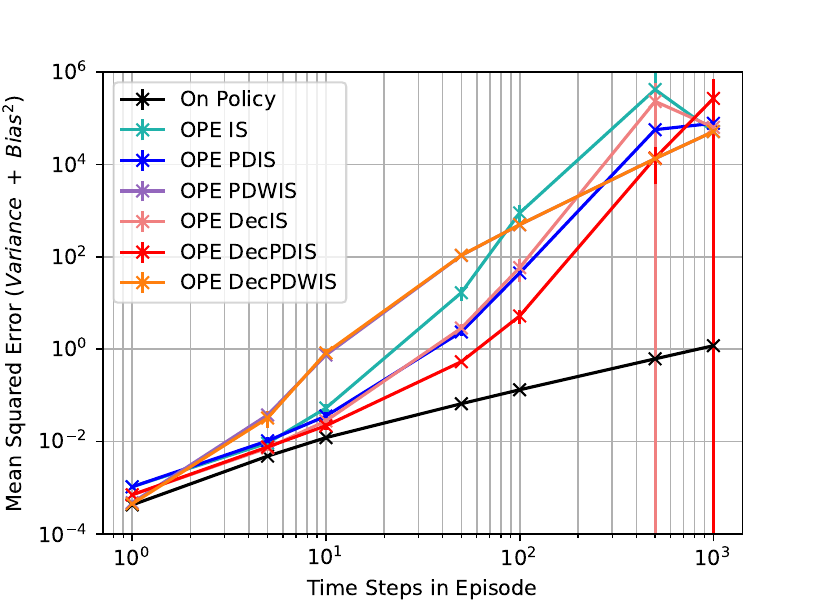}
    \includegraphics[scale=0.55]{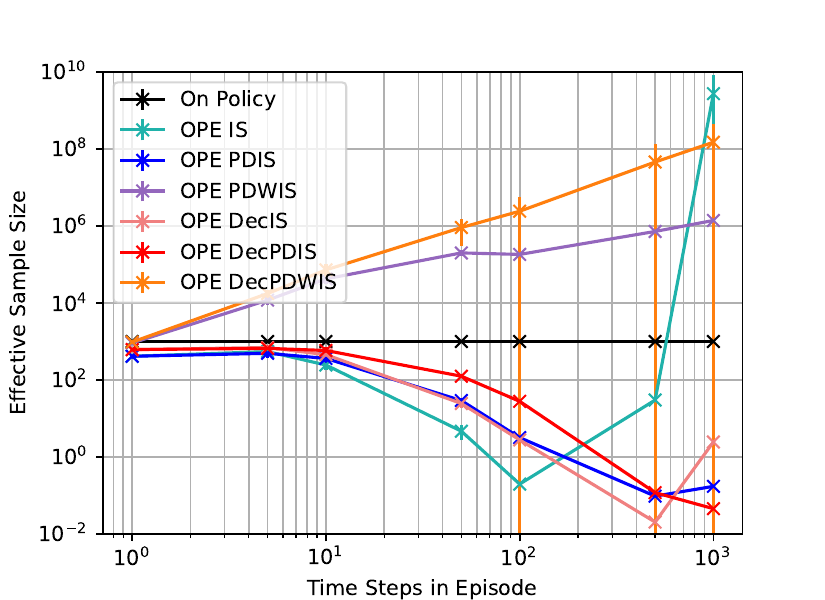}
    \caption{Here, nearly all transitions, even those far into the future, have a high contribution to the value function estimate. It is clear that a high discount factor causes rapid scaling in estimator variance and loss of coverage which leads to scaling in estimator bias. The MSE's of the estimators thus scale faster than the on-policy MSE. Interestingly, however, the variances of the PDWIS estimators seem relatively unaffected by variation in the discount factor; perhaps the effect of loss in coverage is counterbalanced by the effect of denominator weighting.}
    \label{fig:var-vs-trajectories-discount-0-999}
\end{figure}

\newpage

\subsubsection{Varying Policy Divergence in MDP 2 With 1000 Trajectories of Length 10 and Discount Factor 0.7}

\begin{figure}[h!]
    \centering
    \includegraphics[scale=0.5]{images/4-state-MDP-experiments/bias-10-steps-vs-policy-divergence.pdf}
    \includegraphics[scale=0.5]{images/4-state-MDP-experiments/var-10-steps-vs-policy-divergence.pdf}
    \includegraphics[scale=0.5]{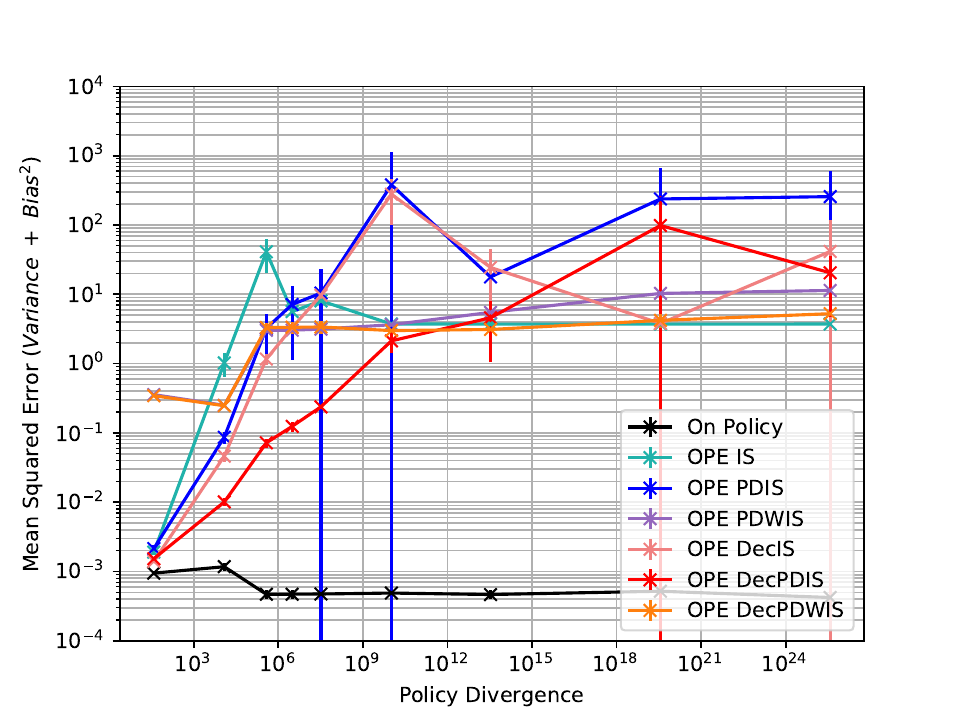}
    \includegraphics[scale=0.5]{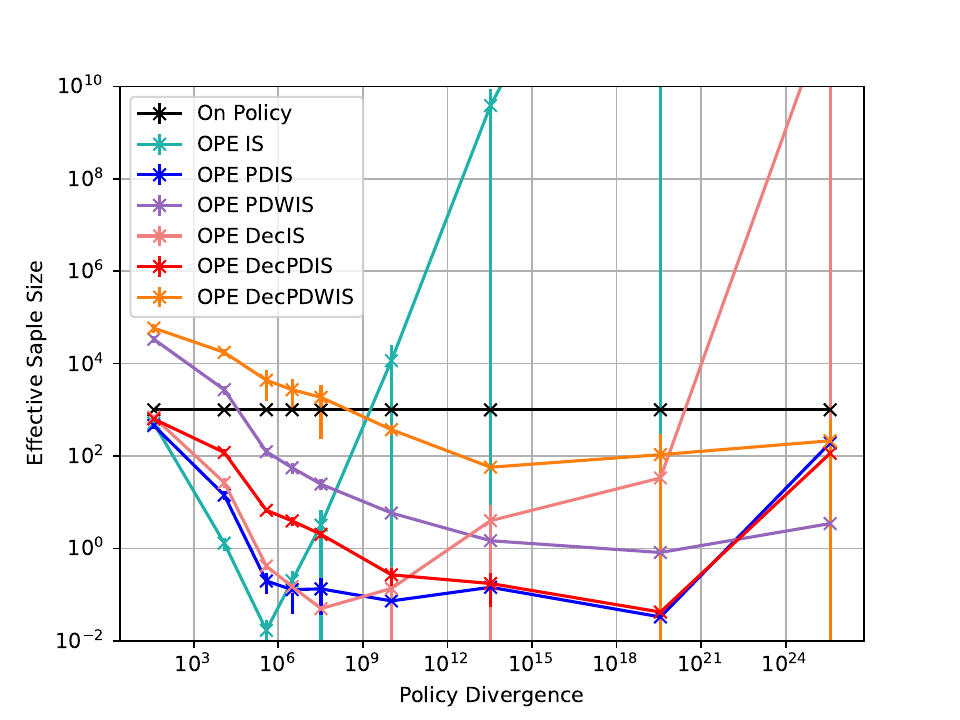}
    \caption{The bias and variance graphs are the same as those in Figure \ref{fig:var-10-steps-vs-policy-divergence} from the main report and have been discussed there in detail. A notable insight is how rapidly coverage of the evaluation policy decreases as policy divergence increases. This can be seen from the rapid scaling in bias, and the variance graphs turning downwards after initial growth. The MSE behaviour is dominated by bias, while ESS seems almost identical to the variance graph but inverted.}
    \label{fig:var-vs-policy-divergence-N-1000}
\end{figure}

\newpage







\end{document}